\documentclass{clv3}

\skip\footins=23.0pt plus 8.0pt
\skip\footins=23.0pt plus 8.0pt
\def\trimmarks{}

\usepackage{hyperref}
\usepackage{xcolor}
\definecolor{darkblue}{rgb}{0, 0, 0.5}
\hypersetup{colorlinks=true,citecolor=darkblue, linkcolor=darkblue, urlcolor=darkblue}

\usepackage{graphicx}
\usepackage{xspace}
\newcommand{\corpusname}{crowd-en\textsc{Vent}\xspace}
\usepackage{booktabs}
\usepackage{tabularx}
\usepackage{array}
\usepackage{multirow}
\usepackage{microtype} 
\usepackage{tikz}
\usetikzlibrary{arrows,shapes,backgrounds,positioning,decorations.pathreplacing}
\usepackage{subfig}
\usepackage{wrapfig}
\usepackage[inline]{enumitem}
\usepackage{longtable}
\usepackage{soul}
\newlist{compactenum}{enumerate}{4}
\setlist[compactenum,1]{nolistsep,label=(\arabic*),leftmargin=6mm}
\newlist{compactitem}{itemize}{4}
\setlist[compactitem,1]{nolistsep,label=$\bullet$}

\newcommand{\TA}{$\text{T$\rightarrow$A}$\xspace}
\newcommand{\TE}{$\text{T$\rightarrow$E}$\xspace}
\newcommand{\aE}{$\text{A$\rightarrow$E}$\xspace}
\newcommand{\TAE}{$\text{TA$\rightarrow$E}$\xspace}

\newcommand{\TAhuman}{$\text{\TA} \atop \text{human}$\xspace}
\newcommand{\TAmodel}{$\text{\TA} \atop \text{model}$\xspace}
\newcommand{\TEhuman}{$\text{\TE} \atop \text{human}$\xspace}
\newcommand{\TEmodel}{$\text{\TE} \atop \text{model}$\xspace}
\newcommand{\TAEmodelGold}{$\text{TA}^{\text{Gold}}\rightarrow\text{E} \atop \text{model}$\xspace}
\newcommand{\TAEmodelPred}{$\text{TA}^{\text{Pred}}\rightarrow\text{E} \atop \text{model}$\xspace}
\newcommand{\AEmodelGold}{$\text{A}^{\text{Gold}}\rightarrow\text{E} \atop \text{model}$\xspace}
\newcommand{\AEmodelPred}{$\text{A}^{\text{Pred}}\rightarrow\text{E} \atop \text{model}$\xspace}

\newcommand{\TAhumanText}{$\text{\TA}_{\text{human}}$\xspace}
\newcommand{\TAmodelText}{$\text{\TA}_{\text{model}}$\xspace}
\newcommand{\TEhumanText}{$\text{\TE}_{\text{human}}$\xspace}
\newcommand{\TEmodelText}{$\text{\TE}_{\text{model}}$\xspace}

\newcommand{\TAEmodelGoldText}{$\text{TA}^{\text{Gold}}\rightarrow\text{E}_{\text{model}}$\xspace}
\newcommand{\TAEmodelPredText}{$\text{TA}^{\text{Pred}}\rightarrow\text{E}_{\text{model}}$\xspace}
\newcommand{\AEmodelGoldText}{$\text{A}^{\text{Gold}}\rightarrow\text{E}_{\text{model}}$\xspace}
\newcommand{\AEmodelPredText}{$\text{A}^{\text{Pred}}\rightarrow\text{E}_{\text{model}}$\xspace}

\newcommand{\emotionname}[1]{\textit{#1}}
\newcommand{\appraisalname}[1]{\textit{#1}}
\newcommand{\fear}{\emotionname{fear}\xspace}
\newcommand{\joy}{\emotionname{joy}\xspace}
\newcommand{\anger}{\emotionname{anger}\xspace}
\newcommand{\boredom}{\emotionname{boredom}\xspace}

\newcommand{\trust}{\emotionname{trust}\xspace}

\newcommand{\guilt}{\emotionname{guilt}\xspace}
\newcommand{\shame}{\emotionname{shame}\xspace}
\newcommand{\pride}{\emotionname{pride}\xspace}

\newcommand{\surprise}{\emotionname{surprise}\xspace}
\newcommand{\sadness}{\emotionname{sadness}\xspace}

\newcommand{\disgust}{\emotionname{disgust}\xspace}
\newcommand{\noemotion}{\emotionname{no-emotion}\xspace}
\newcommand{\relief}{\emotionname{relief}\xspace}

\newcommand{\suddenness}{\appraisalname{suddenness}\xspace}
\newcommand{\familiarity}{\appraisalname{familiarity}\xspace}
\newcommand{\eventpredictability}{\appraisalname{event predictability}\xspace}
\newcommand{\pleasantness}{\appraisalname{pleasantness}\xspace}
\newcommand{\unpleasantness}{\appraisalname{unpleasantness}\xspace}
\newcommand{\goal}{\appraisalname{goal relevance}\xspace}
\newcommand{\ownresponsibility}{\appraisalname{own responsibility}\xspace}
\newcommand{\otherResp}{\appraisalname{other responsibility}\xspace}
\newcommand{\situationalResp}{\appraisalname{situational responsibility}\xspace}
\newcommand{\goalSupport}{\appraisalname{goal support}\xspace}
\newcommand{\anticipationConseq}{\appraisalname{anticip.\ conseq.}\xspace}
\newcommand{\urgent}{\appraisalname{urgency}\xspace}
\newcommand{\acceptConseq}{\appraisalname{accept.\ conseq.}\xspace}
\newcommand{\internalStandards}{\appraisalname{internal standards}\xspace}
\newcommand{\externSocialStandards}{\appraisalname{external norms}\xspace}
\newcommand{\attention}{\appraisalname{attention}\xspace}
\newcommand{\notconsider}{\appraisalname{not consider}\xspace}
\newcommand{\effort}{\appraisalname{effort}\xspace}
\newcommand{\ownControl}{\appraisalname{own control}\xspace}
\newcommand{\othercontrol}{\appraisalname{others' control}\xspace}
\newcommand{\situationalControl}{\appraisalname{situational control}\xspace}

\newcommand{\avg}{avg.}
\newcommand{\F}{$\textrm{F}_1$\xspace}

\issue{x}{x}{2022}

\dochead{}

\runningtitle{Dimensional Modeling of Emotions with Appraisal Theories}

\runningauthor{Troiano, Oberl\"ander, Klinger}

\pageonefooter{Action editor: Saif M. Mohammad. Submission received: 10 June 2022; revised version received: 25 July 2022; accepted for publication: 24 August 2022.}

\begin{document}

\title{Dimensional Modeling of Emotions in Text with Appraisal
Theories: Corpus Creation, Annotation Reliability, and Prediction}

\author{Enrica Troiano\thanks{All authors contributed equally.}}
\affil{Institut f\"ur Maschinelle Sprachverarbeitung\\
  University of Stuttgart\\
  \texttt{enrica.troiano\\@ims.uni-stuttgart.de}}

\author{Laura Oberl\"ander$^*$}
\affil{Institut f\"ur Maschinelle Sprachverarbeitung\\
  University of Stuttgart\\
  \texttt{laura.oberlaender\\@ims.uni-stuttgart.de}}

\author{Roman Klinger$^*$}
\affil{Institut f\"ur Maschinelle Sprachverarbeitung\\
  University of Stuttgart\\
  \texttt{roman.klinger\\@ims.uni-stuttgart.de}}

\maketitle

\begin{abstract}
  The most prominent tasks in emotion analysis are to assign emotions
  to texts and to understand how emotions manifest in language. An important
  observation for natural language processing is that emotions can be
  communicated implicitly by referring to events alone, appealing to
  an empathetic, intersubjective understanding of events, even without
  explicitly mentioning an emotion name. In psychology, the class of
  emotion theories known as appraisal theories aims at explaining the
  link between events and emotions. Appraisals can be formalized as
  variables that measure a cognitive evaluation by people living
  through an event that they consider relevant. They include the
  assessment if an event is novel, if the person considers themselves
  to be responsible, if it is in line with the own goals, and many
  others. Such appraisals explain which emotions are developed based
  on an event, e.g., that a novel situation can induce surprise or one
  with uncertain consequences could evoke fear.
  We analyze the suitability of appraisal theories for emotion
  analysis in text with the goal of understanding if appraisal concepts
  can reliably be reconstructed by annotators, if they can be
  predicted by text classifiers, and if appraisal concepts help to
  identify emotion categories. To achieve that, we compile a corpus
  by asking people to textually describe events that triggered
  particular emotions and to disclose their appraisals. Then, we ask
  readers to reconstruct emotions and appraisals from the text. This
  setup allows us to measure if emotions and appraisals can be
  recovered purely from text and provides a human baseline to judge
  model's performance measures.
  Our comparison of text classification methods to human annotators
  shows that both can reliably detect emotions and appraisals with
  similar performance. Therefore, appraisals constitute an alternative
  computational emotion analysis paradigm and further improve the
  categorization of emotions in text with joint models.
\end{abstract}

\section{Introduction}
\label{sec:introduction}
Voices that have had a say about the affective life of humans have
raised from multiple disciplines. Over the centuries, philosophers,
neuroscientists, cognitive and computational researchers have been
drawn to the study of passions, feelings and sentiment
\cite{solomon1993philosophy,adolphs2017should,oatley2014cognitive,karg2013body}.
Among such affective phenomena, emotions stand out.  For one thing,
they are many: While sentiment can be described with a handful of
categories (e.g., neutral, negative, positive), it takes a varied
vocabulary to distinguish the mental state that accompanies a cheerful
laughter from that enticing a desperate cry, one felt before
a danger from one arising with an unexpected discovery (e.g., joy,
sadness, fear, surprise).  These seemingly understandable experiences
are also complex to define.  Psychologists diverge on the formal
description of emotion -- both of emotion as a coherent whole, and of
emotions as many differentiated facts. What has ultimately been agreed
upon is that emotions can be studied systematically
\cite[cf.][p. 338]{dixon2012emotion}, and that people use specific
``diagnostic features'' to recognize them \cite{Scarantino2016}. They
are the presence of a stimulus event, an assessment of the event based
on the concerns, goals and beliefs of its experiencer and some
concomitant reactions (e.g., the cry, the laughter).

Like other aspects of affect, emotions emerge from language
\cite{wierzbicka1994emotion}; as such, they are of interest for
natural language processing (NLP) and computational linguistics
\cite{sailunaz2018emotion}. The cardinal goal of computational emotion
analysis is to recognize the emotions that texts elicit in the
readers, or those that pushed the writers to produce an utterance in
the first place.  Irrespective of their specific subtask,
classification studies start from the selection of a theory from
psychology, which establishes the ground rules of their object of
focus. Commonly used frameworks are the Darwinistic perspectives of
\namecite{Ekman1992} and \namecite{Plutchik2001}.  They depict
emotions in terms of an evolutionary adaptation that manifests in
observable behaviors, with a small nucleus of experiences that
intersect all cultures. \textit{Discrete} states like anger, fear, or
joy are deemed universal, and thus constitute the phenomena to be looked for
in text.  Besides basic emotions, much research has leveraged a
\textit{dimensional} theory of affect, namely the circumplex model by
\namecite{Posner2005}. It consists of a vector space defined by the
dimensions of valence (how positive the emoter feels) and arousal (how
activated), which enable researchers to represent discrete states in a
continuous space, or to have computational models exploit continuous
relations between crisp concepts, in alternative to predefined emotion
classes.  Further, some works acknowledge the central role of events
in the taking place of an emotion, and the status of emotions as events
themselves. These works constitute a special case of semantic role
labeling, primarily aimed at detecting precise aspects of emotional
episodes that are mentioned in text, like emotion stimuli
\cite{Bostan2020,Kim2018,Mohammad2014,Xia2019a}.

Studies assigning texts to categorical emotion labels
\cite[i.a.]{mohammad-2012-emotional,Klinger2018x}, to subcomponents
of affect \cite[i.a.]{Preotiuc2016,Buechel2017} or of events, have a
pragmatic relationship to the chosen psychological models. They use
theoretical insights about which emotions should be considered (e.g.,
anger, sadness, fear) and how these can be described (e.g., by means
of discrete labels), but they do not account for what emotions
are. In other words, they disregard a crucial diagnostic feature of
emotions, namely, that emotions are reactions to events that are
\textit{evaluated} by people.  The ability of evaluating an
environment allows humans to figure out its properties (if it is
threatening, harmless, requires an action, etc.), which in turn
determine if and how they react emotionally.  Therefore, to overlook
evaluations is to dismiss a primary emotion resource, and most
importantly for NLP, a tool to extrapolate affective meanings from
text.

The relevance of evaluations in text becomes clear considering
mentions of factual circumstances. Writers often omit their emotional
reactions, and they only communicate the eliciting event.  In such
cases, an emotion emerges if the readers carry out an 
interpretation, engaging their knowledge about event participants,
typical responses, possible outcomes, and world relations.  For
instance, it is thanks to an (extra-linguistic) assessment that texts
like ``the tyrant passed away'' and ``my dog passed away'' can be
associated with an emotion meaning, and specifically, with
different meanings. The two sentences describe semantically similar
situations (i.e., death), but their subjects change the comprehension
of how the writer was affected in either case.  Accordingly, the
first text can be charged of relief while the other likely expresses
sadness.

While not directly addressing texts, psychology has produced abundant
literature on the relationship between emotions and evaluations.
Appraisal theories are an entire class of frameworks that has
discussed emotions in terms of the cognitive
\textit{appraisal} of an event, together with the subjective feelings,
action tendencies, physiological reactions, and bodily and vocal
expressions that the event can trigger
\cite{staller2001introducing,gratch2009assessing}. All of these
factors are relevant for computational linguistics because they
realize in language \cite{DeBruyne2021,Casel2021} -- e.g., writers can
describe their verbal (``oh, wow'') or motor response to a situation (``I felt
paralyzed!'') in order to convey an emotion. However,
the appraisal component plays a special part.  Appraisal theorists
elaborate extensively and variously on its contribution in an emotion
experience.  In the OCC view, which is a specific appraisal-based approach
named after its authors Orthony, Clore and Collins \cite{Clore2013}, an
appraisal is a sequence of binary evaluations that concern events,
objects, and actions (i.e., how good or bad, pleasant or unpleasant
they are, whether they match social and personal moral standards).  By
contrast, scientists like \namecite{Smith1985} and \citet{Scherer2005},
who organize the emotion components into a holistic process, qualify
appraisals with more detailed criteria, as dimensions along which
people assess events: ``is it pleasant?'', ``did I see that coming?'',
``do I have control over its development?'', ``do I expect an outcome
in line with my goals?''.  Different combinations of these dimensions
correspond to different emotions. Intuitively, unpleasantness and the
hampering of one's goals could elicit anger; unpleasantness,
unexpectedness and a low degree of control could induce fear.

The latter approach has found its way into computational research,
mainly to make robot agents aware of social processes
\cite{Kim2010,Breazeal2016}.  To us, it represents a promising avenue
also for emotion analysis in text. The evaluation criteria of
\namecite{Smith1985} and \citet{Scherer2005} can be leveraged to
explain why linguistically similar texts convey opposite emotions
(e.g., ``the tyrant passed away'' and ``my dog passed away'' are
assigned different properties, like pleasantness and alignment with
one's goals). Hence, appraisals\footnote{In this work, we use
  ``appraisals'' and ``appraisal dimensions'' interchangeably.} can
bring valuable information for annotation studies. Collecting these
types of judgments might reveal why annotators picked a certain
emotion label (e.g., they appraised the described event differently in
the first place), and might eventually disclose underlying patterns in
their disagreement.  As for emotion classification, the fine-grained
appraisal dimensions discussed above provide a more expressive tool
than basic-, dimensional-, and OCC-based models. Endowed with such
representations, systems might ultimately turn more human-like and
theoretically grounded: since appraisal dimensions are a finite set of
features, they can formalize differences between events, possibly
promoting better classification performances.

In this work, we put these ideas into scrutiny. We aim at
understanding if appraisal theories (specifically, the component
process model) can be used in the field of emotion analysis and
advance it.  Much in the way in which past work has predicted the
emotion of texts' writers via readers and computational models, we
investigate if the evaluations/appraisals carried out by event
experiencers can be reconstructed, given the texts in which they
mention such events, by humans and by automatic classifiers.

Evaluations of emotion-inducing events have actually been leveraged in
NLP, but only by a handful of studies. Of this type are
\citet{Shaikh2009,balahur2011building,Balahur2012,Hofmann2020,Hofmann2021}
and \citet{Troiano2022}. These works proposed approaches to make
emotion categorization decisions motivated by appraisal theories, but
they did not analyze the suitability of these theories for NLP.
Understanding the limits and possibilities of an appraisal-oriented
approach to emotion analysis poses indeed a major challenge: there is
no available corpus that contains annotations of our concern (i.e.,
provided by first-hand event experiencers). A useful and public
resource with a machine-learning appropriate size exists (i.e., ISEAR
by \citet{Scherer1997}), but its texts are unpractical for us, because
they were produced by a combination of native and non-native speakers,
who only consisted of college students. ISEAR was not compiled for
purposes of text analysis, but to investigate the relation between
appraisals and emotions; and no validation of the annotations has been
performed in the same experimental environment.  To solve these
issues, we crowdsource a corpus of emotion-inducing event descriptions
produced by English native speakers, annotated with emotions, event
evaluations (using 21 appraisals), stable properties of the texts'
authors (e.g., demographics, personality traits) and contingent
information concerning their state at the moment of taking our study
(i.e., their current emotion).  The resulting collection, to which we
refer as \corpusname\footnote{This name indicates that is has been
  crowdsourced and is in English. This is in contrast to our corpus
  x-enVENT \cite{Troiano2022}, which has been annotated by trained
  experts with similar variables. It constitutes a preparatory study
  to \corpusname.}, encompasses 6600 instances. Part of it is
subsequently annotated by external crowdworkers, tasked to read the
descriptions and to infer how the authors originally appraised the
events in question.

Dealing with texts that convey subjective experiences, our approach
also relates to some research lines in sentiment analysis aimed at
recognizing ``people’s opinions, sentiments, evaluations, appraisals,
attitudes, [...] towards entities such as products, [...] issues,
events, topics, and their attributes'' \cite{liu2012sentiment}.  Rich
literature can be found on implicit expressions of polarized
evaluations, but it targets specific types of opinions, e.g., those
expressed in business news \cite{jacobs2021fine,jacobs2022sentivent}
and in meeting discussions \cite{wilson-2008-annotating}.  Much of
such work has the goal to understand if texts contain evaluations
\cite{toprak-etal-2010-sentence}, or how their polarity can be traced
back to specific linguistic cues, like negations and diminishers
\cite{musat2010impact}, indirectly valenced noun phrases
\cite{zhang-liu-2011-identifying} and their combination with verbs and
quantifiers \cite{zhang-liu-2011-extracting}.  By contrast, we do not
restrict ourselves to any type of event; most importantly, we relate
evaluations to people's background knowledge with the
theoretically-motivated taxonomy of 21 appraisals, to make the type of
evaluations behind an emotion experience and an emotion judgment
transparent.

Our study revolves around four research questions.  (RQ1) Is there
enough information in a text for humans and classifiers to predict
appraisals?  (RQ2) How do appraisal judgments relate to textual
properties? (RQ3) Can an appraisal or an emotion be reliably inferred
only if the original event experiencer and the text annotator share
particular properties? (RQ4) Do appraisals practically enhance emotion
prediction?  By leveraging \corpusname\footnote{Data and code are
  available at
  \url{https://www.ims.uni-stuttgart.de/data/appraisalemotion}.}, we
investigate if (and to what extent) people's appraisals can be
interpreted from texts, and if models' predictions are more similar to
those who lived the experience first-hand, or resemble more the
external judges' (RQ1). To gain better insight, we analyze the data
and classification models qualitatively (RQ2). Further, we verify if
the sharing of stable/contingent properties between the texts'
generators and validators, including demographics, personality traits
and cultural background, affects the similarity of their judgments
(RQ3). Lastly, narrowing the focus on emotion classification, we
evaluate if and in what case this task benefits from appraisal
knowledge.  More specifically, we compare human performance to that of
computational models that predict emotions and appraisals, separately
or jointly (RQ4).

In sum, we present a twofold contribution to the field. First, we
propose appraisal-based emotion analysis with a rich set of variables
that has never been investigated before in NLP: we cast a novel
paradigm that complements models of basic emotions and dimensional
models of affect, showing that appraisal dimensions can be useful to
infer some mental states from text.  Appraisal information indeed
proves a valuable contribution to emotion classifiers and constitutes
a prediction target itself. It comes with the advantage of being
interpretable, as are basic emotion names, and dimensional, as is 
affect in dimensional models that enable to
measure similarities between emotions. Second, we introduce a corpus
of event descriptions richly annotated with appraisals from the
perspectives of both writers and readers, that we compare. Besides
emotion classification in general, and for the track of investigation
interested in differences between annotation perspectives, our
resource can be a benchmark for research focused on human evaluations
of real-life circumstances.  Lastly, for psychology, our study
represents a computational counterpart of previous work, which
encompasses a large set of appraisal variables and reveals how well
they transfer to the domain of language.

This paper is structured as follows. In
Section~\ref{sec:emotion-theories}, we review research on emotions from
psychology and NLP to draw a parallel between the two.
Section~\ref{sec:reliability} provides an overview of our study and
introduces essential concepts for our study design. It further
illustrates how previous work has (or has not) addressed them. It
presents the problem of emotion recognition in psychology, which is
mostly based on facial interpretations, and from the NLP side it
discusses measures of annotation agreement. Next, we explain our data
collection procedure (Section~\ref{sec:creation}) and analyze it
(Section~\ref{sec:analysis}). The resulting insights constitute a
motivation as well as a baseline for the modeling experiments,
described in Section~\ref{sec:modeling}.  The paper concludes with a
discussion of the limitations of our approach, some possible
solutions, interesting ventures for future work, and ethical points of
concern.

\section{Emotion Theories and Their Application in Natural Language
  Processing}
\label{sec:emotion-theories}

Emotions represent an interdisciplinary challenge. Explaining what they are and
how they arise is an attempt that can take substantially different paths,
depending on the considered types of episodes (anger, joy, etc.), the
underlying mechanisms that one looks at, and their meaning in language. The
insights provided by different directions in the literature share some
commonalities nevertheless. This suggests that the corresponding approaches in
computational emotion analysis are also not in conflict, but rather complement
each other. In the following, we give an overview of previous work both from
psychology and NLP to contextualize the appraisal theories employed in this
study. We specifically follow the organization of \namecite[p.\
8]{Scarantino2016}, who divides psychological currents on the topic into a
\textit{feeling tradition}, a \textit{motivational tradition}, and an
\textit{evaluative tradition}.

\subsection{Feeling and Affect}

In the \textit{feeling tradition}, emotions are not seen as innate
universals.  They are learned constructs whose development relies on
culture and contingent situations.  \textit{Constructionist}
approaches are one instance of this tradition \cite{James1894}.
Pioneered by William James, they theorize that ``bodily changes follow
directly the perception of the exciting fact, and that our feeling of
the same changes as they occur is the emotion''. James claims that
``we feel sorry because we cry, angry because we strike, afraid
because we tremble, and not that we cry, strike, or tremble, because
we are sorry, angry, or fearful'' \citep[reported from][]{Myers1969}.

The perception-to-emotion view has sparked heated debates, with the
counter-argument that humans' emotional processes do not unfold in
such a strict sequential order.  Contemporary constructionists address
this criticism by explaining that emotions are shaped dynamically.
The ``brain prepares multiple competing simulations that answer the
question, what is this new sensory input most similar to?'' \cite[p.\
7]{Barrett2017a}. This similarity calculation is based on perception,
energy costs, and rewards for the body. Therefore, emotions are
constructed thanks to the engagement of resources that are not
specific to an emotion module, similar to the building blocks of an
algorithm that could be arranged to create alternative instructions
\cite[i.a.]{Barrett2017a}. One of the basic pieces out of which
emotions are constructed is \textit{affect}, or ``the general sense of
feeling that you experience throughout each day [...]  with two
features. The first is how pleasant or unpleasant you feel, which
scientists call valence. [...] The second feature of affect is how
calm or agitated you feel, which is called arousal''
\cite[p. 72]{FeldmanBarrett2018}.  Hence, the simulation process links
affect to a complex emotion perception.

\begin{figure}
  \centering
  \includegraphics[scale=0.7]{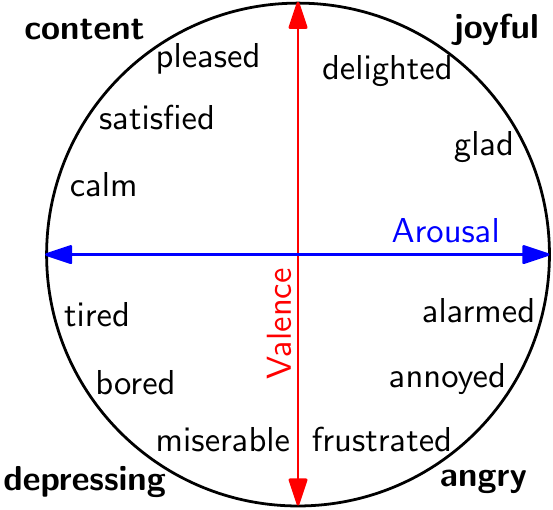}
  \caption{The circumplex model of emotions, a dimensional emotion model.}
  \label{fig:posner}
\end{figure}

Other constructionist theorists relate emotions with affect as well.
\namecite{Posner2005}, for instance, assign emotions to specific positions
within the circumplex model depicted in Figure~\ref{fig:posner}, a continuous
affective space that is defined by the dimensions of valence and arousal.
\citet{Bradley1994} extend this model to a valence-arousal-dominance (VAD) one.
There, emotions vary from one another in regard to the three VAD factors, with
dominance representing the power that an experiencer perceives to have in a
given situation.

A wave of studies based on affect also exists in NLP. It has been dedicated to
predicting the continuous values of valence, arousal, and dominance, defining a
regression task that is commonly solved with deep-learning systems, sometimes
informed by lexical resources \citep[i.a.]{Wei2011,Buechel2016regression,
Wu2019, Cheng2021}. Dimensional models of emotion have indeed many advantages
from a computational perspective. They formalize relations between emotions in
a computationally tractable manner,  e.g., models learn that texts expressing
sadness and those conveying anger are both characterized by low valence. This
means that in an affect recognition task, machine learning systems bypass the
decision of picking one out of various states that are similar to one another
in respect to some dimensions, and which could equally hold for a given text.
In fact, at modeling time, researchers are not compelled to be provided with
categorical emotion information at all. Systems only need to learn relations
between valence and arousal, and in case the final goal is to classify texts
with discrete emotions, the VA(D)-to-emotion mapping can be left as a step
outside the machine learning task.

Still, there have been attempts to integrate the dimensional model with
discrete emotions. \namecite{Park2021} propose a framework to learn a joint
model that predicts fine-grained emotion categories together with continuous
values of VAD. They do so using a pretrained transformer-based model
\citep[namely RoBERTa, ][]{Liu2019Roberta}, fine-tuned with earth movers
distance \citep{Rubner2000} as a loss function to perform classification.
Related approaches learn multiple emotion models at once, showing that a
multi-task learning of discrete categories and VAD scores can benefit both
subtasks \citep{Akhtar2019,Mukherjee2021}. Particularly interesting for our
work is the study by \namecite{Buechel2021}. They define a unified model for a
shared latent representation of emotions, which is independent from the
language of the text, the used emotion model, and the corresponding emotion labels.
In a similar vein, we aim at integrating appraisal theories with discrete
emotion experiences, seeing the dimensions coming from the former as a latent
representation of the latter.

Many efforts in NLP focus on (automatically) creating lexicons. Terms
are assigned VAD scores based on their semantic similarity to other
words, for which manual annotations are provided
\cite{Koper2017,Buechel2016}. To date, lexicons are available for both
English \cite{Bradley1999,Warriner2013,Mohammad2018vad} and other
languages (e.g., \citet{Buechel2020} created lexicons for 91 language,
including Korean, Slovak, Icelandic, Hindi), and so are corpora
annotated at the sentence or paragraph level with (at least a subset
of) VAD information -- among others are \citet{Preotiuc2016},
\citet{Buechel2017readers} and \citet{Buechel2017} for English,
\citet{Yu2016} for Mandarin, \citet{Mohammad2018} for Spanish and
Arabic.  Using corpora, research has investigated how the valence and
arousal that emerge from text co-vary with some attributes of the
writers, such as age and gender \cite{Preotiuc2016}. Moreover, it has
revealed that annotators who infer emotions from text by attempting to
assume the writer's perspective achieve higher inter-annotator
agreement than those who report their personal reactions
\cite{Buechel2017readers}. The finding that the quality of an
annotation effort can change depending on the perspective of text
understanding will turn out crucial for the design decision of our
work.

\subsection{Motivation and Basic Emotions}
\label{sec:basicemotions}
\begin{figure}
  \centering
  \newcommand{\incl}[1]{\includegraphics[width=17mm]{figures/#1}}
  \subfloat[][Ekman's model]{%
    \sf
    \raisebox{\height}{%
      \begin{tabular}{|ccc|}\hline
        \incl{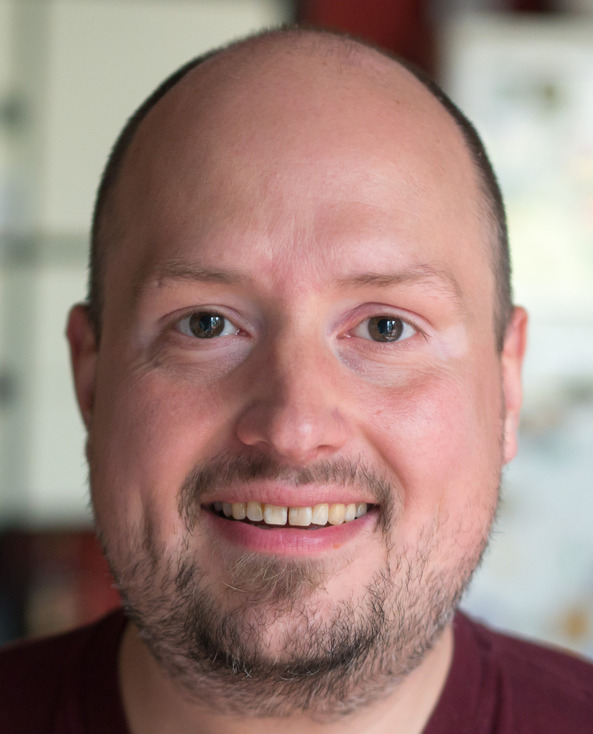}&\incl{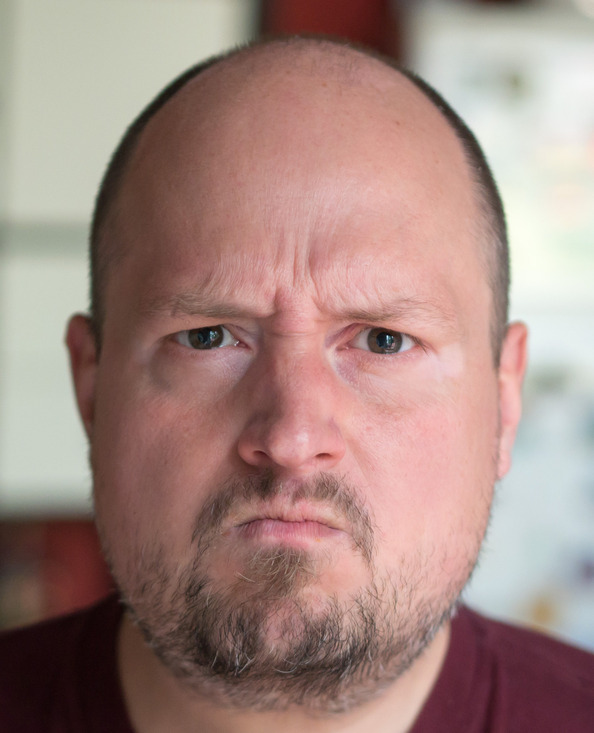}&\incl{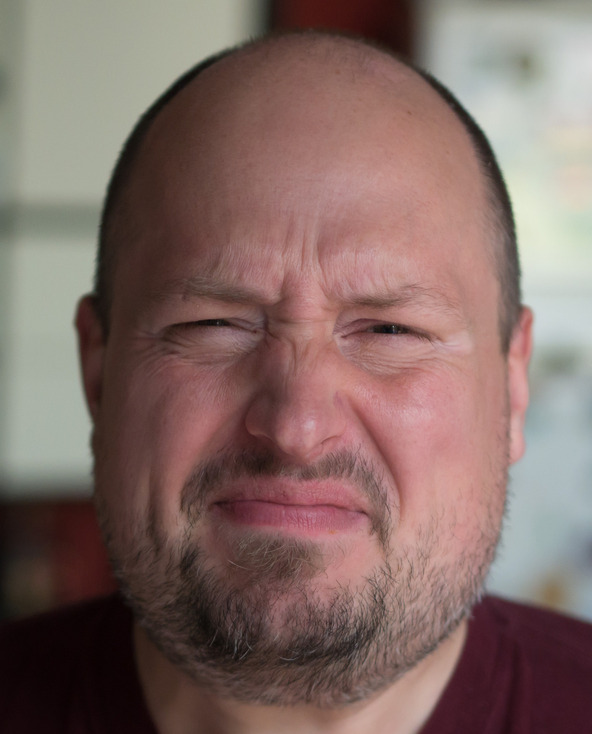}\\[-3pt]
        Joy&Anger&Disgust\\\hline
        \incl{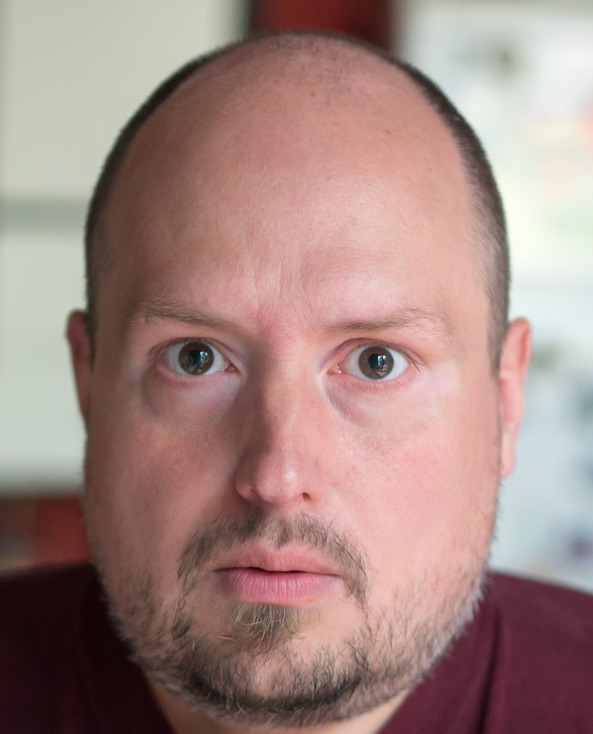}&\incl{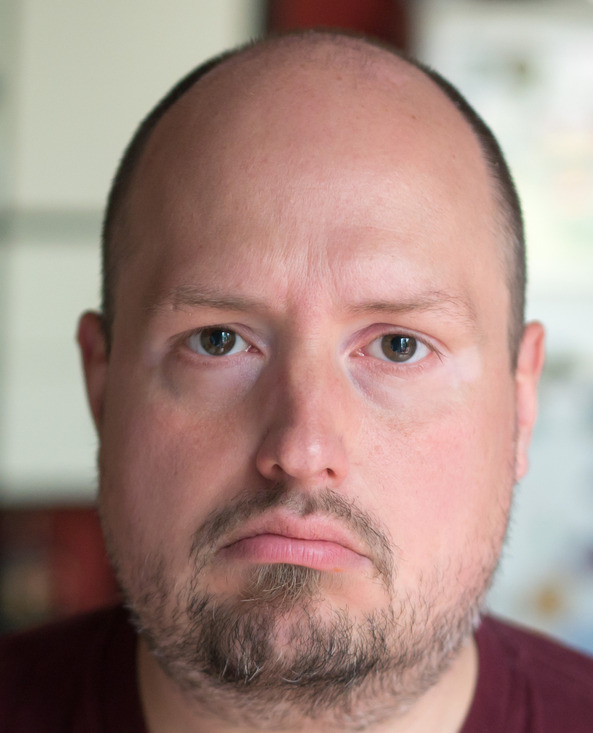}&\incl{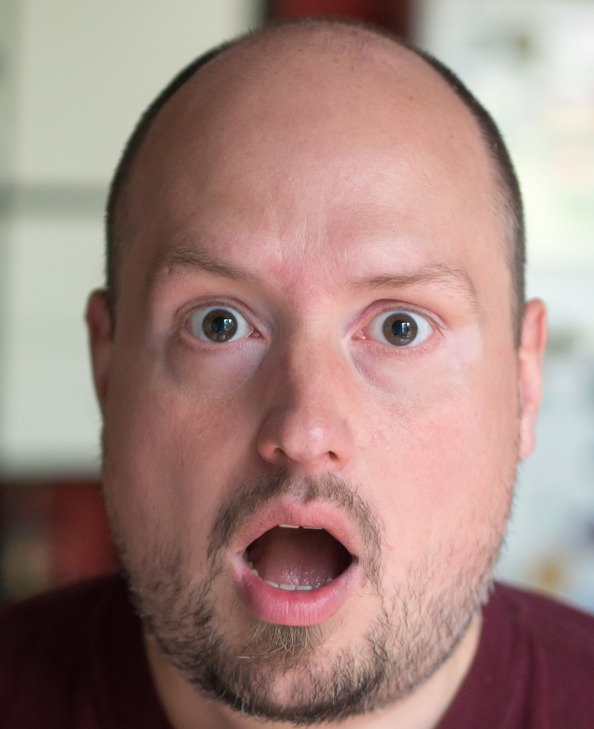}\\[-3pt]
        Fear&Sadness&Surprise\\\hline
      \end{tabular}
    }
    \label{fig:ekman}}\hfill
  \subfloat[][Plutchik's Wheel of Emotions]{\includegraphics[width=0.48\linewidth]{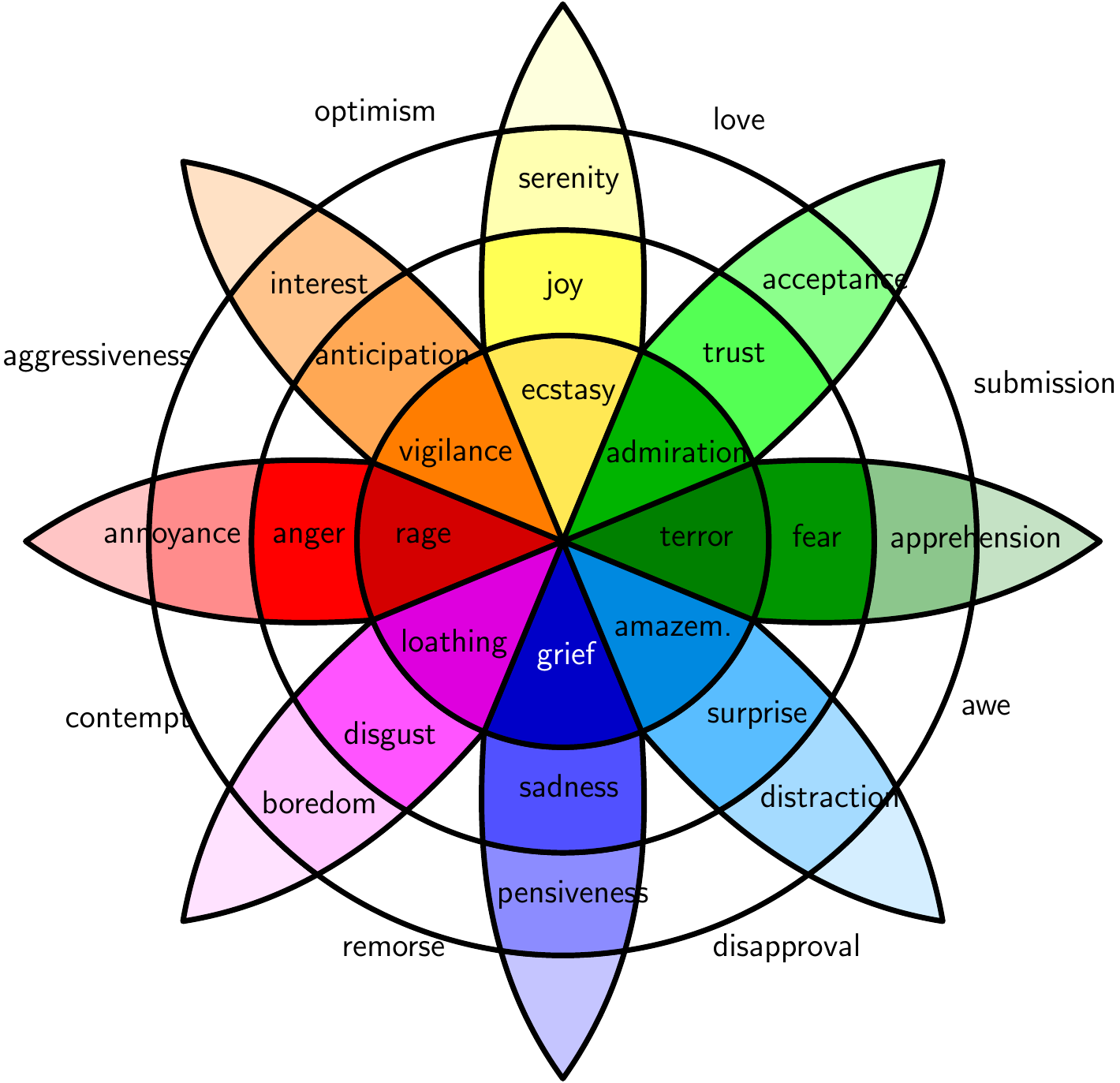}\label{fig:plutchik}}
  \caption{Visualizations of Basic Emotion Models.}
  \label{fig:ekmanplutchik}
\end{figure}
The \textit{motivational tradition} includes ``theories of basic
emotion'', of which \namecite{Ekman1992} is a prominent
representative.  Ekman's research is characterized by a Darwinistic
approach: aimed at measuring observable phenomena, it qualifies as
basic emotions those found among other primates, that have precise
universal signals, a quick onset, a brief duration, an unbidden
occurrence, coherence among instances of the same emotion, distinctive
physiology, and importantly for our work, distinctive universals in
antecedent events and an automatic appraisal.  The idea that emotions
can be distinguished by their physiological manifestation pushed
research in psychology to investigate and code the movements of facial
muscles \cite{Clark2020}, with specific configurations corresponding
to specific emotions. Hence, the basic emotions of fear, anger, joy,
sadness, disgust, and surprise are commonly illustrated with
depictions similar to Figure~\ref{fig:ekman}.

The definition of what constitutes a basic emotion is different in the
Wheel of Emotions \cite{Plutchik2001} illustrated in
Figure~\ref{fig:plutchik}.  As \namecite{Scarantino2016} puts it based
on \namecite{Plutchik1970}, an emotion is ``a patterned bodily
reaction of either protection, destruction, reproduction, deprivation,
incorporation, rejection, exploration or orientation'' (p.\
12). According to Plutchik, each reaction function corresponds to a
primary emotion, namely fear, anger, joy, sadness, acceptance,
disgust, anticipation, and surprise. Primary emotions can be composed
to obtain others, like colors, and they are characterized by their
intensity gradation. The wheel includes indeed a dimension of
intensity (in/outside), similar to the variable of arousal (e.g.,
higher intensity -- darker color: ecstasy; lower intensity -- fairer
gradation: serenity).  In this sense, Plutchik links discrete emotion
theories with dimensional ones.

Theories of basic emotions constitute an (often tacit) argument used
in NLP: different emotions can be clearly recognized not only via
faces but also when the communication channel is text. This is the
main notion that computational studies of emotions borrow from basic
emotion theories in psychology, although the latter offers a much more
varied picture. For example, Ekman also describes non-basic emotions
as ``emotional plots'', moods, and affective personality
traits. Further, he characterizes (basic and non-basic emotions) as
``programs'', which lead to a sequence of changes, when
activated. These changes include action tendencies, alterations in
one's face, voice, autonomic nervous system, and body; plus, they
trigger the retrieval of memories and expectations (cf.\
constructionist theories), which guide how we interpret what is
happening within and around us.  If emotions denote categorical
states, their perception happens thanks to the contribution of
multiple components -- an idea that remains overlooked in NLP.

Early attempts to link language and emotions focus on the construction
of lexicons. An example is the Linguistic Inquiry and Word Count
(LIWC), aimed at providing a list of words that are reliably
associated with psychological concepts across domains and application
scenarios \citep{Pennebaker2001}.  Both this lexicon and the
associated text processing software are well-rooted in psychological
concepts, with emotions being only a subset of the labels. Instead,
the development of WordNet Affect \citep{Strapparava2004} has been
prominently conducted for computational linguistics. It
has enriched the established resource of WordNet with emotion
categories through a semi-automatic procedure.  Taking a more
empirical perspective on data creation, the NRC Emotion Lexicon has
been crowdsourced, resulting in a more comprehensive dictionary
\citep{Mohammad2012b}.

For classification problems, in which pieces of texts are assigned to
one or many discrete emotion labels, lexicons are handy.  They provide
transparent access to the emotion of words, in order to analyze the
emotion of the texts that such words compose.  At the same time,
statistical approaches and deep learning methods can solve the task
without relying on dictionaries.  Models for emotion prediction are by
and large standard text classification approaches, either
feature-based methods with linear classifiers or transfer learning
methods based on pretrained transformers. Various shared tasks 
provide a good overview on the topic 
\cite{Strapparava2007,Klinger2018,Mohammad2018}.

A crucial requirement for these types of automatic systems is the
availability of appropriately sized and representative data: in
emotion analysis, models trained on one domain typically strongly
underperform in another \cite{Bostan2018}. Ready-to-use corpora
nowadays span many domains, including stories \cite{Ovesdotter2005},
news headlines \cite{Strapparava2007}, songs lyrics
\cite{Mihalcea2012}, tweets \cite{Mohammad2012}, conversations
\cite{Li2017,Poria2019}, and Reddit posts \cite{Demszky2020}. Many 
 resources limit their labels to the most frequent or most fitting
emotion categories in the respective domain. Only a handful uses
 more than the eight emotions proposed by Plutchik.
Exceptions are the corpora by \namecite{Abdul2017} and
\namecite{Demszky2020}, who built two large resources for emotion
detection, respectively containing tweets with all 24 emotions present
in Plutchik's wheel, and Reddit comments associated with 27 emotion
categories.  We refer the reader to \citet{Bostan2018} for a more
complete overview of emotion corpora.

\begin{table}
\small
  \caption{The representation of emotions along six appraisal dimensions
    according to \protect\namecite[Table 6]{Smith1985}.}
  \label{tab:smithellsworth}
  \begin{tabular}{lrrrrrr}
    \toprule
    Emotion & Unpleasant & Responsibility & Uncertainty & Attention & Effort & Control \\
    \cmidrule(r){1-1}\cmidrule(rl){2-2}\cmidrule(rl){3-3}\cmidrule(rl){4-4}\cmidrule(rl){5-5}\cmidrule(rl){6-6}\cmidrule(l){7-7}
    Happiness & $-$1.46 & 0.09 & $-$0.46 & 0.15 & $-$0.33 & $-$0.21 \\
    Sadness   & 0.87 & $-$0.36 & 0.00 & $-$0.21 & $-$0.14 & 1.15\\
    Anger     & 0.85 & $-$0.94 & $-$0.29 & 0.12 & 0.53 & $-$0.96  \\
    Boredom   & 0.34 & $-$0.19 & $-$0.35 & $-$1.27 & $-$1.19 & 0.12\\
    Challenge & $-$0.37 & 0.44 & $-$0.01 & 0.52 & 1.19 & $-$0.20\\
    Hope      & $-$0.50 & 0.15 & 0.46 & 0.31 & $-$0.18 & 0.35\\
    Fear      & 0.44 & $-$0.17 & 0.73 & 0.03 & 0.63 & 0.59\\
    Interest  & $-$1.05 & $-$0.13 &$-$0.07 &0.70 &$-$0.07 &$-$0.63\\
    Contempt  & 0.89 &$-$0.50 &$-$0.12 &0.08 &$-$0.07 &$-$0.63\\
    Disgust   & 0.38 &$-$0.50 &$-$0.39 &$-$0.96 &0.06 &$-$0.19\\
    Frustration& 0.88 &$-$0.37 &$-$0.08 &0.60 &0.48 &0.22\\
    Surprise  & $-$1.35 &$-$0.94 & 0.73 &0.40 &$-$0.66 &0.15\\
    Pride     & $-$1.25 &0.81 &$-$0.32 &0.02 &$-$0.31 &$-$0.46\\
    Shame     & 0.73 &1.31 &0.21 &$-$0.11 &0.07 &$-$0.07\\
    Guilt     & 0.60 &1.31 &$-$0.15 &$-$0.36 &0.00 &$-$0.29 \\
    \bottomrule
  \end{tabular}
\end{table}

\subsection{Evaluation and Appraisal}
The \textit{evaluative tradition} is instantiated by appraisal
theories of various kinds. At the core of this stream of thought lies
the idea that an emotion is to be described in terms of many
components. It is ``an episode of interrelated, synchronized changes
in the states of all or most of the five organismic subsystems in
response to the evaluation of a [...]  stimulus-event''
\cite{Scherer2005}. The five subsystems are cognitive,
neurophysiological and motivational components (respectively, an
appraisal, bodily symptoms and action tendencies), as well as motor
(facial and vocal) expressions, and subjective feelings (the perceived
emotional experience).  The change in appraisal, in particular,
consists in weighting a situation with respect to the significance it
holds: ``does the current event hamper my goals?'', ``can I predict
what will happen next?'', ``do I care about it?''.  The emotion that
one experiences depends on the result of such evaluations, and can be
thought of as \textit{being caused} or as \textit{being constituted}
by those evaluations (e.g., in \citet{Scherer2005} appraisals lead to
emotions, in \citet{Ellsworth1988} appraisals are themselves
emotions).

Criteria used by humans to assess a situation are in principle
countless, but there is a finite number that researchers in psychology
have come up with in relation to emotion-eliciting events. For
\namecite{Ellsworth1988}, they are six: \pleasantness (how pleasant an
event is, likely to be associated with joy, but not with disgust),
\effort (that an event can be expected to cause, high for anger and
fear), \textit{certainty} (of the experiencer is about what is
happening, low in the context of hope or surprise), \attention (the
degree of focus that is devoted to the event, low, e.g., with boredom
or disgust), \ownresponsibility (how much responsibility the
experiencer of the emotion holds for what has happened, high when
feeling challenged or proud), and \ownControl (how much control the
experiencer feels to have over the situation, low in the case of
anger).  \namecite{Ellsworth1988} found these dimensions to be
powerful enough to distinguish 15 emotion categories (as shown in
Table~\ref{tab:smithellsworth}). We follow their approach closely, but
regard a larger set of variables based on
\namecite{Smith1985,Scherer1997,Scherer2013}.

\begin{figure}
  \centering
  \includegraphics{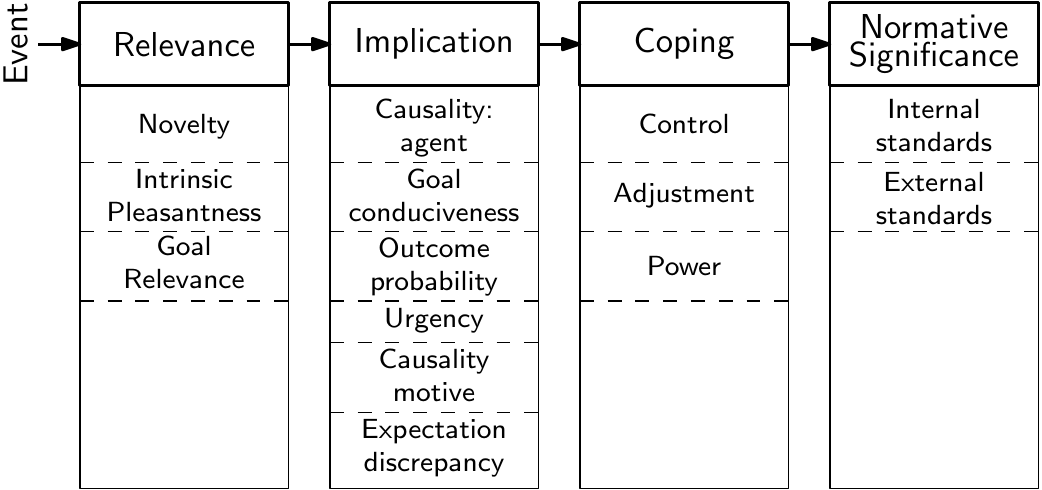}
\caption{Sequence of appraisal criteria adapted from \citet{sander2005systems} and \citet{Scherer2013}. High-level categories represent four appraisal objectives, with the item inside the dashed boxes corresponding to the relative checks.}
\label{fig:appraisal-sequence}
\end{figure}

\citet{Scherer2013} propose a more high-level and structured approach.
Figure~\ref{fig:appraisal-sequence} illustrates their appraisal module
as a multi-level sequential process, which comprises four appraisal
objectives that unfold orderly over time.  First, an event is
evaluated for the degree to which it affects the experiencer
(Relevance) and its consequences affect the experiencers' goals
(Implication).  Then, it is assessed in terms of how well the
experiencer can adjust to such consequences (Coping potential), and
how the event stands in relation to moral and ethical values
(Normative Significance).  Each objective is pursued with a series of
checks. For instance, organisms scan the Relevance of the environment
by checking its novelty, which in turn determines whether the stimulus
demands further examination; the Implication of the emotion stimulus
is estimated by attributing the event to an agent, by checking
if it facilitates the achievement of goals, by attempting to predict
what outcomes are most likely to occur; the Coping potential of the
self to adapt to such consequences is checked, e.g., by appraising who
is in control of the situation; as for the Normative Significance, an
event is evaluated against internal, personal values that deal with
self-concepts and self-esteem, as well as shared values in the social
and cultural environment to which the experiencer belongs.  Therefore,
similar to valence, arousal, and dominance, appraisals can be
interpreted as a dimensional model of emotions, namely, a model that
is based on people's interaction with the surrounding environment.

Despite concerning different objectives, all such checks possess an
underlying dimension of valence \cite{scherer2010blueprint}. That is,
one always represents the result of a check as positive or negative
for the organism: for intrinsic pleasantness, valence amounts to
a concept of pleasure, for goal relevance to an idea of
satisfaction, for coping potential, to a sense of power; it involves
self- or ethical worthiness in the case of internal and external
standards compatibility, and the perceived predictability for novelty
(with a positive valence being a balanced amount of novelty and
unpredictability -- otherwise a too sudden and unpredictable event
could be dangerous, while a too familiar one could be
boredom-inducing).  The outcome of the appraisal process is thus
dependent on subjective features such as personal values, motivational
states and contextual pressures \cite{scherer2010blueprint}. 
Two people with different
goals, cultures and beliefs might produce different
evaluations of the same stimulus.

\begin{figure}
  \centering
  \includegraphics[width=\linewidth]{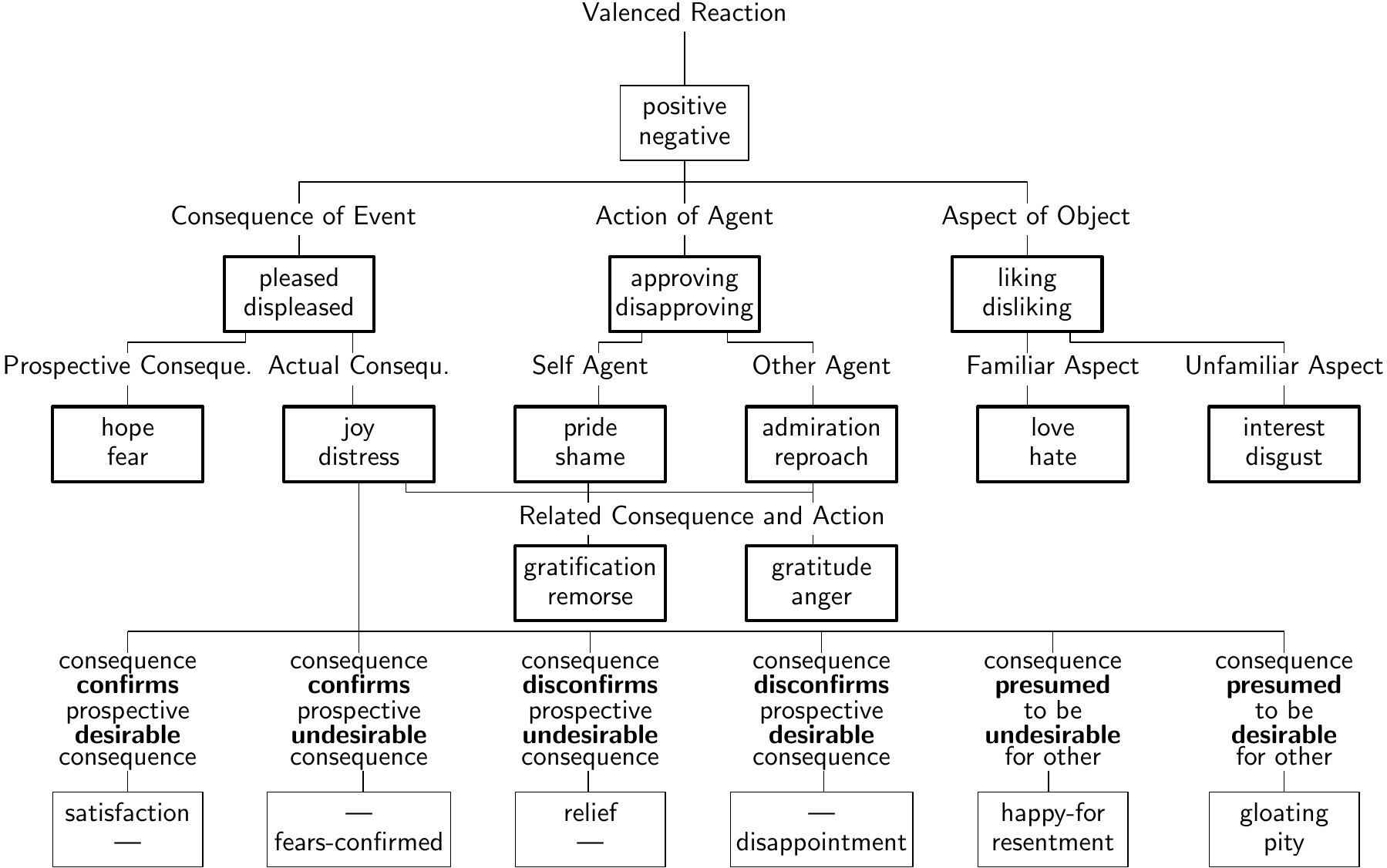}
  \caption{The OCC model, drawn after the depiction of \citet[Figure
    2]{Steunebrink2009}.}
  \label{fig:occ}
\end{figure}

Another model that falls in the evaluative tradition 
 is the OCC model (named after the authors Ortony, Clore and
Collins' initials), in which emotions emerge deterministically from
logic-like combinations of evaluations (e.g., \textit{if} a condition
holds, \textit{then} a specific valenced reaction follows). We
visualize the OCC in Figure~\ref{fig:occ}.  The model formalizes
the cognitive coordinates that rule more than 20 emotion phenomena (shown in the
figure in the bold boxes) within a hierarchy that develops according
to how specific components interact with one another: it starts with
three eliciting conditions, namely consequences of events, agents'
actions and aspects of objects, which spread out according to how they
are appraised with different mental representations (respectively,
goals, norms/standards and tastes/attitudes) based on some binary
criteria, like desirability-undesirability. A path in the hierarchy,
corresponding to a specific instantiation of such components, fires an
emotion (e.g., love stems from the liking of an object).  Like other
appraisal approaches, the OCC model can differentiate emotions
with respect to their situational meanings, but it sees emotions more as
a descriptive structure of prototypical situations than as a process
\cite{Clore2013}.

The rigorously logical view of OCC makes it attractive for
computational studies; in fact, this model is applied also in NLP.
Both \namecite{Shaikh2009} and \namecite{Udochukwu2015} propose rules
to measure some variables that come from the theory of
\namecite{Clore2013}: valence (hence desirability, compatibility with
goals and standards, and pleasantness) is represented with lexicons
that associate objects and events with positivity or negativity, a
confirmation status is associated to the tense of the text, and
causality is modeled with the help of semantic and dependency
parsing. These variables are combined with rules to infer an
emotion category for the text.

This logics-based combination of variables has an arguable
limitation. It treats appraisals in isolation, focusing solely on
those that have a textual realization; consequently, the
classification task is reduced to a deterministic decision that
disregards the probability distributions across all appraisal
variables.  This issue has been bypassed by \namecite{Hofmann2020},
which represents the first attempt to measure emotion-related
appraisals in the NLP panorama.  They annotate a corpus of event
descriptions with the dimensions of \namecite{Smith1985}, and on that,
they train classifiers that predict emotions and
appraisals. Processing the variables in a probabilistic manner, these
systems can handle texts with an opaque appraisal ``substrate'' better
than OCC-based models; they are also better suited for inferring
emotions from the underlying (predicted) appraisals.  However, since
it can count on a comparably small corpus, this work falls short in
showing if emotion analysis benefits from the use of appraisals.

Besides their promising application in classification tasks, appraisal
theories have additional significance for NLP.  The cognitive
component that is directly involved in the emergence of emotion
experiences actually plays a role also in humans' decoding of
emotions.  People's empathy and the ability to assume the affective
perspective of others is guided by their assessment of whether a
certain event might have been important, threatening, or convenient
for those who lived through it \cite{omdahl1995cognitive}.  Motivated
by this, \citet{Hofmann2021} analyze if readers find sufficient
information in text to judge appraisal dimensions, and compare the
agreement among annotators when they have access to the emotion of a
text (as disclosed by the texts' writers) to when they do not.  Their
results show that having knowledge about emotions boosts the
annotator's agreement on appraisals by a substantial amount.  In a
follow-up study \cite{Troiano2022}, we focus on experiencer-specific
appraisal and emotion modeling, thus combining semantic role labeling
with emotion classification.  We annotate the variables that we also
consider in the present paper (described in
Section~\ref{ssec:appraisal-definitions}), but with the help of
trained experts rather than via crowdsourcing and on a smaller scale.

In summary, the components of emotions discussed by appraisal theories
are relevant in this field at various levels, but related studies in
NLP have some pitfalls that are left unresolved. Notably, they use
limited sets of appraisals, fail to provide evidence that appraisals
can help emotion classification, and disregard how well the texts'
annotators can judge appraisals in the first place.  We address these
gaps by building a large corpus of texts annotated with a
broad set of appraisal dimensions, and by comparing the agreement that
other annotators achieve \textit{with the original emotion
experiencer} (i.e., the writers who produced the texts).

\section{Contextualization in  Emotion Annotation Reliability Research}
\label{sec:reliability}
\begin{figure}
  \centering
  \includegraphics[width=0.8\linewidth]{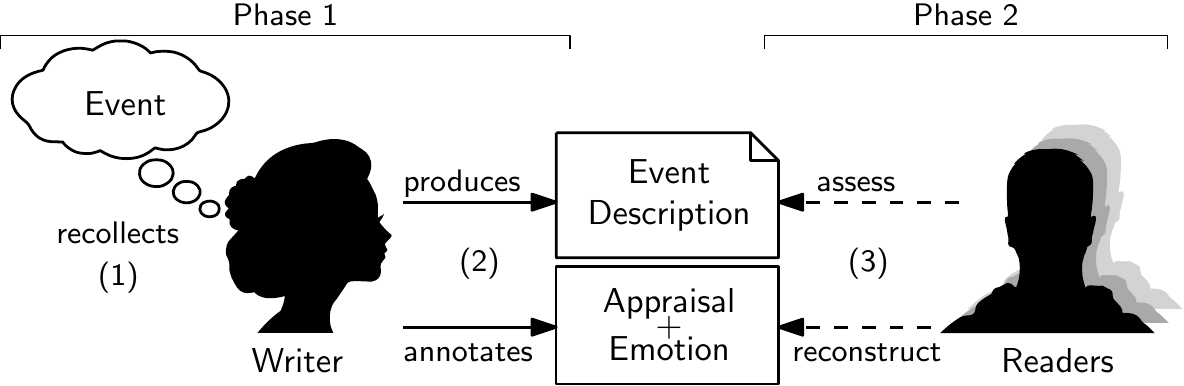}
  \caption{Overview of study design.}
  \label{fig:losinginformation}
\end{figure}
\subsection{Overview of Study Design}
In this paper, we build a novel resource to understand if appraisal
theories are suitable for emotion modeling, and how well computational
models can be expected to perform when interpreting textual event
descriptions.  We visualize our setup in
Figure~\ref{fig:losinginformation}, and discuss it in more detail in
Section~\ref{sec:creation}.  Crowdsourced writers are tasked to
remember an event that caused a particular emotion in them (1).  They
describe it and report their evaluation and subjective experience in
that circumstance (2), including their appraisals. By assessing that
description, other annotators (i.e., readers) attempt to reconstruct
both the original emotion and the appraisal of the event
experiencer~(3).

Like in other fields, corpus creation efforts in emotion analysis
follow the practice of comparing the judgments of multiple coders and
quantifying their agreement.  Typically this is done considering only 
the annotations of the readers, as those of the writers of texts are
often not available.  That way, it is possible to gain insights into
their reliability, but the correctness of their judgments (i.e., if
they agree with the writers) cannot be established -- a design choice
that in fields other than NLP has been shown to affect the
inter-annotator results drastically (see Section~\ref{sec:erapsych}).
Instead, we compare the annotations resulting from (2) with those
collected in (3).

In the following, we review related work in NLP and in psychology that
revolves around the emotion recognition reliability of humans, which
influences our data collection procedure.

\subsection{Emotion Recognition Reliability in Psychology}
\label{sec:erapsych}
The problem of recognizing emotions has concerned the developments of
emotion theories from early on.  In the book ``The Expression of the
Emotions in Man and Animals'' (\citeyear{Darwin1872}), Darwin focuses
on many external manifestations, namely facial expressions and
physiological reactions (e.g., muscle trembling, perspiration, change
of skin color), claimed to be discriminative signals that allow
understanding what others feel.  Such observations are deepened by
Paul Ekman, who introduces a coding scheme of facial muscle movements
to assess emotion expressions quantitatively
\citep{Ekman1980,Ekman1978}.

Ekman also studies quantitatively if emotions can be identified by
people who are not directly experiencing them.  Focusing on the
intercultural aspect of this ability, he asks if ``a particular facial
expression [signifies] the same emotion for all peoples'' \cite[p.\
207]{Ekman1972}. He recites a study in which the culture of emotion
judges did not show a significant impact on their
agreement \citep[p.\ 242f.]{Ekman1972}. In that study, Japanese and
American subjects were presented with depictions of facial
expressions, and they agreed on the recognized emotions with an
accuracy of .79 and .86, respectively.  These numbers measured the
quality of annotation from within the observers' groups. However, by
comparing the coders' decisions with the actual emotion felt by the
depicted subjects, accuracy dropped to .57 and .62 (with .50 being
chance).  Brief, quantifying agreement returns substantially different
results depending on whether it is measured among judges of the
emotion felt by others, or between the same judges and those
``others''.  This constitutes an important insight for our study: we
investigate agreements among external annotators, and compare their
judgments with the self-assessments of the first-hand emotion
experiencers.

The fact that emotions cannot be perfectly identified by interpreting
facial expressions motivates a myriad of studies after Ekman.
Actually, not all emotions are equally difficult to
recognize. \citet{Mancini2018} find that, at least among
pre-adolescents, happiness is easier identified than fear, and
further, that there is a relation between the recognition performance
and the emotion state of the person carrying it out.
Other factors also influence this task.  \citet{Doellinger2021} review
them, pointing to peer status and friendship quality \citep{Wang2019},
to the possible state of depression of the observers
\citep{Dalili2015}, and to their personality traits \citep{Hall2016}
-- conscientiousness and openness are positively correlated to the
ability to recognize nonverbal expressions of emotions, while shyness
and neuroticism are negatively associated with it \citep{Hall2016}.
We also assess personality traits and state-specific variables in our
study.

\subsection{Reliability of Emotion Annotation in Text}
Computational linguistics commonly deals with spontaneously-generated text. 
Domains that received substantial attention are news headlines
and articles, literature, everyday dialogues, and social media. The
field of emotion analysis focuses on these as well, particularly to
learn the tasks of emotion classification and intensity
regression. Depending on the domain in question, the emotion to be
classified is either the one expressed by the writer (e.g., in social
media) or one that the reader experiences (e.g., with poetry and
news).  In both cases, the standard approach to building emotion
corpora is, first, to have multiple people annotating its texts, and
second, to measure their agreement.

If the variables to be predicted/annotated are continuous, agreement
can be calculated with correlation or distance measures, despite not
being originally designed for inter-coder agreement. Examples are
Pearson's $r$ or Spearman's $\rho$, root mean square error ($\textrm{RMSE}$) and
mean absolute error. This holds for annotations taking place both on 
Likert scales (what we do in this paper) and via best-worst scaling
\citep{Louviere2015}.  Various measures have been formulated
specifically for the comparison of annotations with discrete
categories.  Cohen's $\kappa$, for instance, quantifies agreement
between two annotators, and Fleiss' $\kappa$ \citep{Cohen1960} is its
generalization to multiple coders. Cohen's $\kappa$ is defined as
$\kappa ={\frac {p_{o}-p_{e}}{1-p_{e}}}$, where $p_o$ is the observed
probability of agreement, and $p_e$ is the expected agreement based on
the distribution of labels assigned by the annotators individually. In
multi-class classification problems, it is common to calculate
$\kappa$ across all classes, while in multi-label problems, this is
done for each class separately.

With skewed label distributions, $\kappa$ might underestimate
agreement and assume low scores. For this reason, authors often report
other evaluations in addition. Typical options are a between-annotator
accuracy
($\textrm{acc}=\frac{\mathrm{TP}+\mathrm{TN}}{\mathrm{TP}+\mathrm{FP}+\mathrm{FN}+\mathrm{TN}}$),
where the decision of one annotator is considered a gold standard and
the other is treated as a prediction, and an inter-annotator agreement
F$_1=\frac{\mathrm{\mathrm{TP}}}{TP+\frac{1}{2}(\mathrm{FP}+\mathrm{FN})}$
(where TP is the count of true positives, FP of false positives, TN of
true negatives, and FN of false negatives).  Since classification
models are also evaluated with the latter two measures, their
performance can be directly compared to humans'.  This is valuable for
at least two reasons: first, one can treat inter-annotator agreement
as a reasonable upper bound for the models. For instance, if
annotators agree with one another or with the original emotion label
of text only to a certain extent, models showing analogous performance
are still acceptable. In fact, the purpose and plausibility of models
that achieve better results than humans is hard to interpret.  Second,
agreement can be leveraged to assess the quality of datasets.  For
instance, \citet{Mohammad2012} provide a large corpus of tweets
labeled with (emotion) hashtags. Such an approach can be
considered noisy, because a hashtag does not necessarily express the
emotion of the writer or of the text content.  Still, its creators
find that an emotion classifier reaches similar results on the
``self-labeled'' data as it does on manually labeled texts (40.1
F$_1$), suggesting that the quality of labels is comparable in the
human and the automatic settings.

Inter-annotator agreement scores vary based on the domain of focus.
\citet{Haider2020} find an average $\kappa$=.7 and F$_1$=.77 on poems
for the annotation of the perceived emotion.  \citet{Aman2007} report
a $\kappa$ between .6 and .79 for blogs, where joy shows the highest
agreement and surprise the lowest.  Similar numbers are obtained by
\citet{Li2017} on dialogues (.79, although the measure is
unspecified).  The $\kappa$ of annotators judging the tweets in
\citet{Schuff2017} ranged from .57, for trust, and .3, for disgust and
sadness. Looking at correlation measures, for news headlines,
\citet{Strapparava2007} compute an average emotion intensity
correlation between annotators of .54, with sadness having the highest
score (.68) and surprise the lowest (.36).  \citet{Preotiuc2016}, who
annotate Facebook posts, report correlations of .77 for valence and
.83 for arousal.

Previous work shows that the agreement between annotators in emotion
analysis is limited in comparison to other NLP tasks. In the domain of
fairy tales, \citet{Ovesdotter2005} find a $\kappa$ between .24 and
.51, depending on the annotation pair.  Building their corpus of news
headlines, \citet{Bostan2020} report an agreement of $\kappa$=.09,
likely due to the fact that headline interpretation can be
sensitive to one's context and background.  Another factor that
influences (dis)agreements is the annotation perspective that coders
are required to assume.  \citet{Buechel2017readers} compare judgments
about the readers' and the writers' emotion (where the latter is
inferred by the readers themselves and not indicated by the authors of
the texts), providing evidence that taking the perspective of writers
promotes the overall annotation quality.  In a similar vein,
\citet{Mohammad2018b} analyses the role of personal information on
VAD-based judgments, much in line with the multiple works in
psychology (introduced in Section~\ref{sec:erapsych}) which delve into
the annotators' personal information (e.g., mental disorders,
personality traits) in order to better understand their annotation
performance.  While creating a VAD lexicon, \citet{Mohammad2018b}
collects data about the annotators' age, gender, agreeableness,
conscientiousness, extraversion, neuroticism, and openness to
experience, and points out a significant relation between
(nearly all) the demographic/personality traits of people and their
task agreement.

Across such a broad literature, agreement between readers and writers
is mostly disregarded. The texts' authors are rarely leveraged as
annotators. In fact, corpora containing information about their
emotion are typically constructed via self-labeling, either with
hashtags \cite{Mohammad2012} and emojis \cite{Felbo2017} or through
emotion-loaded phrases that are looked for in the text
\cite{Klinger2018}.  The only work that we are aware of, and which
involves text writers, is that of \citet{Troiano2019}. They ask
crowdworkers to generate event descriptions based on a prompting
emotion, and then compare it to the emotion inferred by the readers
from text in terms of accuracy.  Their work is a blueprint for our
crowdsourcing setup, but it does not contain any appraisal-related
label. In a follow-up work, they assign appraisal dimensions to the
same descriptions with the help of three carefully trained annotators
\cite{Hofmann2020}, who achieve average $\kappa$=.31 for the variable
of \attention, .31 for \textit{certainty}, .32 for \effort, .89 for
\pleasantness, .63 for \ownresponsibility, .58 for \ownControl, .37
for \situationalControl, and .53 as an overall average. These numbers
are the only agreement scores for appraisal dimensions that are
available up-to-date (but they are only computed among readers).

\section{Corpus Creation}
\label{sec:creation}
To the best of our knowledge, there are no linguistic resources to
study affect-oriented appraisals. Therefore, as a starting point for
our investigation, we built an emotion and appraisal-based corpus of
event descriptions.  The creation of \corpusname took place over a
period of 8 months (from March to December 2021), and it was divided
into two consecutive phases: a first phase for generating the data and
a second one to validate it. These phases are both represented in
Figure~\ref{fig:losinginformation}. Phase 1 consists of generators
\textit{recollecting} personal events (Step (1)) and \textit{writing
  and annotating} them ((Step (2)); Phase 2 consists of validators
\textit{assessing} the events produced in Phase 1 and
\textit{reconstructing the emotion and the appraisals} (Step (3)).

The two phases were designed to mirror each other with respect to the
considered variables, the formulation of questions, and the possible
answers.  In the generation phase, participants produced event
descriptions and informed us about their appraisals and emotions. The
authors' appraisals and emotions were then reconstructed in the
validation phase by multiple readers for a subsample of texts.
In both, participants disclosed their emotional state at present,
their personality traits, and demographic information.  As a result,
part of \corpusname is annotated from two different
perspectives. One, corresponding to \textit{generation}, is based
on the recollection of evaluations as they were originally made when
the event happened; the other, the \textit{validation}, is about
inferred evaluations. In this paper, we refer to the authors/writers
of the event descriptions also as generators and to the readers as
validators (Phase 1 and Phase 2). Both are considered participants in
the study and act as text annotators. The full annotation
questionnaires, including the comparison between the generation and
the validation phases, is depicted in the Appendix,
Table~\ref{tab:questionnairetemplate}.\footnote{PDF-printouts of the
  questionnaires showing the original design are part of the
  supplementary material.}

Annotating well-established corpora with emotions and appraisals could
have been a viable alternative to generating texts from scratch, but
such a choice would have faced principled criticism.  Available
resources provide no ground truth appraisals, impeding to evaluate
if the readers' annotations are correct. This is a problem, because
judgments concerning emotions are highly subjective, and this is also
assumed to be the case for the cognitive evaluations of events -- they
hinge on people's world knowledge and on their perception of the stimulus
event, which is not necessarily shared between the texts' writers and
the annotators. \citet{Hofmann2020} and \citet{Hofmann2021}
have enriched an existing corpus of event descriptions with evaluative
dimensions, asking annotators to interpret how the texts' authors
assessed such events in real life (similar to our validation setup).
By operating in the absence of a ground truth annotation, they could
not determine if the evaluations were well reconstructed. This is the
gap that we fill in with \corpusname.

\subsection{Variable Definition}
\label{ssec:variable-definitions}
The formulation of a task concerning appraisal-related judgments
depends on the specific theory that one considers. As a matter of
fact, different research lines are rooted in a common conceptual
framework, but they are still characterized by internal
differences. For example, appraisal dimensions change from one work to
the other, or are qualified in different ways.  Below we establish the
theoretical outset of our questionnaire, describing how we defined the
variables of interest: appraisals
(Section~\ref{ssec:appraisal-definitions}), emotions
(Section~\ref{emotions-definitions}), and some supplementary variables
(Section~\ref{ssec:other-variables}).

\subsubsection{Appraisals}
\label{ssec:appraisal-definitions}

We adopt the schema proposed by \citet{sander2005systems},
\citet{scherer2010blueprint} and \citet{Scherer2013}.  They group
appraisals into the four categories shown in
Figure~\ref{fig:appraisal-sequence}, which represent specific
evaluation objectives. There is a first assessment aimed at weighing
the relevance of an event, followed by an estimate of its
consequences, and of the experiencer's own capability to cope with
them; last comes the assessment of the degree to which the event
diverges from personal and social values.

Each objective is instantiated by a certain number of evaluation checks, and each
check can be broken down into one or many appraisal
dimensions. Namely,
\begin{enumerate*}[start=1,before=\itshape,font=\normalfont]
\item suddenness,
\item familiarity,
\item predictability,
\item pleasantness,
\item unpleasantness,
\item goal-relatedness,
\item own responsibility,
\item others' responsibility,
\item situational responsibility,
\item goal support,
\item consequence anticipation,
\item urgency of response,
\item anticipated acceptance of consequences,
\item clash with one's standards and ideals,
\item violation of norms or laws.
\end{enumerate*}
These dimensions illustrate properties of events and their relation to
the event experiencers. Used by \citet{Scherer1997} to create the
corpus ISEAR\footnote{Original questionnaire:
  \url{https://www.unige.ch/cisa/files/3414/6658/8818/GAQ\_English\_0.pdf}.},
they constitute the majority of appraisals judged by the annotators in
our study as well.  Figure~\ref{fig:appraisal-sequence-ours}
collocates them (as numbered items) under the corresponding checks
(the underlined texts).

The above items can also be found in other studies.  For instance,
while formulating the questions differently, \citet{Smith1985} analyse
pleasantness, certainty, and responsibility (they merge
\textit{others'} and \textit{situational responsibility} together). In
addition, they directly tackle a handful of dimensions which are only
implicit in \citet{Scherer1997}, specifically
\begin{enumerate*}[start=16,before=\itshape,font=\normalfont]
\item attention, \normalfont{and}
\item attention removal,
\end{enumerate*}
two assessments that can be considered related to the relevance and
the novelty of an event, and
\begin{enumerate*}[start=18,before=\itshape,font=\normalfont]
\item effort,
\end{enumerate*}
which is the understanding that the event requires the exert of physical or
mental resources, and is therefore close to the assessment of
one's potential. \citet{Smith1985} also divide the check of control into the more
fine-grained dimensions of
\begin{enumerate*}[start=19,before=\itshape,font=\normalfont]
\item own control of the situation,
\item others' control of the situation,
\item chance control
\end{enumerate*}.

We integrate the two approaches of \citet{Scherer1997} and
\citet{Smith1985}, by adding the latter six criteria to our
questionnaire. We include \attention and \textit{attention removal}
under Novelty in Figure~\ref{fig:appraisal-sequence}, \effort as part
of the Adjustment check, and \textit{own, others'} and \textit{chance
 control} inside Control. This enables us to align with the NLP setup
described in \citet{Hofmann2020} and
\citet{Hofmann2021}\footnote{\citet{Hofmann2020} and
  \citet{Hofmann2021} use a subset of our dimensions but a different
  nomenclature. The following is the mapping between their variables
  and ours: attention $\rightarrow$ attention, responsibility
  $\rightarrow$ own responsibility, control $\rightarrow$ own control,
  circumstantial control $\rightarrow$ chance control, pleasantness
  $\rightarrow$pleasantness, effort $\rightarrow$ effort; certainty
  $\rightarrow$ consequence anticipation. Certainty (about what was
  going on during an event) and consequence anticipation are close
  but not identical concepts. After including the first in a pre-test
  of our study, we observed that its annotation was monotonous across
  both emotions and workers (an event about which people can produce a
  text is likely judged as one that they understood). We discard it.},
and to have a much larger coverage of dimensions motivated by
psychology.  Note however that we disregard a few dimensions from
\citet{Scherer1997}. In Figure~\ref{fig:appraisal-sequence-ours}
(adapted from \citet{Scherer2013}), they correspond to the checks
``Causality: motive'', ``Expectation discrepancy'' and ``Power''. As
they differ minimally from other appraisals, they would complicate the
task for the annotators.\footnote{While our crowdsourcing setup
  requires laypeople to accomplish the task with no previous training,
  no formal knowledge about appraisals, nor their relation to
  emotions, Scherer's (\citeyear{Scherer1997}) questionnaire was
  carried out in-lab.}

Research in psychology also proposes some best practices for
collecting appraisal data. \citet{yanchus2006development} in particular
casts doubt on the use of questions that annotators typically answer
to report their event evaluations (e.g., ``Did you think that the
event was pleasant?'', ``Was it sudden?''). Asking questions might
bias the respondents because it allows people to develop a theory
about their behavior in retrospect. Statements instead leave them free
to recall if the depicted behaviors applied or not (e.g., ``The event
was pleasant.'', ``It was sudden.'').  In accordance with this idea,
we reformulate the questions used in \citet{Scherer1997} and
\citet{Smith1985} as affirmations, aiming to preserve their meaning
and to make them accessible for crowdworkers.
Section~\ref{comparison-appendix} in the Appendix reports a comparison
between our appraisal statements and the original questions, as well
as the respective answer scales.

The resulting affirmations are detailed below.  In our study,
each of them has to be rated on a 1-to-5 scale, considering how much
it applies to the described event (1:``not at all'',
5:``extremely''). The concept names in parentheses are canonical names
for the variables that we use henceforth in this paper.

\begin{figure}
  \sf\small
  \begin{tabularx}{\linewidth}{|X|X|X|X|}\hline
              &             &        & Normative \\
    Relevance & Implication & Coping & Significance\\
    \hline
    \parbox[t][6cm]{\linewidth}{%
    \footnotesize
    \ul{Novelty}
    \begin{compactenum}
    \item suddenness
    \item familiarity
    \item predictability
    \end{compactenum}
    \begin{compactenum}[start=16]
    \item attention$^*$
    \item att. removal$^*$
    \end{compactenum}
    \vspace{\baselineskip}
    \ul{Intrinsic Pleasantness}
    \begin{compactenum}[start=4]
    \item pleasant
    \item unpleasant
    \end{compactenum}
    \vspace{\baselineskip}
    \ul{Goal Relevance}
    \begin{compactenum}[start=6]
    \item goal-related
    \end{compactenum}
    }
              &
    \parbox[t][6.5cm]{\linewidth}{%
    \footnotesize
    \ul{Causality: agent}
    \begin{compactenum}[start=7]
    \item own responsibility
    \item other's respons.
    \item situational respons.
    \end{compactenum}
   \vspace{\baselineskip}
   \ul{Goal conduciveness}
   \begin{compactenum}[start=10]
   \item goal support
   \end{compactenum}
   \vspace{\baselineskip}
   \ul{Outcome probability}
   \begin{compactenum}[start=11]
   \item consequence anticipation
   \end{compactenum}
   \vspace{\baselineskip}
   \ul{Urgency}
   \begin{compactenum}[start=12]
   \item response urgency
   \end{compactenum}
   \vspace{\baselineskip}
   (\ul{Causality: motive})\par
   (\ul{Expectation discrepancy})
   }

&
   \parbox[t][6cm]{\linewidth}{%
   \footnotesize
   \ul{Control}
   \begin{compactenum}[start=19]
   \item own control$^*$
   \item others' control$^*$
   \item chance control$^*$
   \end{compactenum}
   \vspace{\baselineskip}
   \ul{Adjustment}
   \begin{compactenum}[start=13]
   \item anticipated\par acceptance
   \end{compactenum}
   \begin{compactenum}[start=18]
   \item effort$^*$
   \end{compactenum}
  \vspace{\baselineskip}
  (\ul{Power})
  }
         &
   \parbox[t][6cm]{\linewidth}{%
   \footnotesize
   \ul{Internal standards}\par \ul{compatibility}
   \begin{compactenum}[start=14]
   \item clash with own standards/ideals
   \end{compactenum}
    \vspace{\baselineskip}
    \ul{External standards}\par \ul{compatibility}
    \begin{compactenum}[start=15]
    \item clash with laws/norms
    \end{compactenum}
    \vspace{\baselineskip}
   }
    \\
    \hline
  \end{tabularx}
  \caption{Appraisal objectives (on top) with their
    relative checks (underlined) and the appraisal dimensions
    investigated in our work (numbered). Checks in parenthesis have been proposed by
    \citet{Scherer1997} but are not included in our study. Items
    marked with an asterisk come from \citet{Smith1985}.}
  \label{fig:appraisal-sequence-ours}
\end{figure}

\paragraph{Novelty Check} According to \citet{Smith1985}, a key facet
of emotions is that they arise in an environment that requires a
certain level of attention.  Kin to the assessment of novelty, the
evaluation of whether a stimulus is worth attending or worth ignoring
can be considered the onset of the appraisal process. Their study
treats attention as a bipolar dimension, which goes from a strong
motivation to ignore the stimulus to devoting it full
attention. Similarly, we ask:
\begin{enumerate}[start=16]
\item I had to pay attention to the situation. (\attention)
\item I tried to shut the situation out of my mind. (\notconsider)
\end{enumerate}

Stimuli that occur abruptly involve sensory-motor processing other
than attention.  To account for this, the check of novelty develops
along the dimensions of suddenness, familiarity and event
predictability, respectively formulated as:
\begin{enumerate}[start=1]
\item The event was sudden or abrupt. (\suddenness)
\item The event was familiar. (\familiarity)
\item I could have predicted the occurrence of the event. (\eventpredictability)
\end{enumerate}

\paragraph{Intrinsic Pleasantness}  An
emotion is an experience that feels good/bad \cite{Clore2013}. This 
feature is unrelated to the current state of the
experiencer but is intrinsic to the eliciting condition (i.e., it bears
pleasure or pain):
\begin{enumerate}[start=4]
\item The event was pleasant. (\pleasantness)
\item The event was unpleasant. (\unpleasantness)
\end{enumerate}

\paragraph{Goal relevance check} As opposed to intrinsic pleasantness,
this check involves a representation of the experience for the goals
and the well-being of the organism (e.g., one could assess an event as
threatening). We define goal relevance as:
\begin{enumerate}[start=6]
\item I expected the event to have important consequences for me. (\goal)
\end{enumerate}

\paragraph{Causal attribution} Tracing a situation back to the cause
that initiated it can be key to understanding its significance. The
check of causal attribution is dedicated to spotting the agent
responsible for triggering an event, be it a person or an external
factor (one does not exclude the other):
\begin{enumerate}[start=7]
\item The event was caused by my own behavior. (\ownresponsibility)
\item The event was caused by somebody else’s behavior. (\otherResp)
\item The event was caused by chance, special circumstances, or
  natural forces. (\situationalResp)
\end{enumerate}
\citet{Scherer2013} also include a dimension related to the causal
attribution of motives (``Causality: motive'' in
Figure~\ref{fig:appraisal-sequence-ours}), which is similar to the
current one but involves intentionality. We leave intentions
underspecified, such that for 7., 8. and 9., the agents'
responsibility does not necessarily imply that they purposefully
triggered the event.

\paragraph{Goal conduciveness check}
The check of goal conduciveness is dedicated to assessing whether the
event will contribute to the organism's well-being:
\begin{enumerate}[start=10]
\item I expected positive consequences for me. (\goalSupport)
\end{enumerate}
Goal relevance (6.) differs from this appraisal: an event might be
relevant to one's goals and needs while not being compatible with them
(it might actually be deemed important precisely because it hampers
them).

\paragraph{Outcome probability check} 
Events can be distinguished based on whether their outcome can be
predicted with certainty.  For instance, the loss of a dear person
certainly implies a future absence, while taking a written exam could
develop in different ways.  Annotators recollected whether they could
establish the consequences of the event, at the moment in which it
happened, by reading:
\begin{enumerate}[start=11]
\item I anticipated the consequences of the event. (\anticipationConseq)
\end{enumerate}
\citet{Scherer2013} identify one more check about
consequences: people picture the potential outcome of an event based
on their prior experiences, and then evaluate if the actual
outcome fits what they expected. We refrain from introducing
\textit{expectation discrepancy} (under ``Implication'', in
Figure~\ref{fig:appraisal-sequence-ours}) in our repertoire. For one,
it is hard to distinguish from \textit{outcome probability check} in a
crowdsourcing setting; but mainly, such a dimension clashes with our attempt to 
induce the mental evocation of their state \textit{at the time in which the event happened}
(e.g., when taking an emotion-eliciting exam), and not when its consecutive developments
became known (e.g., when learning, later, if they passed). Brief, 
11. aims at understanding if people could
picture potential outcomes of the event, 
and not if their prediction turned out correct.

\paragraph{Urgency check} One feature of events is how urgently they
require a response. This depends on the extent to which they affect
the organism. High priority goals compel immediate adaptive
actions:

\begin{enumerate}[start=12]
\item The event required an immediate response. (\urgent)
\end{enumerate}

\paragraph{Control check} This group of evaluations concerns the
ability of an agent to deal with an event, specifically to influence
its development. At times, ``event control'' is in the hands of the
experiencer (irrespective of whether they are also responsible for
initiating it), other times it is held by external entities, and yet others
the event is dominated by factors like chance or natural forces
\cite{Smith1985}. Accordingly, we formulate the following three
statements:

\begin{enumerate}[start=19]
\item I was able to influence what was going on during the event. (\ownControl)
\item Someone other than me was influencing what was going on. (\othercontrol)
\item The situation was the result of outside influences of which
  nobody had control. (\textit{chance} or \situationalControl)
\end{enumerate}

We do not focus on ``Power'' (under Coping in 
Figure~\ref{fig:appraisal-sequence-ours}), the assessment
of whether agents can control the event at least in principle 
(e.g., if they possess the physical or intellectual resources to 
influence the situation).

\paragraph{Adjustment}
Related to control is the evaluation of how well an experiencer will 
cope with the foreseen consequences of the event, particularly with those
that cannot be changed:

\begin{enumerate}[start=13]
\item I anticipated that I would easily live with the unavoidable
  consequences of the event. (\acceptConseq)
\end{enumerate}
A different dimension of adjustment check is motivated by
\citet{Smith1985}.  Emotions can be differentiated on the basis of
their physiological implications, similar to the notion of arousal in
the dimensional models of emotion.  More precisely, subjects
anticipate if and how they will expend any effort in response to an
event (e.g., fight or flight, do nothing).  We phrase this idea as:
\begin{enumerate}[start=18]
\item The situation required me a great deal of energy to deal with
  it. (\effort)
\end{enumerate}

\paragraph{Internal and External standards compatibility}
The significance of an event can be weighted with respect to
one's personal ideals and to social codes of conduct.
Two appraisals can be defined on the matter:
\begin{enumerate}[start=14]
\item The event clashed with my standards and ideals. (\internalStandards)
\item The actions that produced the event violated laws or socially
  accepted norms. (\externSocialStandards)
\end{enumerate}
The first pertains to an event colliding with desirable attributes for
the self, with one's imperative motives of righteous behavior. The
second concerns its evaluation against the values shared in a social
organization. Both guide how experiencers react to events.

\subsubsection{Emotion Selection}
\label{emotions-definitions}
Our choice of emotion categories is closely related to that of
appraisals, because different emotions are marked by different
appraisal combinations. In the literature, such a relationship is
addressed only for specific emotions. Therefore, we motivate the
selection of this variable following appraisal scholars once more.

We consider the emotions that one or several studies claim to be
associated to the appraisals of
Section~\ref{ssec:appraisal-definitions}.  We include all emotions
from \citet{Scherer1997} as a first nucleus. They are \textit{anger},
\textit{disgust}, \textit{fear}, \textit{guilt}, \textit{joy}, and
\textit{sadness} (i.e., Ekman's basic set), plus \textit{shame}.  On
top of these, we use \textit{pride}, which is tackled with respect to
the objectives of \textit{relevance}, \textit{implication},
\textit{coping} and \textit{normative significance}
\citep{Manstead1989,Roseman1996,Roseman2001a,Smith1985,Scherer2001a}.
The last two works also comprise a discussion of \textit{boredom}, and
\citet{Roseman1990} and \citet{roseman1984cognitive} examine
\textit{surprise}, as well as the positive emotion of \textit{relief}.
\textit{Trust}, an emotion present in Plutchik's wheel, is linked to
the appraisal of goal support \cite{lewis2001personal}, and to the
check of control \cite{dunn2005feeling}.

We regard the reference to appraisal theories as a sufficiently strong
motivation to make use of these discrete emotions. It enables us, for
instance, to verify if the patterns of appraisals found in our data
correspond to those proposed by theorists, as a signal that the
annotators' understanding of the variables under consideration match
the experts'. Moreover, compared to dimensional models of affect,
discrete categories facilitate our attempt to explain how the
annotators' emotion judgments vary as their appraisal ratings
vary. Lastly, VAD concepts can be deemed implicit to the chosen
appraisals dimensions (e.g.,
valence~$\approx$~\pleasantness$-$\unpleasantness,
arousal~$\approx$~\attention~$-$~\notconsider,
dominance~$\approx$~\ownControl). In this sense, opting for a VAD
annotation would be redundant.

 We define our questionnaires with these 12 emotion labels. We add
in addition a \textit{no-emotion} category, because events can be
appraised along our 21 dimensions even if they elicit no emotion.  The
neutral class serves as a control group to observe
differences in appraisal between emotion- and non-emotion-inducing
events. However, not all texts generated for this label in \corpusname 
describe uninfluential or unemotional events. As pointed out later, 
many of them depict rather dramatic circumstances that, perhaps exceptionally,
did not stir the experiencers up.

\subsubsection{Other Variables}
\label{ssec:other-variables}
We use two other groups of variables regarding the described
emotion-inducing circumstances and the type of personas providing the
judgments.  The first group deals with emotion and event properties;
the other focuses on features of the study participants. Note that we
do not aim at analyzing all these variables in the current paper --
they potentially serve future studies based on our data.

\paragraph{Properties relative to Emotions and Events}
It is reasonable to assume that the same event is appraised
differently depending on its specific instantiation.  For example,
while standing in a queue, an emoter of boredom could feel more in
control of the situation than another, depending on how long each of
them persists in it, or how intensely the event affects them. Motivated
by this, we consider the \textit{duration of the event}, the
\textit{duration of the emotion} (with the possible answers
``seconds'', ``minutes'', ``hours'', ``days'', and
``weeks''\footnote{For the study of neutral events, the emotion
  duration variable comprises the option ``I had none''.}), and the
\textit{intensity} of the experience (to be rated on a 1 to 5 scale,
ranging from ``not at all'' to ``extremely'').

\paragraph{Properties of Annotators}
Annotation endeavours in emotion analysis show comparably low
inter-coder agreements, as discussed in
Section~\ref{sec:reliability}. We hence collect some properties of the
annotators, in order to understand how they influence (dis)agreements
among emotion and appraisal judgements.

One property concerns demographic information.  The self-perceived
belonging to a socio-cultural group can determine one's associations
to specific events. For that, we request participants to disclose
their \textit{gender} (``male'', ``female'', ``gender variant/non
conforming'', and ``prefer not to answer'') and \textit{ethnicity}
(either ``Australian/New Zealander'', ``North Asian'', ``South
Asian'', ``East Asian'', ``Middle Eastern'', ``European'',
``African'', ``North American'', ``South American'',
``Hispanic/Latino'', ``Indigenous'', ``prefer not to answer'', or
``other'').  We further ask them about their \textit{age} (as an
integer), as well as their \textit{highest level of education} (among
``secondary education'', ``high school'', ``undergraduate degree'',
``graduate degree'', ``doctorate degree'', ``no formal
qualifications'' and ``not applicable''), which might affect the
clarity of the texts they write, or the way in which they interpret
what they read.

People's personality traits are another attribute that guides their
judgments about mental states. We follow the Big-Five personality
measure of \citet{gosling2003very}.  As an alternative to lengthy
rating instruments, it is a 10-item measure corresponding to the
dimensions of
\textit{openness to experience} (measured positively via ``open to new
experiences and complex'' and negatively via ``conventional and
uncreative''),
\textit{conscientiousness} (measured positively via ``dependable and
self-disciplined'' and negatively via ``disorganized and careless''),
\textit{extraversion} (measured positively via ``extraverted and
enthusiastic'' and negatively via ``reserved and quiet'',
\textit{agreeableness} (measured positively via ``sympathetic and
warm'' and negatively via ``critical and quarrelsome''), and
\textit{emotional stability} (measured positively via ``calm and
emotionally stable'' and negatively via ``anxious and easily upset'').
Participants self-assign traits by rating each pair of adjectives on
a 7-point scale, from ``disagree strongly'' to ``agree strongly''.

As an extra link between the annotator and the annotation, we ask
participants what \textit{emotion they feel} right before entering the
task on a 1--5 scale (i.e., ``not at all'', ``intensely'').  For that,
the labels presented in Section~\ref{emotions-definitions} need to be
scored, except for the neutral label.  Further, we demand that they judge
the \textit{reliability} of their own answers.  This variable is
instantiated in different ways for the two phases.  Since writers can
recall events that happened at any point in their life, some memories
of appraisals might be more vivid than others, which can affect their
annotations.  Therefore, we deem confidence as the trustworthiness of
this episodic memory, quantifying people's belief that what they
recall corresponds to what actually happened. In the validation phase,
this variable measures the annotators' confidence that the emotion
they inferred from text is correct. Both are assessed on a 5-point
scale, with 1 corresponding to the lowest degree of confidence.

Lastly, we notice that the goal of building and validating a corpus of
self-reports potentially suffers from a major flaw. On the one hand,
there is no guarantee that the described events happened in the
writers' life. It is reasonable to think that, running out of ideas,
writers resorted to events that are typically emotional.  On the
other, readers' judgments might depend on whether they had an
experience comparable to the descriptions that they are presented
with. Therefore, we ask the writers if they actually experienced the
event they described, and the validators if they experienced a similar
event before. We cannot assess the
honesty of this answer either, but assuming it can be trusted, it
represents an additional level of information to look at patterns of
appraisals (e.g., how well the appraisal of events that were not
really lived in first person can be reconstructed).

\subsection{Generation}
\label{ssec:data-generation}
In the generation phase, annotators had the goal to describe an event
that made them feel one predefined emotion (out of those in
Section~\ref{emotions-definitions}) and to label such description. We
collected their answers using Google Forms.  Participants were
recruited on Prolific\footnote{\url{https://www.prolific.co}}, a platform
that allows prescreening workers based on several features (e.g.,
language, nationality).

We adopted a few strategies to promote data quality.  First, we opened
the study only to participants whose first language is English, with a
nationality from the US, UK, Australia, New Zealand, Canada, or
Ireland, and with an acceptance rate of $\geq$80\% to previous
Prolific jobs.  Second, we interspersed our questionnaires with two
types of attention tests: a strict test, in which a specified box on a
scale had to be selected, and one in which a given word had to be
typed.  Third, we intervened to make automatic text corrections
unlikely, by impeding the completion of our surveys via smartphones.

As we sought to have the same number of descriptions for all emotions,
we organized data generation into 9 consecutive rounds. A round was
aimed at collecting a certain number of tasks, based on different
emotions.  The first round served to verify whether our variables were
understandable, record the feedback of the annotators, and adjust the
questionnaire accordingly. We do not include it in \corpusname.  The
three final rounds balanced out the data. They comprised
questionnaires only for those emotions with insufficient data points,
due to rejections in the previous rounds. A special treatment was
reserved to \textit{shame} and \textit{guilt}: we considered them as
two sides of the same coin, and for each we collected half the items
than for the other emotions, motivated by the affinities between the
two \cite{tracy2006appraisal} and the difficulty for crowdworkers to
discern them \cite{Troiano2019}.

Annotators could fill in more than one questionnaire (for more than
one emotion, in more than one round). On average, people took our
study 2.8 times, with the most productive worker contributing with 33
questionnaires.  Since our expected completion time for a
questionnaire was of around 4 minutes, we set the payment to \textsterling\,0.50,
i.e., \textsterling\,7.50 per hour, in the respect of the minimum Prolific wage --
more details in the Appendix (Section~\ref{study-details}, 
Table~\ref{tab:generation-overview}).  The 6600 approved questionnaires
were submitted by 2379 different people, for a total cost of \textsterling\,4825.20
(including service fees, VAT, and the pre-test round).  We used these
answers to compile \corpusname.

\begin{figure}
\centering
  \tikzstyle{arrow} = [thick,->,>=stealth, minimum width=50mm]
  \tikzstyle{box} = [rectangle, draw=black, fill=white, xshift=15mm, minimum width=20mm, minimum height=10mm]
  \sf
  \begin{tikzpicture}[node distance=15mm]
      \node (current) [box] {Current State};
      \node (picture) [box, right of=current,xshift=5mm] {Picture the Event};
      \node (appraisal) [box, right of=picture,xshift=5mm] {Appraisal};
      \node (personal) [box, right of=appraisal] {Personal};
      \draw [arrow] (current) -- (picture);
      \draw [arrow] (picture) -- (appraisal);
      \draw [arrow] (appraisal) -- (personal);
    \end{tikzpicture}
    \caption{Questionnaire overview. The two phases of data creation
      mainly differ with respect to the block ``Picture the Event'':
      in the generation phase, the event is recalled and described; in
      the successive phase, the text is read for the validators to put
      themselves in the shoes of the writers.}
\label{fig:sketch-questionnaire}
\end{figure}
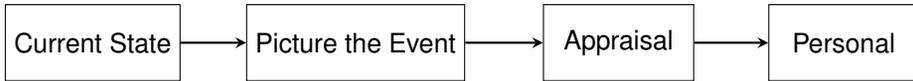

While each questionnaire was dedicated to a different prompting
emotion $E$, all of them instantiated the same template. As shown in
Figure~\ref{fig:sketch-questionnaire}, there are four blocks of
information.  At the very beginning, participants were asked about
their current emotion state. They then addressed the task of
recalling a real-life event in which they felt emotion $E$, indicating
the duration of the event, the duration of the emotion, the intensity
of the experience, and their confidence.  They described such
experience by completing the sentence ``\textit{I felt $E$
  when/because\ldots''}. For instance, people saw the text ``\textit{I
  felt anger when/because\ldots}'' for the prompting emotion
$E$=anger, and ``\textit{I felt no particular emotion
  when/because\ldots}'' in the \textit{no-emotion}-related
questionnaire.  We encouraged them to write about any event of their
choice, and to recount a different event each time they took our
survey, in case they participated multiple times.  As complementary
material, workers were provided with a list of generic life areas
(i.e., health, career, finances, community, fun/leisure, sports, arts,
personal relationships, travel, education, shopping, learning, food,
nature, hobbies, work) that could help them pick an event from their
past, in case they found such choice troublesome.  Moving on to the
third block of information, people rated the 21 appraisal dimensions,
considering the degree to which each of them held for the described
event.  The survey concluded with a group of questions on demographic
information, personality traits and event knowledge.\footnote{Event
  knowledge was included from round 5 afterwards.} People who
participated multiple times needed to provide their
demographics and personality-related data only once.

After the first three rounds, we observed that a substantial number of
participants had mentioned similar experiences. For instance,
\textit{sadness} triggered many descriptions of loss or illness, and
\textit{joy} tended to prompt texts about births or successfully
passed exams.  The risk we incurred was to collect over-repetitive
appraisal combinations.  To solve the issue, we aimed at inducing
higher data diversity.  Starting from round 4, we re-shaped the text
production task with two contrasting approaches.  One served to
stimulate the recalling of idiosyncratic facts. In the questionnaires
based on this solution, people were invited to talk about an
experience that was special to them -- one that other participants
unlikely had in their life. The other strategy attempted to refrain
them from talking about specific events.  We manually inspected the
collected texts, and compiled a repertoire of recurring topics,
emotion by emotion (see Table~\ref{tab:off-limits}); hence, we
presented the new participants with the topics usually prompted by
$E$, and we asked them to write about something different.  Since this
strategy appeared to diversify the data more than the other, we kept
using it in the last three rounds, updating the list of off-limits
topics.

\begin{table}
\centering
  \small
\caption{Fully-updated list of off-limits topics used to induce event
  variability.}
\label{tab:off-limits}
\begin{tabular}{p{1.8cm}p{10.5cm}}
\toprule
Emotion & Off-limits topics\\
\cmidrule(r){1-1}\cmidrule(r){2-2}
Anger & reckless driving, breaking up, being cheated on, dealing with abuses and racism\\
Boredom, No Emo. & attending courses/lectures, working, having nothing to do, standing in cues/waiting, shopping, cooking/eating \\
Disgust & vomit, defecation, rotten food, experiencing/seeing abusive behaviors, cheating\\
Fear & being home/walking alone (or followed by strangers), being involved in accidents, loosing sights of own kids/animals, being informed about an illness, getting on a plane\\
Guilt, Shame & stealing, lying, getting drunk, overeating, and cheating\\
Joy, Pride, Relief & birth events, passing tests, being accepted at school/for a job, receiving a promotion, graduating, being proposed to, winning awards, team winning matches\\
Sadness & death and illness, loosing a job, not passing an exam, being cheated on\\
Surprise & surprise parties, passing exams, getting to know someone is pregnant, getting unexpected presents, being proposed to\\
Trust & being told/telling secrets, opening up about mental health\\
\bottomrule
\end{tabular}
\end{table}

We acknowledge the artificiality of this setup: the texts were
produced by filling in a partial sentence and being tasked to recall
certain events but not others.  At the same time, constraining
linguistic spontaneity resulted in high-quality data: compared to a
free text approach, the sentence completion framework represented a
way to reduce the need for writers to mention emotion names -- which
we would need to remove for the validation phase -- and to minimize
the occurrence of ungrammaticalities.  Moreover, the descriptions
present constructs that are similar to productions occurring on
digital communication channels (e.g., those that can be found in the
corpus by \citet{Klinger2018}).

Having concluded the nine rounds, we compiled the generation side of
\corpusname. We discarded submissions with heavily ungrammatical
descriptions and incorrect test checks (i.e., those based on box
ticks, while we were lenient with type-in checks containing
misspellings). For individual annotators who completed various
questionnaires, we removed descriptions paraphrasing the same event,
and for those who filled the last block of questions more than once,
we averaged the personality traits scores.  In total, we obtained 6600
event descriptions, balanced by emotion: 275 descriptions for
\textit{guilt} and \textit{shame}, and 550 for all other prompting
emotions.

\subsection{Validation}
 During the second phase of building \corpusname, the texts 
 previously produced were annotated from the perspective of 
 the readers.  This was a ``validation'' process
in the sense that the resulting judgments can shed light on the 
inter-subjective validity of emotions and
appraisals. We are here in
line with the study by \citet{Hofmann2020} and \citet{Hofmann2021},
with the difference that we move to a crowdsourcing setup,
with non-binary judgments and a larger number of annotators,
texts, appraisals, and emotions.

The validation was developed in multiple rounds, preceded by a
pre-test that verified the feasibility of the study on a small number
of texts. The initial attempt was completed successfully and the
results were included in \corpusname.  This motivated us to proceed
using the same questionnaire (without any refinement).  Five
additional rounds were launched, until the target number of
annotations was achieved.

We validated only a subset of \corpusname, sampled with heuristic- and
random-based criteria: the data was balanced by emotion (100 per
label, except for guilt and shame, each of which received half the
items), and it was extracted from the answers of different generators
to boost the linguistic variability shown to the annotators --
assuming that personal writing styles emerged from the
descriptions. From a set of generation answers that respected these
conditions, we randomly extracted 1200 texts. Of those, 20 constituted
the material for the pre-test.  In each text, we replaced words that
correspond to an emotion name with three dots (e.g., ``\textit{I
  felt\ldots when I passed the exam}''), for the emotion
reconstruction task to be non-trivial. This preprocessing step was
accomplished through rules and heuristics. The first served to mask,
e.g., all words in an $E$-related text with the same lemma as $E$, or
synonyms of $E$ (e.g., the word ``furious'' in texts prompted by
\textit{anger}); the other to remove emotion words that contained
typos.\footnote{The full list of masked words and phrases is in the
  supplementary material.}

Answers were collected with the
software SoSciSurvey\footnote{\url{https://www.soscisurvey.de}}
which provides the possibility to create a questionnaire dynamically,
with different annotation data for each participant.
Specifically, each annotator judged 5 different texts placed in a
questionnaire, and each text was annotated by 5
different people,  for a total of 6000 collected judgments (i.e., 1200
texts$\times$5 annotations). Moreover, to prevent texts from being
re-annotated by their writers, the study was made inaccessible for 
all those who performed generation.

Participants were enlisted via Prolific, where we adopted the same
filtering and quality checking strategies used before.  Workers could
take our study only once, such that the judgments of each of them
would appear an equal number of times and would return a picture of
the crowd's impressions appropriate to study inter-subjectivity.  We
encouraged them to follow the instructions with a bonus of
\textsterling\,\,5 for the 5\% best performing respondents (i.e., 60
crowdworkers whose appraisal reconstruction is the closest to the
original ones).  We estimated the completion time of a questionnaire
to 8 minutes, and set the reward to \textsterling\,1 per
participant.\footnote{Breakdown of costs and number of participants in
 the Appendix, Table~\ref{tab:validation-overview}.}  As we approved
1217 submissions, constructing the validation side of \corpusname
costed \textsterling\,2188.09 (VAT, service fees and bonus included).

The validation questionnaire followed the one for generation.  We made
a few adjustments (a full a comparison of the questionnaires in the
two phases is in the Appendix, Table~\ref{tab:questionnairetemplate}),
but its template corresponded to that depicted in
Figure~\ref{fig:sketch-questionnaire}, with most answering options
mirroring those used before. Each questionnaire in the validation was
not dedicated to one predefined prompting emotion. It included 5 texts
that could be related to any of the emotions included in the
generation phase.

The block of questions opening the survey asked people to rate their
current emotion. Next, annotators were presented with a description
and they were asked to put themselves in the shoes of the writers at
the moment in which they experienced the event. They had to attempt
and infer the emotion that the writer was elicited by the event,
corresponding to $E$.  Our choice to work in a mono-label setting was
influenced by our compliance with the framework of
\citet{Scherer1997}.  Although their ISEAR corpus only contains
writers' annotations, the validation step we added instantiates an
opposite but corresponding task (i.e., emotion decoding). Thus, we put
the readers in the position to provide their predominant impression
about $E$, as were the participants in the previous (emotion encoding)
phase. The alternative of picking multiple emotion alternatives for a
text might have changed the annotation of the related appraisals,
making \corpusname and previous studies on the emotion--appraisal
relationship incomparable.

The validators also had to estimate the duration of the described
event and the duration of the emotion, as well as the intensity of
such experience. They rated their confidence in the annotations given
up to that point (i.e., how well they believed to have assessed
emotion, event duration, emotion duration, and intensity). As for the
variable of event knowledge, we asked workers if they had ever had an
experience comparable to the one they judged.  After that, they
reconstructed the original appraisals of the writers.  Participants
repeated these steps (included in \textit{Picture the event} and
\textit{Appraisal} in Figure~\ref{fig:sketch-questionnaire})
consecutively for the 5 texts.  Lastly, they provided personal
information related to their age, gender, education, ethnicity and
personality traits, as detailed in Section~\ref{ssec:other-variables}.

Overall, the answers we collected surpassed our target number of
answers -- i.e., some texts were annotated more than 5 times.  We
randomly removed some of these accepted submissions, and we achieved
this way the same amount of judgments per emotion, that we included 
in \corpusname.

\section{Corpus Analysis}
\label{sec:analysis}

In this section, we answer RQ1 (can humans predict
appraisals from text?) and RQ3 (do annotators' properties play a role
in their agreement?) with a quantitative discussion, and we address RQ2 qualitatively (how
do appraisal judgments relate to textual realizations of events?). 
Since \corpusname
contains annotations from two different perspectives, we describe each
of them separately and in comparison to one another.

Section~\ref{ssec:descriptive} provides general descriptive statistics
about the generation side of the corpus, including patterns across
variables and their correspondence to the validation counterpart.
Section~\ref{ssec:reliability} sharpens the focus on the relationship
between appraisals and emotions in the generation phase. We then
compare such a relationship to the readers' perspective (partially
addressing RQ1).  Section~\ref{ssec:reliability-conditions} narrows
down to inter-annotator agreement computed both on the raw data (RQ1)
and subsampling annotations conditioned on the annotators' properties
(for RQ3). Lastly, in Section~\ref{sec:qualitativedataanalysis}, we
inspect instances in which the validators were either particularly
successful or unsuccessful in recovering the writers' emotions and/or
appraisals (RQ2). This qualitative analysis sheds light on some patterns of
judgments that will be later investigated also in the automatic predictions.

\begin{table}
  \centering\small
  \setlength{\tabcolsep}{3.5pt}
\caption{Data statistics of the generated data (phase 1). \#T: Number
  of Texts, $\overline{\#s}$/$\overline{\#t}$:
  average number of sentences/tokens. s: seconds, m: minutes, h:
  hours, d: days, w: weeks. I: Intensity}
  \label{tab:stats}
 \begin{tabular}{lrrr rrrrr rrrrr r}
 \toprule
   &&&&\multicolumn{5}{c}{event duration}& \multicolumn{5}{c}{emotion duration} \\
   \cmidrule(lr){5-9}\cmidrule(lr){10-14}
   Emotion & \#T & $\overline{\#s}$ & $\overline{\#t}$ & s&m&h&d&w  & s&m&h&d&w & I \\
   \cmidrule(r){1-1}\cmidrule(l){2-2}\cmidrule(l){3-3}\cmidrule(l){4-4}\cmidrule(l){5-5}\cmidrule(l){6-6}\cmidrule(l){7-7}\cmidrule(l){8-8}\cmidrule(l){9-9}\cmidrule(l){10-10}\cmidrule(l){11-11}\cmidrule(l){12-12}\cmidrule(l){13-13}\cmidrule(l){14-14}\cmidrule(l){15-15}
   Anger      & 550 & 1.3  & 21.8 & 69  & 202  & 107 & 68  & 104 & 16 & 108 & 142 & 114 & 170 & 4.2 \\
   Boredom    & 550 & 1.4  & 20.4 & 3   & 105  & 306 & 48  & 88  & 6  & 123 & 297 & 53  & 71  & 3.6 \\
   Disgust    & 550 & 1.4  & 20.6 & 145 & 238  & 58  & 44  & 65  & 30 & 154 & 133 & 97  & 136 & 4.1 \\
   Fear       & 550 & 1.4  & 22.4 & 97  & 233  & 105 & 46  & 69  & 16 & 142 & 143 & 112 & 137 & 4.5 \\
   Guilt      & 275 & 1.3  & 21.9 & 45  & 92   & 62  & 28  & 48  & 9  & 34  & 55  & 58  & 119 & 4.0 \\
   Joy        & 550 & 1.3  & 19.4 & 61  & 156  & 189 & 65  & 79  & 7  & 57  & 150 & 150 & 186 & 4.3 \\
   No Emo.    & 550 & 1.3  & 17.2 & 73  & 256  & 125 & 42  & 54  & 66 & 106 & 65  & 22  & 13  & 2.1 \\
   Pride      & 550 & 1.3  & 19.0 & 67  & 186  & 137 & 49  & 11  & 11 & 54  & 134 & 171 & 180 & 4.2 \\
   Relief     & 550 & 1.4  & 21.7 & 78  & 175  & 140 & 74  & 83  & 32 & 101 & 155 & 121 & 141 & 4.3 \\
   Sadness    & 550 & 1.4  & 20.7 & 55  & 142  & 111 & 85  & 157 & 7  & 27  & 76  & 112 & 328 & 4.5 \\
   Shame      & 275 & 1.3  & 20.6 & 37  & 114  & 59  & 24  & 41  & 1  & 32  & 65  & 74  & 103 & 4.1 \\
   Surprise   & 550 & 1.2  & 18.4 & 110 & 235  & 97  & 51  & 57  & 29 & 107 & 153 & 129 & 132 & 4.1 \\
   Trust      & 550 & 1.3  & 22.4 & 35  & 203  & 153 & 61  & 98  & 15 & 93  & 136 & 93  & 213 & 4.0 \\
   \cmidrule(r){1-1}\cmidrule(l){2-2}\cmidrule(l){3-3}\cmidrule(l){4-4}\cmidrule(l){5-5}\cmidrule(l){6-6}\cmidrule(l){7-7}\cmidrule(l){8-8}\cmidrule(l){9-9}\cmidrule(l){10-10}\cmidrule(l){11-11}\cmidrule(l){12-12}\cmidrule(l){13-13}\cmidrule(l){14-14}\cmidrule(l){15-15}
   $\sum$/Avg.& 6600& 1.3 & 20.4 &  67.3 & 179.8 & 126.8& 52.7 & 81.1 & 18.8 & 87.5 & 131.1 & 100.5 & 148.4 & 4.0 \\
   \bottomrule
\end{tabular}
\end{table}

\subsection{Text Corpus Descriptive Statistics}
\label{ssec:descriptive}

Table~\ref{tab:stats} illustrates features of the generation side in
\corpusname.  The corpus contains 6600 texts, 550 per emotion, except
for \guilt and \shame, having 275 items each.  A text consists of one
or more sentences. As shown in column $\overline{\#s}$, the average
number of sentences is similar across emotions. Texts are also consistent
in terms of length (see $\overline{\#t}$). They comprise
20.43 tokens on average, with \fear and \trust receiving the longest
descriptions (avg.\ 22.36) and \surprise the shortest (avg.\
18.38). Non-emotional expressions have fewer words overall, indicating
that annotators provided less context to communicate non-affective
content. In total, the corpus encompasses 134,851 tokens, excluding
punctuation.\footnote{Tokenization via \textit{nltk},
  \url{https://www.nltk.org}.}

Most texts describe events that took place within minutes or hours
(``event duration'' in Table~\ref{tab:stats}). By contrast, \sadness
has an outstandingly high number of week-long events, and \surprise
and \fear are characterized by a substantial amount of events that
lasted only a few seconds. Interestingly, many texts report on
emotions that persisted over days or weeks (``emotion
duration''). This collides with the view that emotions are short-lived
episodes \cite{Scherer2005}, but it is unsurprising in our annotation
setup. The annotators might have recalled longer emotion episodes in
greater detail, and therefore, they might have recounted those to
focus on a vivid memory. They might also have perceived long-lasting
emotional impacts as being of particular importance (i.e., as special
circumstances fitting one of our text diversification strategies).

It is reasonable to assume that another criterion by which they picked
an episode from their past was the emotion intensity connected to it
(column ``I'' in the table): for all labels but \boredom and
\noemotion, the reported intensity is high. This also translates into
high scores of confidence in the generation phase.  Generally, the
participants trusted their memory about the events they described, with
average self-assigned confidence above 4.4 across all emotions. The
confidence of readers about their own performance is lower, ranging
between 3.4 for the \noemotion instances and 4.1 for \joy, with an
average of 3.9.

Besides confidence, we have a number of other annotation layers that
are not reported in the table. One of them is the emotional state
prior to participation in our study. The values for this variable
are by and large uniformly distributed within each prompting
category. However, they differ across emotion categories: the
highest average value is held by current states of \boredom (2.24)
followed by \joy (2.06), \trust (1.95), and \relief (1.69). The lowest
value is observed for \disgust (1.17). Results are
similar for the validation phase. Concerning personality traits, the
participants reported high scores of \textit{Conscientiousness} (avg.\ 2.32/2.60
in the generation/validation phases) and \textit{Openness} (2.24/1.97).

\begin{table}[t]
  \centering\small
  \renewcommand{\arraystretch}{0.6}
  \caption{Most frequent nouns in each emotion category, sorted by frequency.}
  \label{tab:topics}
\begin{tabularx}{\linewidth}{lX}
  \toprule
  Emotion & Most Frequent Nouns \\
  \cmidrule(r){1-1}\cmidrule(r){2-2}
  Anger & work friend time partner car people child year day job
          husband family boyfriend son member school mother colleague
          week house daughter thing person ex\\
  \cmidrule(r){1-1}\cmidrule(r){2-2}
  Boredom & work time hour day home job friend class room
            night meeting game week one thing house training task
            phone flight tv school lecture weekend traffic lot\\
  \cmidrule(r){1-1}\cmidrule(r){2-2}
  Disgust & friend people man food dog work time child family
            day house person partner colleague car floor boyfriend
            street room parent job school night member cat\\
  \cmidrule(r){1-1}\cmidrule(r){2-2}
  Fear & car night time friend day house dog year child work
         hospital road man people accident family dad spider son partner
         front job hour door way phone park life\\
  \cmidrule(r){1-1}\cmidrule(r){2-2}
  Guilt & friend time child work money partner girlfriend day
          thing family brother school mother son sister relationship
          daughter year dog ex dad parent lot kid father\\
  \cmidrule(r){1-1}\cmidrule(r){2-2}
  Joy & time friend year day child family boyfriend son job dog
        partner birthday birth baby work school life daughter car week
        room month wife song sister holiday\\
  \cmidrule(r){1-1}\cmidrule(r){2-2}
  No Emo. & morning job time work day friend boyfriend year
               school car thing grocery today event life situation shop
            tv task shopping people partner family college\\
  \cmidrule(r){1-1}\cmidrule(r){2-2}
  Pride & work job year son time school daughter university friend
          day degree award team lot week child game student class
          college exam family company result\\
  \cmidrule(r){1-1}\cmidrule(r){2-2}
  Relief & time day work job test house year week friend daughter
           result car surgery school month dog exam cancer university
           partner money home health son night\\
  \cmidrule(r){1-1}\cmidrule(r){2-2}
  Sadness & friend year time family job dog dad day week child
            month boyfriend sister mum life parent daughter cat work
            husband school house home thing people\\
  \cmidrule(r){1-1}\cmidrule(r){2-2}
  Shame & friend work school money day time parent front family
          test thing people sister member exam situation sex lot dad
          class child year wife store partner job\\
  \cmidrule(r){1-1}\cmidrule(r){2-2}
  Surprise & friend birthday year time job party boyfriend work
             sister partner gift car wife week parent girlfriend month
             money day trip person husband house college\\
  \cmidrule(r){1-1}\cmidrule(r){2-2}
  Trust & friend partner time boyfriend husband work life secret
          family car relationship people job doctor day girlfriend
          situation hospital colleague money year person\\
  \bottomrule
\end{tabularx}
\end{table}

The majority of people who disclosed their gender were female (generation:
1639, validation: 710), followed by male (690, 480), and a handful
identifying with gender variants (43, 22).  Their age distribution
has a median of 28 at generation time and 36 in the validation
step. Most participants had a high school-equivalent degree
(generation: 738, validation: 356), an undergraduate degree (975,
527), or a graduate degree (379, 223), and only a few did not have any
formal qualification (9, 5).  Moreover, most people identified as
European (1247, 808) or North American (550, 178).

For an overview of the semantic content of the corpus, we show the
most frequent noun lemmata\footnote{Calculated via SpaCy v.3.2,
  \url{https://spacy.io/api/lemmatizer}.} as a proxy of the described
topics in Table~\ref{tab:topics}. Besides reoccurring terms (e.g.,
family- and work-related ones), which are used to contextualize the
events themselves, some words are more specific to certain emotions,
and they indicate concepts that have a prototypical emotion meaning in
the collective imagination, like ``spider'' and ``night'' for \fear,
``birthday'' for \surprise, ``degree'' and ``award'' for \pride.

\begin{figure}
  \centering
  \includegraphics{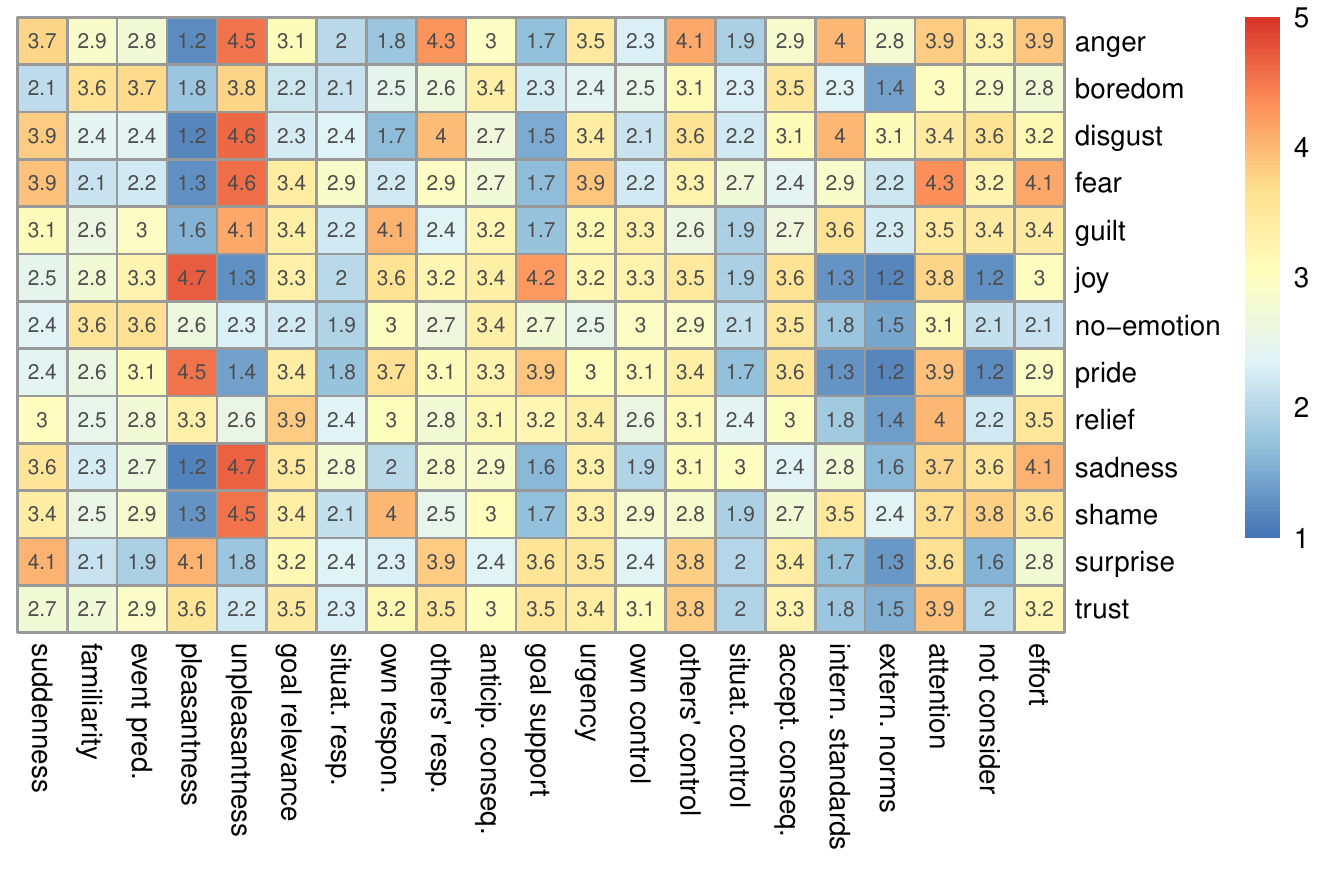}
  \caption{Average appraisal values as found among the writers' judgments,
  divided by emotion. 
  Numbers range between 1 (dark blue) and 5 (dark red).}
  \label{fig:generation-appraisal-emotion}
\end{figure}

\subsection{Relation between Appraisals and Emotions}
\label{ssec:reliability}
Moving on to our core annotation analysis, we investigate the
relationship between appraisal and emotion variables.  We start by
focusing on the generation phase:
Figure~\ref{fig:generation-appraisal-emotion} shows the distribution
of appraisals across emotions as it emerges from the judgments of the
writers.  Each cell reports the value of an appraisal dimension (on
the columns) averaged across all descriptions prompted by a given
emotion (on the rows). High numbers indicate that the appraisal
and emotion in question are strongly related. Low
values tell us that the appraisal hardly holds for that affective
experience.

These results are not only intuitively reasonable but also in line
with past studies in psychology \cite[cf.][]{Smith1985}. We see, for
instance, that events bearing a high degree of \suddenness are related
to \textit{surprise}, \textit{disgust}, \textit{fear}, and
\textit{anger} more than to other emotions.  \textit{Familiarity},
instead, commonly holds for events associated to \noemotion and
\boredom.  Another dimension that stands out for these two labels is
\eventpredictability: its values are comparable to
\textit{familiarity} across all emotions, except for \surprise and
\anger, where it is lower. As expected, \pleasantness and
\unpleasantness are high for positive emotions (i.e., \joy, \pride,
\trust) and negative ones (e.g., \sadness, \shame), respectively.
Among the positive categories, \trust has the highest \unpleasantness
value.  Also \internalStandards and \externSocialStandards
discriminate positive from negative classes, with some within-emotion
differences (events sparking negative emotions, e.g., \disgust, are
deemed to violate self-principles more than social norms).

Next, \boredom and \disgust are associated with low values for the
\goal of events, while the combination of the three responsibility-oriented
appraisals distinguishes a set of emotions: \anger, \disgust, and
\surprise stem from events initiated by others (\otherResp $>$
\situationalResp $>$ \ownresponsibility), \guilt and \shame are
attributed to the self (\ownresponsibility $>$ \otherResp $>$
\situationalResp) and so are \joy and \pride, although to a lower
degree. Once more, \trust differs from the other positive emotions, as
it accompanies events triggered by other individuals or by the
experiencers themselves (e.g., lending someone a precious object) but
not by chance.  It is interesting to compare the
responsibility-specific annotations of \guilt and \shame to the three
dimensions focused on one's ability to influence events. Also there,
the writers felt that the development of the facts was in their
\ownControl more than in the hands of external factors
(\othercontrol/\situationalControl).  Among the two, however,
\ownControl is especially related to \guilt, an emotion stemming
from behaviors that can be regulated rather than from stable traits of
the experiencer (which contribute instead to episodes of \shame
\cite{tracy2006appraisal}). The \textit{anticipation of consequences}
reaches particularly low values for \surprise, \disgust, and \fear,
with the latter being characterized by the strongest level of \effort
(together with \sadness) and of \attention, as opposed to \shame,
\disgust, and \sadness, for which the texts' authors reported their
attempt to dismiss the event.

\begin{figure}
  \centering
  \includegraphics{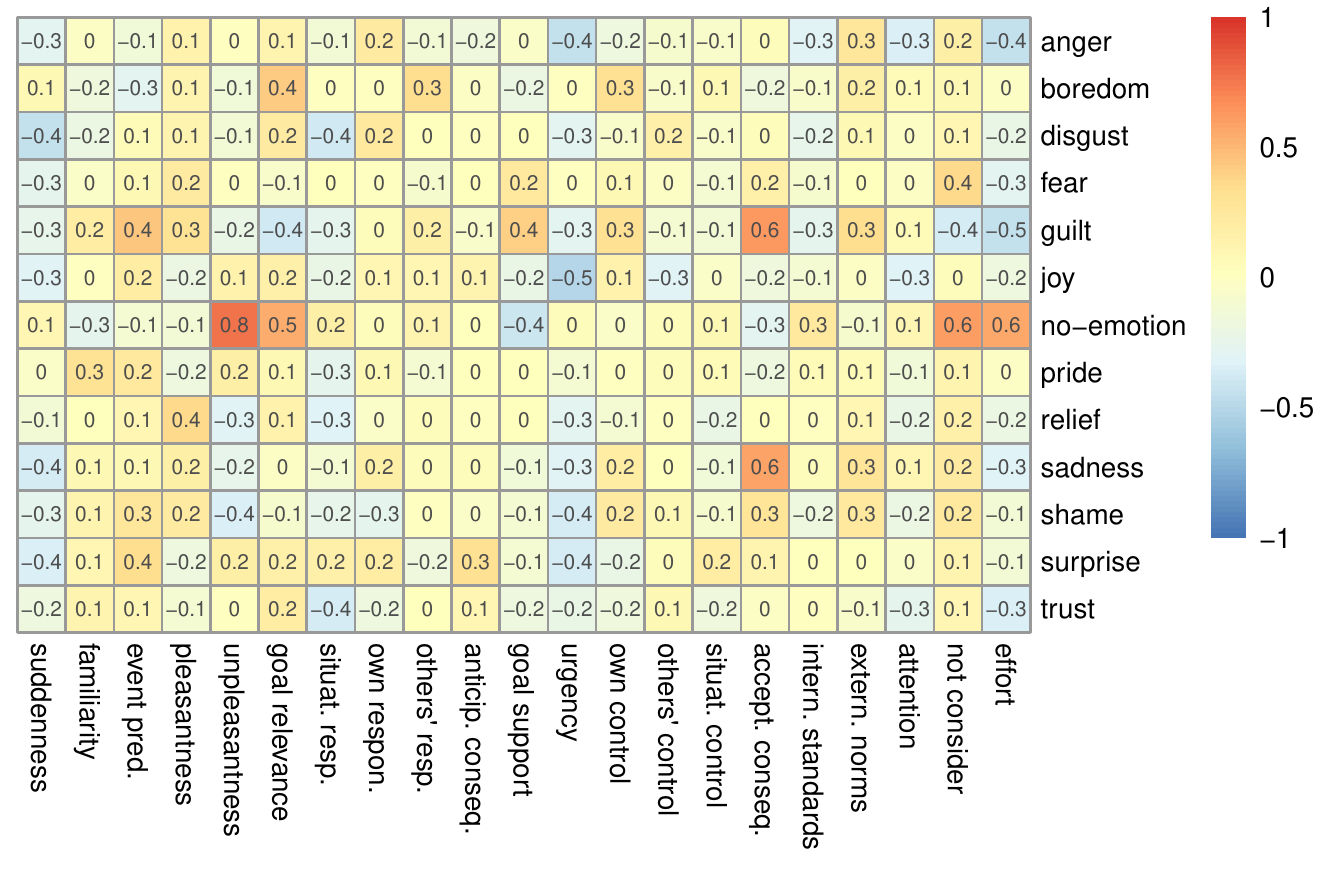}
  \caption{Comparison between the average appraisal values assigned by the generators and the validators,
  divided by emotion. Cells (in the red spectrum) indicate that the generators on average picked higher scores, and vice versa for cells with negative numbers (in blue). Zero values indicate a perfect match between the average scores of the two phases.}
  \label{fig:generation-validation-comparison}
\end{figure}

While these numbers provide a picture of the cognitive dimensions
underlying emotions, they do not answer RQ1 in itself. For that, we
inspect the same information by including the validation side of
\corpusname.  We compare the two batches of judgments in
Figure~\ref{fig:generation-validation-comparison}.  To create this
heatmap, we calculate the average appraisal values across the
prompting emotions -- like in
Figure~\ref{fig:generation-appraisal-emotion}, but using the
validators' appraisal answers and the 1200 corresponding generators'
answers, separately; then, we subtract the results of the former from
the latter. Therefore, a cell here shows the difference between the
average gold standards given by the experiencers and the readers'
assessments.  Should the validators' appraisals be similar to those of
the people who lived through the events (thus approaching 0 throughout
Figure~\ref{fig:generation-validation-comparison}), we could conclude
that it is possible to obtain corpora with reliable appraisal labels via
traditional annotation methods, based on external judges who
determine the affective import of existing texts.

The figure illustrates some interesting patterns: divergent ratings
stand out for \unpleasantness, \goal, \notconsider and
\effort in the row \noemotion, as well as for \urgent in \joy, \effort
in \guilt, and the \acceptConseq in both \guilt and \sadness.
\textit{Suddenness}, \effort and \urgent have lower values across all
emotions, while for \eventpredictability, \externSocialStandards, and
\notconsider, the validators tended to choose ratings that surpassed
the original ones.

Overall, these differences are comparably low (all absolute values are
below 1).
We conclude that readers of event descriptions successfully
reconstruct appraisal dimensions. We now move to a more detailed
analysis of agreements.

\subsection{Conditions of Inter-Annotator Agreement}
\label{ssec:reliability-conditions}
We discuss inter-annotator agreement based on a comparison between
generators and validators (for RQ1) and among the validators. Further,
we scrutinize if their agreement is influenced by some of their
personal characteristics (for RQ3), to understand if there is
a tendency to agree more if the judges share specific properties.  The
datapoints that we consider are extracted from \corpusname as follows.
First, we take all study participants who generated/validated the same
texts and pair them. In total we obtain 6,600 generator--validator
(G--V) pairs (each generator is coupled with 5 validators) and 12,000
validator--validator (V--V) pairs (${5\choose2}\cdot 1,200$). We then
filter the G--V and V--V pairs according to various properties (e.g.,
the age difference between both members of a pair) that either
characterize them or not. This leads to various subsets of annotated
texts (e.g., the subset of texts in which the age difference of the
paired judges is higher than a particular threshold, and the subset
where it is lower).  We only consider the intersection of these text
subsets, i.e., that have been annotated by pairs of all properties
under analysis for one variable.

The properties in question are those collected during corpus
construction, corresponding to the rows in
Table~\ref{tab:iaaconditions}. For gender, annotators can be both
males, both females, or of each of a different gender; for age, we
focus on age differences, as greater or lower than 7 years.
Familiarity with the event concerns the validators only. The
generators know the event by definition; hence, 
only one member of G--V could be unfamiliar with it, while
familiarity can hold or not hold for
both annotators in V--V. Lastly, we
take into account personality traits because past research found that
people with particular traits are better at recognizing
emotions from facial expressions (cf. Section~\ref{sec:reliability}).
We investigate if a similar phenomenon happens in text, and filter the
pairs like so: did the validator(s) turn out to be open,
conscientious, extraverted, agreeable, or emotionally stable?

\begin{table}
  \centering
  \caption{Conditions of reconstruction agreement. Emotion agreement
    is shown as \F and Acc. between pairs of labels from a generator
    and a validator (G--V) or two validators (V--V). The appraisal
    agreement is an average root mean square error. \#Pairs denotes
    the number of G--V (1st) or V--V (2nd) pairs for each
    condition. Boxes indicate measures computed on the same textual
    instances, which can therefore be directly compared. $^*$ indicates
    all pairs that are significantly different from each other inside
    a box; calculated with 1000$\times$ bootstrap resampling,
    confidence level .95.}
  \label{tab:iaaconditions}
  \includegraphics{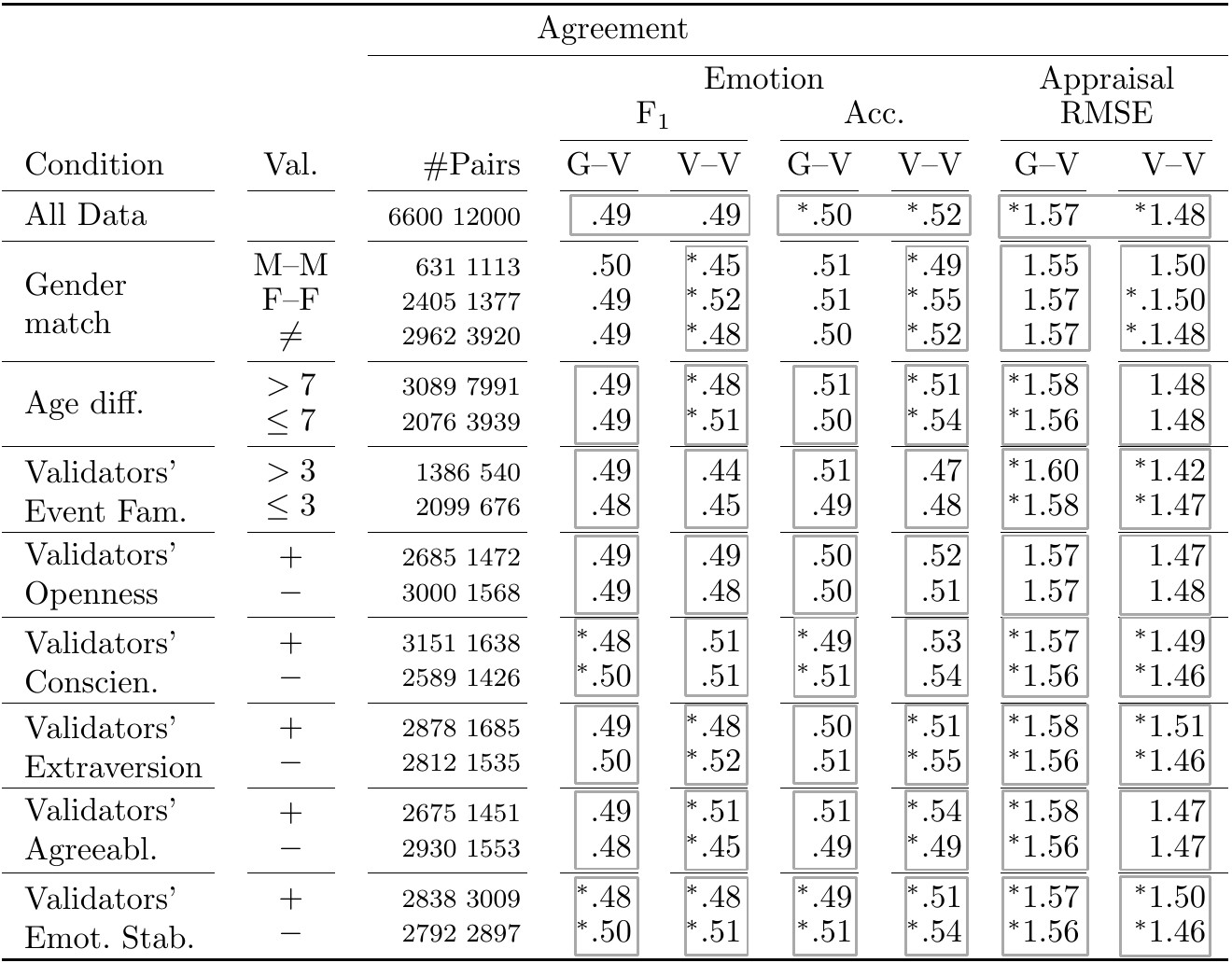}
\end{table}

From all of these data subsets, we compute agreement through multiple
measures. For emotions, we use average \F and accuracy, for
appraisal annotations, we employ average $\textrm{RMSE}$ scores.  We
do not normalize for expected agreement, as it is commonly done
with $\kappa$ measures, because we do not have unique annotators that
remain stable over a considerable amount of texts -- which prevents us
from assigning a meaningful value for the expected agreement to each
individual.

Table~\ref{tab:iaaconditions} summarizes the results.  Note that the
number of pairs varies depending on the property under consideration,
as different properties might hold for different numbers of people.
Boxes indicate that the results have been obtained from the same
textual instances. We can therefore compare numbers inside the same
box, but not across boxes (either because they refer to different
evaluation methods or different concepts, or because they were calculated
on different textual instances). We calculate the significance under a
.95 confidence level via bootstrap resampling (1000 samples) on the
textual instances for each evaluation measure, pairwise for all
results inside each box \citep{Canty20211,Davison1997}. Pairs of
asterisks indicate pairs of numbers that are significantly different
(all of them are, if three values are marked with a $^*$).

The row ``All Data'' in the table contains all annotation pairs, not
filtered by any property. In this row, G--V and V--V values can be
compared.  We see that the agreement on emotions of the V--V pairs is
higher than that achieved by the G--V pairs, with 2pp (significant)
difference in accuracy (no difference in \F). We take this as a
first sign of the validators' reliability and correct interpretation
of the guidelines: they agree with the point of view that they are
attempting to reconstruct and with all other judges undertaking
the same task. The significant accuracy difference between G--V and
V--V pairs stems mostly from a mismatch between generators and
validators on \joy/\surprise (prompting/validated emotion), \joy/\pride, \joy/\relief,
\sadness/\anger, and \noemotion/\boredom. We will analyze these cases in more detail
later on. The difference in agreement for the annotation of appraisals is
also significant, and more noticeable (1.57 for G--V vs.\ 1.48 for
V--V). The biggest difference holds for \notconsider, followed by
\otherResp and \situationalResp. There is no appraisal dimension in
which G--V pairs outperform V--V pairs.

In all other rows are results obtained from annotator pairs filtered
by property.  For the gender matches, we tackle the groups with the
most common self-reported answers (i.e., male, female).  Note that the
numbers under G--V and V--V do not come from the same texts (in our
data, being male and female are mutual exclusive
properties\footnote{For a text, e.g., M--M might be among the
  validator pairs, but not in the generator-validator ones in case the
  generator is female.}). For emotions, mixed-gender pairs disagree
significantly more than the female subsets. This is also the case for
male pairs, with a 7pp difference in \F. Mostly, females agree more on
what is considered to cause \shame and \guilt, where males tend to
disagree.

To evaluate the impact of age on agreement, we
separate the pairs at a threshold of 7 years (we tested other
thresholds, which lead to smaller differences). Interestingly, all
differences are comparably small ($<$3pp in \F and acc.), but
still significant for emotions among V--V pairs and for appraisals among G--V
pairs.

The property of event familiarity (self-assessed on a 1--5
scale) leads to a significant difference in the appraisal
assessments. Interestingly, \textit{non}-familiar validators tend to
agree more with generators and each other than those that indicated to
be familiar with an event. A possible explanation is
that readers who did not experience an event similar to the
description rely purely on the information emerging from the text and
are not biased from their own experiences.

For the analysis of the influence of validators' personality traits,
we split the validators with a threshold that approximates a balanced 
separation of all judges. The trait of \textit{Openness} does not show any
significant relation to agreement across all measures and
annotation variables. The traits of \textit{Conscientiousness} and
\textit{Introversion} show a small but
significant positive impact on the agreement measures. Validators that
indicated to be \textit{Agreeable} show significant and considerably
higher agreement with each other in the emotion labeling task. A lack of
\textit{Emotional Stability} corresponds to a small but significant improvement
in agreement across both emotions and appraisals and G--V/V--V pairs.

We did not find any substantial difference between the general
agreement and the agreement between groups of the same ethnicity or
education.

In summary, the analysis of inter-annotator agreement conditioned on
self-reported personal information revealed that better emotion and
appraisal reconstructions are favored by specific properties.
However, differences between groups of judges with diverse
properties are small, and the within-validation phase
agreement compares to that between generators and validators,
considering agreement on all data
irrespective of group filterings.  Hence,
we conclude that \textit{the annotations provided by the readers are
  reliable.}

\begin{table}[t]
  \newcommand{\twoc}[1]{\multicolumn{2}{c}{#1}}
  \centering\small
  \setlength{\tabcolsep}{4pt}
  \renewcommand{\arraystretch}{0.9}
   \caption{Examples where all
    validators (V) correctly reconstruct the emotion of the generators (G). 
    The top (bottom) examples have
    high (low) agreement on appraisals.}
  \label{tab:agreementexamples1}
  \begin{tabularx}{1.0\linewidth}{lcccX}
    \toprule
    & \multicolumn{2}{c}{Emo.} & Appr. \\
    \cmidrule(r){2-3}\cmidrule(r){4-4}
    Id & G & V & $\textrm{RMSE}$ & Text \\
    \cmidrule(r){1-1}\cmidrule(lr){2-2}\cmidrule(lr){3-3}\cmidrule(lr){4-4}\cmidrule(l){5-5}
1 & \twoc{pride} & 0.65 & I baked a delicious strawberry cobbler. \\
2 & \twoc{fear} & 0.69 & I was running away from a shooting and a car was trying to run me down \\
3 & \twoc{fear} & 0.72 & I felt ... when there was a power outage in my home. That day, my wife and I were cuddling in the sitting room when a thunderstorm started. Then ... filled me when thunder hit our roof and all the lights went off. \\
4 & \twoc{pride} & 0.82 & I felt ... when I ran a marathon at a decent pace and finished the race in a good place \\
5 & \twoc{fear} & 0.84 & A housemate came at me with a knife. \\
6 & \twoc{fear} & 0.86 & I was surrounded by four men; they hit me in the face before I offered to give them everything I had in my pockets. \\
7 & \twoc{pride} & 0.89 & I felt ... when I accomplish my goals through a team effort. I take part in team sports and have a pivotal role in success, and being able to do my job and make my team proud of me gives me a strong sense of .... \\
    \cmidrule(r){1-1}\cmidrule(lr){2-2}\cmidrule(lr){3-3}\cmidrule(lr){4-4}\cmidrule(l){5-5}
203 & \twoc{fear} & 1.68 & I felt ... when I was in a public place during the coronavirus pandemic \\
204 & \twoc{pride} & 1.73 & I helped out a friend in need \\
205 & \twoc{fear} & 1.74 & I felt ... when i had a night terror. \\
206 & \twoc{boredom} & 1.81 & I went on holiday abroad for the first time. I felt ... because I didn't enjoy being on the beach doing nothing. \\
207 & \twoc{sadness} & 1.86 & I felt ... when I graduated high school because I remember that I'm growing up and that means leaving people behind. \\
208 & \twoc{disgust} & 2.03 & His toenails where massive \\
209 & \twoc{fear} & 2.08 & I felt ... going in to hospital \\
210 & \twoc{trust} & 2.35 & my husband is always there for me and i
                            can ... that no matter what he will be
                            there for our child and do what ittakes to
                            provide for us as a family \\
    \bottomrule
  \end{tabularx}
\end{table}

\subsection{Qualitative Discussion of Differences}
\label{sec:qualitativedataanalysis}
A manual inspection of the data deepens our understanding of
inter-annotator agreement. We investigate the texts on which
judges (dis)agree and divide them into two categories, i.e., those that
turned out ``easy'' to label and those where the correct inference is
difficult to draw.  Table~\ref{tab:agreementexamples1} shows examples
in which all readers correctly reconstructed the writers' emotion.
Table~\ref{tab:agreementexamples2} reports items where all validators
inferred the same emotion, but that emotion does not correspond to the
gold label -- as revealed in the quantitative discussion, agreeing on
the emotion does not imply agreeing on the appraisals. We report
this observation by dividing the tables in two blocks.  The top block
corresponds to texts with high G--V agreement in appraisal (as an
average $\textrm{RMSE}$), while the bottom to high disagreement.

The top examples in Table~\ref{tab:agreementexamples1} describe events, varying from
ordinary circumstances (e.g., baking) to peculiar ones (e.g., being threatened
by a housemate) that have unambiguous implications for the well-being of the
experiencer. It can be argued that these texts describe situations with shared underlying
characteristics graspable even by people who did not experience them (e.g.,
most likely, being threatened spurs \unpleasantness, scarce \goal, and
inability to \anticipationConseq).  By contrast, the examples with low
agreement on appraisals seem to require a more elaborate empathetic
interpretation.  They might be easily understandable with regard to the
emotion, but they underspecify many details about the described situation,
which would be necessary for a reader to infer how it was evaluated along
fine-grained dimensions. For instance, going to the hospital is attributed to
fear, but it remains unclear under which circumstances this situation occurs (a
planned surgery? an accident? to visit someone?).

\begin{table}[t]
  \centering\small
  \setlength{\tabcolsep}{4pt}
  \renewcommand{\arraystretch}{0.9}
  \caption{Examples in which all validators (V) agree with each other, but
    not with the generators (G) of the event descriptions. The top (bottom) blocks shows texts
    where the agreement is high (low) on appraisals.}
  \label{tab:agreementexamples2}
  \begin{tabularx}{1.0\linewidth}{lcccX}
    \toprule
    & \multicolumn{2}{c}{Emo.} & Appr. \\
    \cmidrule(r){2-3}    \cmidrule(r){4-4}
    Id & G & V & $\textrm{RMSE}$ & Text \\
    \cmidrule(r){1-1}\cmidrule(lr){2-2}\cmidrule(lr){3-3}\cmidrule(lr){4-4}\cmidrule(l){5-5}
1 & joy & pride & 0.81 & finally mastered a song i was practising on guitar \\
2 & pride & joy & 0.83 & my band got signed to a label run by an artist i admire \\
3 & trust & joy & 0.87 & I am with my friends \\
4 & joy & pride & 0.90 & I bought my own horse with my own money I had worked hard to afford \\
5 & surprise & pride & 0.93 & when I built my first computer \\
6 & surprise & joy & 1.00 & I felt ... when my partner put their arms around me at a concert and started to dance with me to a song we listen to. \\
7 & trust & joy & 1.01 & I felt ... when my boyfriend drove out of town to see me at 2 in the morning. \\
8 & anger & fear & 1.09 & My waters broke early during pregnancy \\
9 & joy & pride & 1.11 & I was able to complete a challenge that I didn't think I would do \\
    \cmidrule(r){1-1}\cmidrule(lr){2-2}\cmidrule(lr){3-3}\cmidrule(lr){4-4}\cmidrule(l){5-5}
43 & pride & sadness & 1.65 & That I put together a funeral service for my Aunt \\
44 & surprise & joy & 1.66 & I got a dog for my birthday \\
45 & joy & relief & 1.68 & I was diagnosed with PMDD because it meant I had answers \\
46 & no-emotion & anger & 1.69 & I saw an ex-friend who stabbed me in the back with someone I considered a friend \\
47 & shame & relief & 1.81 & I tasked with sorting out some files from the office the previous day and I slept off when I got home \\
48 & disgust & sadness & 1.82 & I was left out of a family chat. \\
49 & sadness & relief & 1.83 & when I returned to my apartment after being away during COVID. \\
50 & shame & sadness & 1.84 & Not being around my son \\
51 & surprise & joy & 1.90 & I found the perfect man for me, and the more time goes on, the more I realized he was the best person for me. Every day is a .... \\
52 & no-emotion & sadness & 1.93 & Breaking up with my partner \\
    \bottomrule
  \end{tabularx}
\end{table}

Table~\ref{tab:agreementexamples2} contains texts from which readers did not
recover the actual emotion experienced by the author. Instances of
high appraisal agreement are associated to labels with similar
affective meanings, and are therefore more likely to be confused
than, for instance, a positive and a negative emotion. Mislabeling occurs mostly
between \joy and \pride, both of which are (arguably)
appropriate, and in one case between \anger and \fear.  Instead, the
bottom block of the table reports texts in which a positive emotion is
misunderstood for a negative one.  For instance, Id 43 was produced
for \pride but was validated as \sadness. These mistakes might be due to the
readers focusing on a portion of text different from that
considered salient by the writer (e.g., Id 49, ``being away with
covid'': \sadness, ``returning home'': \relief), or to the readers
drawing a presupposition from the text (e.g., Id 43, a funeral took
place: \sadness) different from what the author intended to convey
(he/she was able to organize it: \pride).  It is also possible that
some of these G--V disagreements derive from the sequence of tasks
in the survey.  The readers were first prompted to assign an emotion to
the event and only later were they guided to evaluate it in detail.  Going
the other way around might have led the crowdworkers to reflect on the
events in a more structured way, and might have elicited different
judgements.  There are also examples in which an emotion is assigned
while none was felt by the event experiencer (e.g., Id 46 and 52).
On the one hand, this is a sign of the subjectivity of emotions.
On the other, it tells something about how some writers
tackled the task: they likely decided to recount a
circumstance that usually would not leave individuals in apathy
but that, unexpectedly, turned out to not perturb their own general sense of feeling.

Brought together, these observations illustrate features of
\corpusname, and suggest some systematic patterns in its annotation
that are informative about agreement.  To begin with, part of the
instances that we collected convey enough information for readers to
understand emotions, independent of if and how they also understand
the underlying evaluation.  From this, we derive that \textit{at least
  in some cases, grasping appraisals from text is not necessary to
  grasp the corresponding emotion} -- which is an insight that we
further explore in our modeling experiments, by using
systems for emotion recognition that can decide to leverage or ignore
appraisals information.

Second, by contrasting the high-vs.-low appraisal agreements blocks of
Table~\ref{tab:agreementexamples2}, we learn that the ``semantic
difference'' between emotions that are incorrectly reconstructed is
lower if the appraisals are inferred acceptably well (e.g., readers
picked \pride instead of \joy, while they face confusions between more
incongruent labels, e.g., \pride/\sadness, by disagreeing also on the
appraisals). Put differently, the annotators can share the underlying
understanding of an affective experience, even if they disagree on a
discrete label to name it. Hence, the labels they choose can be
considered compatible alternatives.  As our single-label experimental
setup did not request the description authors to indicate multiple
emotion labels for their experience, a follow-up study would be needed
to confirm this hypothesis.

Third, there are instances where humans fail to reconstruct emotions,
and differences between such judgments are mirrored in differences in
their appraisal measures. We hypothesize that the correct appraisal
information can be valuable for improving the emotion classification
of these instances -- it might disambiguate alternatives by offering
information that is not described in the text. In the modeling setup,
we explore this idea by looking at how an
appraisal-aware emotion recognition model improves as it accesses
evaluations-centered knowledge.

\begin{figure}
  \centering
  \includegraphics[width=\linewidth]{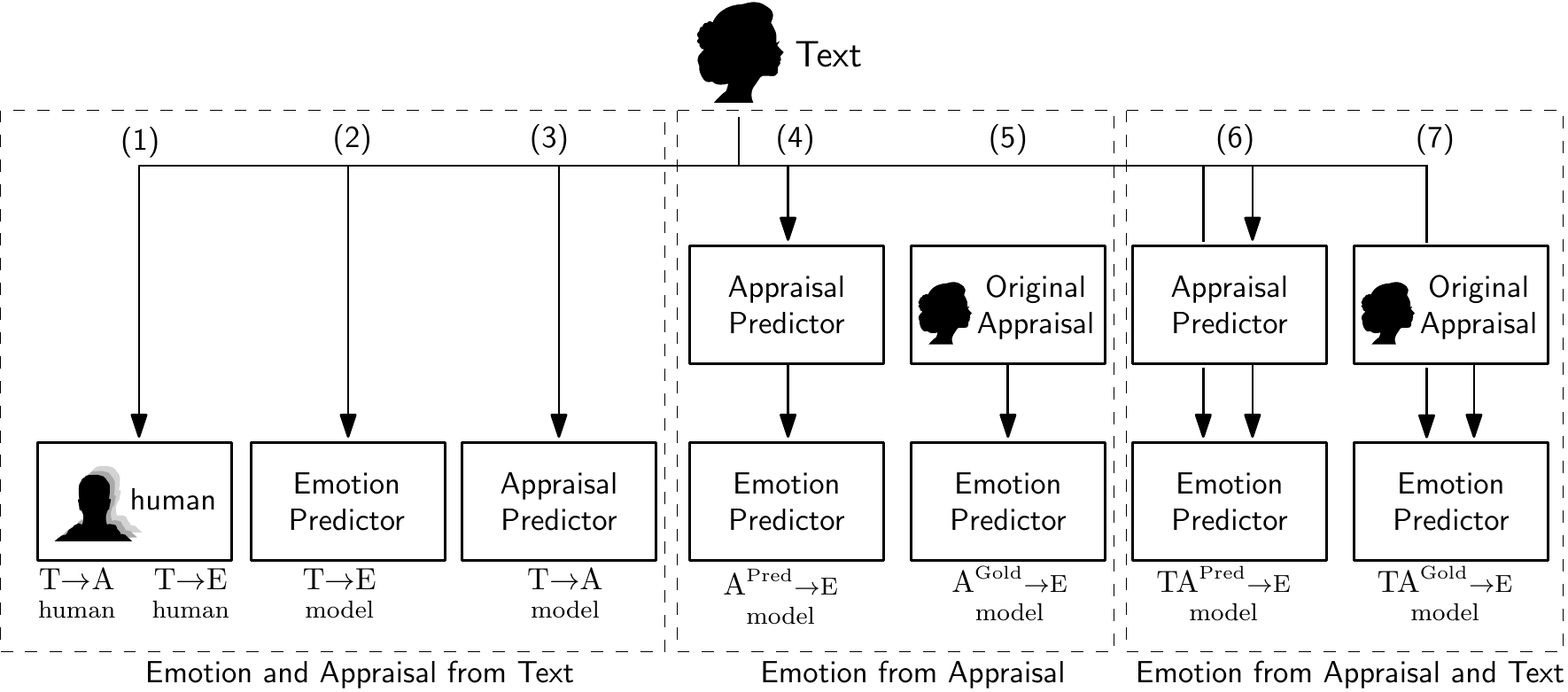}
  \caption{Models to predict
    emotions and appraisals, and to understand interactions among
    tasks.}
  \label{fig:modelingsetup}
\end{figure}

\section{Modeling Appraisals and Emotions in Text}
\label{sec:modeling}
In the preceding section, we answered RQ1 from an annotators-based
perspective (Is there enough information in a text for humans to
predict appraisals?). We answer the same question here, but by turning
to a computational modeling discussion: Is there enough information in
a text for classifiers to predict appraisals? Our ultimate goal is
to understand if these psychological models are not only usable but
also advantageous for emotion analysis. Therefore, we also address
RQ4: Do appraisals practically enhance emotion predictions?

We formalize different models and motivate the relationship between
them. Next, we put them to use to predict emotion categories and appraisal
dimensions. Such models consist of three main classes that vary with
respect to their input, output, and sequence of steps.  We have:
models that take text as their only input and that output either an
emotion category (\TE) or appraisal dimensions (\TA); models that use
only appraisal patterns as input to predict emotions (\aE); emotion
predictors with mixed input, informed by both text and appraisals
(\TAE).

As explained below, each model mirrors a precise view on the emotion
component process theory.  By evaluating their predictions against the
ground truth labels, we can validate the underpinning theory from a
text classification perspective. For instance, if emotions arise
deterministically from the 21 dimensions that we study, our
appraisal-to-emotion classifiers (\aE) should work acceptably
well. Moreover, if the information concentrated in the event
description is enough to reconstruct appraisals, then \TA should show
a good performance, and the consequent step from there to emotions
(\aE) should be straightforward.

\subsection{Model Architecture and Experimental Setup}
\newlength{\mylen} \settoheight{\mylen}{A}
Figure~\ref{fig:modelingsetup} illustrates our experimental framework.
In total, we consider 7 models. A box in the depiction indicates a
model (the head \includegraphics[height=\mylen]{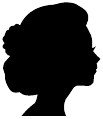}\
 indicates data that directly stems from the generator
of a textual instance). The lines correspond to the flow of information
used by the box connected with an arrowhead.  The
left-most model (denoted as (1) in the depiction) is not a
computational system, but represents the classification
performance of the validators of \corpusname.  We include that in
order to understand how well people performed in the task undertaken by
our machine learning-based systems (indicated by the numbers from (2)
to (7)).  Specifically, we focus on how the readers predicted the
prompting emotions and the correct appraisals from text, treating
these two ``human models'' separately (i.e., \TEhumanText and
\TAhumanText). Under the assumption that humans outperform
computational models, (1) will act as an upper bound
for the automatic classifiers.

We use (2) as a baseline computational model to predict emotion
categories for a given text (\TEmodelText), learning the task in an
end-to-end fashion.  From a psychological perspective, this classifier
aligns with theories of basic emotions discussed in
Section~\ref{sec:basicemotions}, as it is purely guided by the
definition of the output categories -- although only a subset of our
11 emotion labels would be considered ``basic'' in the strict definition
of \citet{Ekman1992}.  The model in (3) is set up analogously, but it
predicts a vector of appraisal values. This \TAmodelText can be
considered in line with a constitutive theoretical approach (as
described by \citet{Smith1985} or \citet{Clore2013}) where the
appraisal variables instantiated in response to an event represent the
emotion itself -- hence, they do not serve as input to predict a
consequent discrete emotion label.  Even without such an additional
step, this model can be practically useful, similar to emotion
analysis systems that output scores of valence or arousal.

We further employ (3) in the pipeline represented in (4), which
performs the additional appraisal-to-emotion step.
There, the emotion predictor is trained on the appraisal-based output of (3).
To evaluate this \AEmodelPredText, we compare it against
\AEmodelGoldText (5), which is required to accomplish the same
emotion prediction task, but is trained on the
writers' original appraisal judgments.

Lastly, we instantiate two combined models, \TAEmodelPredText (6) and
\TAEmodelGoldText (7), which have access to both the texts and the
corresponding appraisals. These consist of the predictions of
\TAmodelText for \TAEmodelPredText, and of the judgments provided by
the event experiencers for \TAEmodelGoldText.  Being pipelines, all
models from (4) to (7) have a structural affinity to the evaluative
tradition of emotion theories (Section~\ref{sec:emotion-theories})
that involves a deterministic perspective on emotions: the appraisals
of an event cause the emotion experience \cite{Scherer2001,
  Scherer2013}.  However, as opposed to (4) and (5), models (6) and
(7) do not follow a strict pipeline architecture, as they can decide
to bypass the appraisal information, if not needed for the emotion
prediction.

To bring all these models together into an evaluation agenda,
we conduct three experiments.
\begin{description}[noitemsep,topsep=0pt]
\item [\textbf{Experiment 1:}] We use \TAmodelText and \TAhumanText.
  The human model enables us to assess the task's difficulty. It
  informs us about what we can reasonably expect from the
  systems. Therefore, we use it as a benchmark to evaluate
  \TAmodelText and consequently answer RQ1.
\item [\textbf{Experiment 2:}] Before assessing if appraisals are
  beneficial for the prediction of emotions from text (RQ4), we need
  to verify if emotions can be inferred from such 21 appraisal
  dimensions.  According to psychology, humans do the appraisal-to-emotion
  mapping in real life. Here we investigate if that
  is the case also for machine learning-based models
  (\AEmodelGoldText/\AEmodelPredText).
\item [\textbf{Experiment 3:}] We use \TAEmodelPredText,
  \TAEmodelGoldText, \TEmodelText and \TEhumanText.  We investigate if
  the appraisal-informed models have any advantage over the latter
  two, which are only based on text. Hence, we answer RQ4.
\end{description}

Since each of the 1200 validation instances was evaluated by 5
different annotators, we aggregate all judgments (instance by
instance) into a final, adjudicated label, thus obtaining the same
level of granularity that we have for the automatic predictions.  We
use the majority vote for the aggregation of both emotions and
appraisals. We do not opt for averaging the appraisal judgments as
this would flatten the annotation of the various dimensions and not
account for differences in their reconstruction.  Whenever the
majority vote leads to a tie, we resolve it by assigning a higher
weight to the appraisal judgments of annotators who self-assigned a
strong degree of confidence.

For all computational models, we use the same 1200 instances that have
been validated by the human annotators as a test set. We randomly
split the remaining 5400 generation instances into training and
validation data (90\,\% for training, i.e., 4860 instances; 10\,\% for
validation, i.e., 540 instances) without strictly enforcing
stratification by prompting emotion label.

The emotion predictors are classification models which choose one
single label from the set of prompting emotions. The appraisal models
are instantiated twice: as regressors that predict a continuous value
in [0:1], and as classifiers in a discretized variant of the problem.
For that, we map the 5-point scales of the appraisal
ratings to 0, corresponding to \{1,2,3\} in the original answer, and
1, if the original answer was \{4,5\}.\footnote{The decision on this
  threshold derives from the distribution analysis shown in
  Figure~\ref{fig:plot-violins} in the Appendix.}  Approaching this
problem in a classification setup allows us to compare our results to
previous work \cite{Hofmann2020}, and see if the systems agree with
humans at least about an appraisal holding or not (more than
recognizing its fine-grained value).

Our implementation builds on top of Huggingface Transformers
\cite{Wolf2020}.  All experiments take the pretrained RoBERTa-large
model \cite{Liu2019} as a backbone, implemented in the AllenNLP
library \cite{Allennlp}. Depending on the task, we use a
classification or regression layer on top of the average-pooled output
representations. The training objective is to minimize the
cross-entropy loss for all text-based classifiers, and the mean square
error loss ($\textrm{MSE}$) for all regressors. We report the mean of
the results across 5 runs and use validation data to perform early
stopping. The learning rate is $3\cdot10^{-5}$, the maximal number of
epochs is 10, and the batch size is 16.

For \AEmodelGoldText and \AEmodelPredText, we use a single-layer
neural network with 64 hidden nodes, ReLU \cite{Nair2010} activation,
and a dropout rate \cite{Srivastava2014} of 0.1 between the hidden and
input layer. The training objective is to minimize the cross-entropy
loss. For \TAEmodelPredText and \TAEmodelGoldText, we concatenate the
vector of appraisal values to the pooled vector representation of the
textual instance before the output layer.

\subsection{Results}
\label{sec:Evaluation}
\noindent We analyze the modeling results with traditional evaluation
metrics for text classification, namely, macro-averaged precision,
recall and macro-\F .  The appraisal regressors are evaluated via
$\textrm{RMSE}$.  All reported scores are averages across 5 runs with
different seeds. Standard deviations are in Appendix,
Table~\ref{tab:table-appraisals-sd} (for Experiment~1),
Table~\ref{tab:table-appraisals2emotions-sd} (for Experiment~2), and
Table~\ref{tab:table-emotions-sd} (for Experiment~3).\footnote{Note
  that the evaluation of the human-based models leads to different \F
  values than those in the corpus analysis section, where we
  considered multiple pairs of human-generated labels for each text --
  here, we have aggregated judgments. The individual predictions for
  all models are part of the supplementary material.}

\begin{table}
  \centering\small
  \newcommand{\sd}[2]{#1{\tiny$\pm$#2}}
  \setlength{\tabcolsep}{6.4pt}
  \caption{Appraisal prediction performance of human validators
    (\TAhumanText) and computational models (\TAmodelText). For the classification setup, we report
    Precision, Recall and \F. For the regression
    setup, we report the root mean square error ($\textrm{RMSE}$).
    $\Delta$: difference between \TAmodelText and \TAhumanText. Standard deviations are reported in Table~\ref{tab:table-appraisals-sd} in the Appendix.}
  \label{tab:text2appraisal}
\begin{tabular}{lccccccrccr}
    \toprule
    & \multicolumn{7}{c}{Classification} & \multicolumn{3}{c}{Regression} \\
    \cmidrule(lr){2-8}\cmidrule(lr){9-11}
    & \multicolumn{3}{c}{\TAhuman}  & \multicolumn{3}{c}{\TAmodel} &&  \TAhuman & \TAmodel & \\
    \cmidrule(r){2-4}\cmidrule(lr){5-7}\cmidrule(lr){9-9}\cmidrule(lr){10-10}
    Appraisal       & P    & R    & \F   & P    & R    & \F   & $\Delta_{\small{\textsf{\F}}}$ & $\textrm{RMSE}$  & $\textrm{RMSE}$ & $\Delta_{\small{\textrm{RMSE}}}$ \\
  \cmidrule(r){1-1}\cmidrule(rl){2-2}\cmidrule(rl){3-3}\cmidrule(rl){4-4}\cmidrule(rl){5-5}\cmidrule(rl){6-6}\cmidrule(rl){7-7}\cmidrule(rl){8-8}\cmidrule(rl){9-9}\cmidrule(rl){10-10}\cmidrule(l){11-11}
  Suddenness       & $.75$ & $.61$ & $.68$ & $.70$ & $.79$ & $.74$ & $+.06$       & $1.47$ & $1.33$ & $-.14$ \\
  Familiarity      & $.66$ & $.45$ & $.53$ & $.77$ & $.82$ & $.79$ & $+.26$      & $1.49$ & $1.42$ & $-.07$ \\
  Event Predict.   & $.60$ & $.54$ & $.56$ & $.76$ & $.74$ & $.75$ & $+.19$      & $1.46$ & $1.47$ & $+.01$ \\
  Pleasantness     & $.82$ & $.84$ & $.83$ & $.88$ & $.87$ & $.88$ & $+.05$       & $1.10$ & $1.30$ & $+.20$ \\
  Unpleasantness   & $.85$ & $.84$ & $.85$ & $.79$ & $.80$ & $.80$ & $-.05$       & $1.22$ & $1.26$ & $+.04$ \\
  Goal Relevance   & $.65$ & $.67$ & $.66$ & $.73$ & $.69$ & $.71$ & $+.05$       & $1.52$ & $1.57$ & $+.05$ \\
  Situat.\ Resp.   & $.70$ & $.37$ & $.48$ & $.83$ & $.87$ & $.85$ & $+.37$      & $1.55$ & $1.43$ & $-.12$ \\
  Own Resp.        & $.75$ & $.71$ & $.73$ & $.81$ & $.77$ & $.79$ & $+.06$       & $1.32$ & $1.40$ & $+.08$ \\
  Others' Resp.    & $.75$ & $.74$ & $.74$ & $.74$ & $.72$ & $.73$ & $-.01$       & $1.54$ & $1.57$ & $+.03$ \\
  Anticip.\ Conseq.& $.57$ & $.48$ & $.52$ & $.67$ & $.71$ & $.69$ & $+.17$      & $1.61$ & $1.50$ & $-.11$ \\
  Goal Support     & $.74$ & $.62$ & $.67$ & $.80$ & $.82$ & $.81$ & $+.14$      & $1.36$ & $1.33$ & $-.03$ \\
  Urgency          & $.66$ & $.46$ & $.54$ & $.63$ & $.60$ & $.61$ & $+.07$       & $1.68$ & $1.43$ & $-.25$ \\
  Own Control      & $.57$ & $.48$ & $.53$ & $.78$ & $.81$ & $.79$ & $+.26$      & $1.48$ & $1.35$ & $-.13$ \\
  Others' Control  & $.76$ & $.76$ & $.76$ & $.64$ & $.60$ & $.62$ & $-.14$      & $1.55$ & $1.36$ & $-.19$ \\
  Situat.\ Control & $.71$ & $.40$ & $.51$ & $.84$ & $.90$ & $.87$ & $+.36$      & $1.53$ & $1.35$ & $-.18$ \\
  Accept.\ Conseq. & $.48$ & $.39$ & $.43$ & $.63$ & $.65$ & $.64$ & $+.21$      & $1.44$ & $1.36$ & $-.08$ \\
  Internal Standards       & $.68$ & $.51$ & $.57$ & $.82$ & $.83$ & $.82$ & $+.25$      & $1.16$ & $1.34$ & $+.18$ \\
  External Norms   & $.63$ & $.52$ & $.56$ & $.90$ & $.95$ & $.92$ & $+.36$      & $1.77$ & $1.44$ & $-.33$ \\
  Attention        & $.74$ & $.75$ & $.74$ & $.50$ & $.48$ & $.48$ & $-.26$      & $1.38$ & $1.27$ & $-.11$ \\
  Not Consider     & $.55$ & $.53$ & $.54$ & $.83$ & $.71$ & $.77$ & $+.23$      & $1.56$ & $1.53$ & $-.03$ \\
  Effort           & $.70$ & $.54$ & $.61$ & $.69$ & $.70$ & $.70$ & $+.09$       & $1.47$ & $1.38$ & $-.09$ \\
  \midrule
  Macro avg.       & $.68$ & $.58$ & $.62$ & $.75$ & $.75$ & $.75$ & $+.13$      & $1.46$ & $1.40$ & $-.06$ \\
  \bottomrule
\end{tabular}
\end{table}

\subsubsection{Experiment 1: Reconstruction of appraisal from event
  descriptions}
Table~\ref{tab:text2appraisal} illustrates the outcome of
appraisal reconstruction from text, carried out computationally and
by the validators. Both the automatic classifier and the regressor have
an acceptable performance, with a .75 macro-average \F and an averaged 
$\textrm{RMSE}$=$1.40$. 

Focusing on the classification task, the dimensions of
\externSocialStandards, \pleasantness and \situationalControl
correspond to better quality outputs, especially compared to \urgent,
\othercontrol, and \acceptConseq where \F is the lowest.  Dimensions
that are easy/hard to reconstruct from a computational perspective are
so also for the validators.  Overall, however, the computational model
achieves higher results than human validators. As we see in the column
$\Delta_{\small{\textsf{\F}}}$, which reports the differences between
\TAmodelText and \TAhumanText, classification models show 13pp higher
\F.

The same trend emerges from the regression framework, where the
average error drops by 6 points.  The improvement is not equally
distributed across appraisals.  It stands out on the dimensions of
\externSocialStandards, \urgent, \othercontrol, \situationalControl,
and \suddenness. In many of these variables with a
$\Delta_{\small{\textrm{RMSE}}} <$.10, the original ratings are spread
more uniformly across the 5-point answer scale (cf. \othercontrol,
\suddenness, \anticipationConseq in Figure~\ref{fig:plot-violins},
Appendix). This suggests that the regression task on appraisals might
be easier for dimensions that take on a varied range of values in the
training data. By contrast, in the classification setup, the gap
between automatic and human performances characterize dimensions whose
original judgments concentrate on either end of the rating spectrum,
like \externSocialStandards, \situationalControl, \situationalResp,
\internalStandards, \ownControl and \familiarity, all surpassing the
judges by more than 20pp.

Hence, both the classifier and the regressor outdo \TAhumanText, that
we hypothesized to represent an upper bound for their
performance. They grasp some information about the writers'
perspective that an aggregation of validators does not account
for. This is a hint at the value of collecting judgments directly from
the event experiencers, as a more appropriate source for the systems to
learn first-hand evaluations -- past NLP research did
that through the readers' reconstructions alone.  Therefore, from
Experiment~1, we conclude that the task of inferring the 21 appraisal
dimensions from a text that describes an event is viable: the systems can
model both emotional and non-emotional states using fine-grained,
dimensional information rather than emotion classes. Their
classification performance also improves upon past work based on a
smaller set of appraisal variables (i.e., \citet{Hofmann2020} obtained
\F=.70 on a different data set, labeled only by readers).

\begin{table}
    \centering\small
    \caption{Models' performance (Precision, Recall, \F) on recognizing emotions,
      having access only to the annotated (\AEmodelGoldText) or predicted
      appraisal dimensions (\AEmodelPredText). Standard deviations are reported in Table~\ref{tab:table-appraisals2emotions-sd} in the Appendix. Discretized/Scaled: input appraisals
      are represented as boolean values/as values in the [0:1] interval. $\Delta$: \F difference between the models leveraging appraisal predictions and gold ratings.}
    \label{tab:appraisal2emotion}
    \setlength{\tabcolsep}{5.4pt}
\begin{tabular}{lrrrrrrrrrrrrrr}
    \toprule
     & \multicolumn{7}{c}{Discretized}  & \multicolumn{7}{c}{Scaled} \\
      \cmidrule(r){2-8} \cmidrule(r){9-15}
& \multicolumn{3}{c}{\AEmodelGold}& \multicolumn{3}{c}{\AEmodelPred}& & \multicolumn{3}{c}{\AEmodelGold}& \multicolumn{3}{c}{\AEmodelPred}& \\
    \cmidrule(r){2-4}\cmidrule(r){5-7}\cmidrule(r){9-11}\cmidrule(r){12-14}
    Emotion & P & R & \F & P & R & \F & $\Delta_{\small{\textsf{\F}}}$ & P & R & \F & P & R & \F &$\Delta_{\small{\textsf{\F}}}$\\
    \cmidrule(r){1-1}
    \cmidrule(r){2-4}\cmidrule(r){5-7}\cmidrule(r){8-8}\cmidrule(r){9-11}\cmidrule(r){12-14}\cmidrule(r){15-15}
Anger & $.35$ & $.46$ & $.40$ & $.35$ & $.33$ & $.34$ & $-.06$ & $.35$ & $.46$ & $.40$ & $.28$ & $.57$ & $.37$ & $-.03$\\
    Boredom & $.44$ & $.47$ & $.46$ & $.54$ & $.69$ & $.60$ & $+.14$ & $.47$ & $.62$ & $.54$ & $.46$ & $.60$ & $.52$ & $-.02$\\
    Disgust & $.36$ & $.32$ & $.34$ & $.42$ & $.33$ & $.37$ & $+.03$ & $.49$ & $.44$ & $.46$ & $.58$ & $.20$ & $.29$ & $-.17$\\
    Fear & $.22$ & $.33$ & $.26$ & $.25$ & $.47$ & $.32$ & $+.06$ & $.27$ & $.34$ & $.30$ & $.26$ & $.46$ & $.33$ & $+.03$\\
    Guilt & $.30$ & $.23$ & $.26$ & $.25$ & $.08$ & $.12$ & $-.14$ & $.32$ & $.26$ & $.28$ & $.29$ & $.15$ & $.19$ & $-.09$\\
    Joy & $.28$ & $.30$ & $.28$ & $.31$ & $.31$ & $.30$ & $+.02$ & $.29$ & $.30$ & $.29$ & $.31$ & $.24$ & $.25$ & $-.04$\\
    No-emo. & $.46$ & $.46$ & $.46$ & $.46$ & $.29$ & $.35$ & $-.11$ & $.50$ & $.46$ & $.47$ & $.53$ & $.23$ & $.31$ & $-.16$\\
    Pride & $.33$ & $.39$ & $.35$ & $.27$ & $.48$ & $.34$ & $-.01$ & $.35$ & $.38$ & $.35$ & $.29$ & $.33$ & $.29$ & $-.06$\\
    Relief & $.28$ & $.13$ & $.18$ & $.33$ & $.13$ & $.19$ & $+.01$ & $.32$ & $.18$ & $.23$ & $.36$ & $.21$ & $.26$ & $+.03$\\
    Sadness & $.31$ & $.29$ & $.30$ & $.30$ & $.39$ & $.34$ & $+.04$ & $.37$ & $.35$ & $.36$ & $.36$ & $.24$ & $.28$ & $-.08$\\
    Shame & $.26$ & $.22$ & $.24$ & $.25$ & $.19$ & $.21$ & $-.03$ & $.27$ & $.24$ & $.25$ & $.29$ & $.37$ & $.33$ & $+.08$\\
    Surprise & $.44$ & $.44$ & $.43$ & $.46$ & $.44$ & $.44$ & $+.01$ & $.46$ & $.43$ & $.44$ & $.63$ & $.22$ & $.31$ & $-.13$\\
    Trust & $.21$ & $.14$ & $.17$ & $.29$ & $.10$ & $.15$ & $-.02$ & $.30$ & $.23$ & $.26$ & $.23$ & $.35$ & $.27$ & $+.01$\\
    \cmidrule(r){1-1}\cmidrule(r){2-4}\cmidrule(r){5-7}\cmidrule(r){8-8}\cmidrule(r){9-11}\cmidrule(r){12-14}\cmidrule(r){15-15}
    Macro avg. & $.33$ & $.32$ & $.32$ & $.34$ & $.32$ & $.32$ & $+.00$ & $.37$ & $.36$ & $.35$ & $.38$ & $.32$ & $.31$ & $-.05$\\
    \bottomrule
\end{tabular}
\end{table}

\subsubsection{Experiment 2: Reconstruction of emotions from appraisals}
\label{sec:exp2}
Using the systems above, which go from text to appraisals, allows us
to characterize emotional contents without predefining their possible
discrete values (\anger, \disgust, etc.).  With the second experiment,
we move our attention to such values, which are the phenomena of
interest \textit{par excellence} for emotion analysis.  Our goal is to
investigate the link between the 21 cognitive dimensions and the
recognition of emotions only from a computational perspective, thus
verifying if the appraisal-to-emotion mapping is feasible: we analyze
models that take appraisals as inputs and produce a discrete emotion
label as an output.  Specifically, we represent events either with the
self-reported appraisals (for \AEmodelGoldText) or with the appraisals
predicted by \TAmodelText (for \AEmodelPredText) from Experiment 1.
In both cases, the emotion classifiers do not have (direct) access to
the text.

Results are in Table~\ref{tab:appraisal2emotion}, separated between
the setting where appraisals are booleans, and one where they are
treated as continuous values (scaled within the interval [0:1]).
Surprisingly, the gold appraisals do not systematically yield better
performance than the predicted dimensions. In fact, by looking at the
Discretized framework, we find that each type of input enhances the
detection of different emotions. For instance, \guilt, \anger, \shame,
and \trust are better identified when the gold appraisal ratings are
available to the model, while the recognition of \boredom is
facilitated by predicted appraisal values.  Still, on a macro-level
numbers indicate no discrepancy between the accesses to the gold
annotations and to the output produced by \TAmodelText. The two models
perform on par ($\Delta_{\small{\textsf{\F}}}=0$), with an average
macro-\F of .32.  This finding is per se promising, as it sets the
ground for appraisal-based emotion classifiers that operate
independently of the help of gold information, without producing
worse-quality output.
Differences between the gold and predicted inputs are more marked in
the Scaled setup.  The gold variants lead to a better performance than
the predicted appraisal scores (macro-\F=.35 vs.\ macro-\F=.31), with
\disgust, \surprise, and \pride benefitting the most from such
information.  Compared to the Discretized results, the
\AEmodelPredText here reaches a 1pp-lower \F.\footnote{We experimented
  with the appraisal representation in [1:5] instead of scaling them
  to [0:1], and we obtained an overall macro-\F=.31 for
  \AEmodelGoldText and a macro-\F=.22 for \AEmodelPredText, thus
  $\Delta$=.09.}

Focusing on the emotion differences within \AEmodelPredText, we see a
remarkable gap of 48pp between the lowest and highest \F (33pp in the
Scaled scenario): the information learned by the model is
substantially more useful for a subset of emotions, which suggests
that appraisals might not come equally handy in classifying all
events. At the same time, we acknowledge that all obtained \F scores
seem tepid, irrespective of the input representation and the input
type.  Our results should be interpreted by taking into account that a
random decision in the scaled setting leads to .08 \F, and that
similar performances can be found in psychological studies that
predict emotion from appraisals (with the difference that they report
accuracy instead of \F). \namecite{Smith1985} achieve an accuracy of
42\,\% for the task of classifying 15 emotions based on 6 appraisal
variables.  \namecite{Frijda1989} have an accuracy of 32\,\% in
recognizing 32 emotions using 19 appraisals, \namecite{Scherer1997}
report a score of 39\% in discriminating between 7 emotions using 8
input dimensions, and \namecite{Israel2019} obtain an overall accuracy
of 27\% when recognizing 13 emotions with 25 appraisals.  Thus, the
data-derived mapping from appraisals to emotions aligns with past
research. This is an important indicator for the quality of our data:
the ratings in \corpusname are comparable to those collected in the
past by experts who did not conduct their studies via
crowdsourcing. In practice, this means that our models can exploit the
link between appraisal variables and emotions similarly well as found
in psychology.

\medskip\noindent To understand if the performance of the emotion
prediction based on appraisals is promising for joint models that also
consider text, we now compare the predictive power of
\AEmodelGoldText and \AEmodelPredText against those based on text and
text only.  Such a comparison provides a partial answer to RQ4,
because it shows if the appraisal-based systems capture information
that the text-based models (typically used in emotion analysis)
cannot, and vice versa.  Table~\ref{tab:text2emotion} shows the
results. It summarizes how humans reconstruct the prompting emotion
labels (\TEhumanText), and how automatic systems carry out the same
task (\TEmodelText).  Among the positive emotions, \joy seems the most
difficult to recognize for the system ($.45$\,\F ), which achieves
$.63$ and $.74$\,\F on \relief and \trust, respectively.  The lowest
automatic performance on negative emotions regards those with fewer
annotated samples, i.e., \shame ($.51$\,\F) and \guilt ($.48$\,\F).
Classes that are predicted better by the computational model than by
human validators are \boredom, \disgust, \shame, \surprise and \trust,
as well as \noemotion. It should be noted here that correctly
recognizing \noemotion is challenging, as many participants reported
events in which they remained apathetic but which are typically
emotional (e.g., the loss of a dear person).

The relative improvement of the systems over the validators is less
pronounced than in Experiment~1 but is still present: the system
surpasses humans by 3pp (Macro-\F=.56 for the human validators vs.\
.59\,\F for \TEmodelText). We take it as evidence of the success of
the automatic models, but also of the subjective nature of the emotion
recognition task: even when aggregated into the most representative
judgment of the crowd, the readers' annotation does not necessarily
correspond to the original emotion experience (as spelled out in
Section~\ref{sec:qualitativedataanalysis}, some of their
misclassifications happen between similar emotion classes).

Importantly, the results of \TEmodelText are substantially higher than
what we achieved with the appraisal-informed classification of
\AEmodelPredText and \AEmodelGoldText. Some \F scores of the predicted
appraisal-based model are on par with the textual classification,
either with \TEmodelText (e.g., \boredom, \surprise) or \TEhumanText
(e.g., \noemotion, \surprise).  However, overall numbers clearly
pinpoint to the conclusion that appraisals alone do not bear an
advantage over the textual representations of events. This is
unsurprising, because appraisals are grounded in (and in fact stem
from) a salient experience, while the two models under consideration
are aware of \textit{how} a circumstance is evaluated but not
\textit{what} circumstance is evaluated, i.e., the opposite of
\TEmodelText.  Therefore, as we move on to the next experiment, we
contextualize appraisals with textual information, to understand if
they can complement each other.

\begin{table}
  \newcommand{\tripcol}[1]{\multicolumn{3}{c}{#1}}
    \centering\small
    \caption{Emotion recognition performance (Precision, Recall, \F) of the models based only on text (\TE) or both text and appraisals (\TAE). Standard deviations are reported in Table~\ref{tab:table-emotions-sd} in the Appendix. Delta values show the differences between the Macro-\F scores of the indexed models.}
\label{tab:text2emotion}
    \setlength{\tabcolsep}{3.5pt}
\begin{tabular}{lrrrrrrrrrrrrrrrr} %
  \toprule
  & \tripcol{(a)} & \tripcol{(b)} & & \tripcol{(c)} & &\tripcol{(d)} & & \\
    & \tripcol{\TEhuman} & \tripcol{\TEmodel} & $\Delta^{(b)}_{(a)}$ & \tripcol{\TAEmodelGold} & $\Delta^{(c)}_{(b)}$ &\tripcol{\TAEmodelPred} & $\Delta^{(d)}_{(c)}$ & $\Delta^{(d)}_{(b)}$ \\
    \cmidrule(r){2-4}\cmidrule(r){5-7}\cmidrule(r){8-8}\cmidrule(r){9-11}\cmidrule(r){12-12}\cmidrule(r){13-15}\cmidrule(r){16-16}\cmidrule(r){17-17}
    Emotion & P & R & \F & P & R & \F & \F & P & R & \F & \F & P & R & \F & \F & \F \\
    \cmidrule(r){1-1}\cmidrule(r){2-4}\cmidrule(r){5-7}\cmidrule(r){8-8}\cmidrule(r){9-11}\cmidrule(r){12-12}\cmidrule(r){13-15}\cmidrule(r){16-16}\cmidrule(r){17-17}
    Anger      & $.50$ & $.66$ & $.57$ & $.57$ & $.52$ & $.53$ & $-.04$   & $.56$ & $.58$ & $.57$ & $+.04$  & $.56$ & $.58$ & $.57$ & $.00$  & $+.04$  \\
    Boredom    & $.78$ & $.69$ & $.73$ & $.81$ & $.87$ & $.84$ & $+.11$ & $.83$ & $.84$ & $.83$ & $-.01$   & $.83$ & $.83$ & $.83$ & $.00$  & $-.01$ \\
    Disgust    & $.85$ & $.53$ & $.65$ & $.74$ & $.59$ & $.66$ & $+.01$  & $.70$ & $.63$ & $.66$ & $.00$   & $.70$ & $.63$ & $.66$ & $.00$  & $.00$  \\
    Fear       & $.66$ & $.83$ & $.73$ & $.65$ & $.66$ & $.65$ & $-.08$   & $.69$ & $.66$ & $.67$ & $+.02$  & $.69$ & $.66$ & $.67$ & $.00$  & $+.02$  \\
    Guilt      & $.48$ & $.58$ & $.53$ & $.63$ & $.39$ & $.48$ & $-.05$   & $.64$ & $.54$ & $.58$ & $+.10$ & $.63$ & $.52$ & $.56$ & $-.02$  & $+.08$  \\
    Joy        & $.41$ & $.62$ & $.49$ & $.53$ & $.40$ & $.45$ & $-.04$   & $.49$ & $.48$ & $.48$ & $+.03$  & $.49$ & $.46$ & $.47$ & $-.01$  & $+.02$  \\
    No-emotion & $.72$ & $.21$ & $.33$ & $.66$ & $.50$ & $.55$ & $+.22$ & $.61$ & $.54$ & $.56$ & $+.01$  & $.62$ & $.53$ & $.56$ & $.00$  & $+.01$  \\
    Pride      & $.52$ & $.69$ & $.59$ & $.48$ & $.64$ & $.54$ & $-.05$   & $.51$ & $.61$ & $.55$ & $+.01$  & $.50$ & $.62$ & $.55$ & $.00$  & $+.01$  \\
    Relief     & $.56$ & $.74$ & $.64$ & $.65$ & $.63$ & $.63$ & $-.01$   & $.58$ & $.67$ & $.62$ & $-.01$   & $.58$ & $.68$ & $.62$ & $.00$  & $-.01$ \\
    Sadness    & $.54$ & $.76$ & $.63$ & $.52$ & $.68$ & $.59$ & $-.04$   & $.61$ & $.69$ & $.65$ & $+.06$  & $.59$ & $.69$ & $.63$ & $-.02$  & $+.04$  \\
    Shame      & $.48$ & $.48$ & $.48$ & $.53$ & $.50$ & $.51$ & $+.03$  & $.55$ & $.47$ & $.50$ & $-.01$   & $.55$ & $.45$ & $.49$ & $-.01$  & $-.02$ \\
    Surprise   & $.57$ & $.33$ & $.42$ & $.53$ & $.54$ & $.53$ & $+.11$ & $.58$ & $.44$ & $.49$ & $-.04$   & $.58$ & $.44$ & $.50$ & $+.01$ & $-.03$ \\
    Trust      & $.95$ & $.36$ & $.52$ & $.73$ & $.75$ & $.74$ & $+.22$ & $.76$ & $.71$ & $.73$ & $-.01$   & $.76$ & $.70$ & $.72$ & $-.01$  & $-.02$ \\
    \cmidrule(r){1-1}\cmidrule(l){2-4}\cmidrule(l){5-7}\cmidrule(l){8-8}
    \cmidrule(r){9-11}\cmidrule(r){12-12}\cmidrule(r){13-15}\cmidrule(r){16-16}\cmidrule(r){17-17}
    Macro \avg & $.62$ & $.58$ & $.56$ & $.62$ & $.59$ & $.59$ & $+.03$  & $.62$ & $.60$ & $.61$ & $+.02$  & $.62$ & $.60$ & $.60$ & $-.01$  & $+.01$  \\
  \bottomrule
\end{tabular}
\end{table}

\subsubsection{Experiment 3: Reconstruction of emotions via text and appraisals}
\label{sec:exp-3}
We gauge understanding of the extent to which appraisals
contribute to emotion classification by comparing models that
have access to both the text and the (predicted or original) appraisals with
the automatic emotion predictor based solely on the text.

Columns (c) and (d) in Table~\ref{tab:text2emotion} show the results
of the pipelines \TAEmodelGoldText and \TAEmodelPredText. Column
$\Delta^{(c)}_{(b)}$ shows the improvement of the pipeline that
integrates text and appraisal information.  The current experiment
returns a different picture than Experiment~2.  Here we see that
appraisals \textit{enhance} emotion classification to various degrees
for the different emotions.  Overall, they allow the model to gain 2pp
\F. While this might seem a minor improvement, for some classes the
increase is more substantial, namely for \guilt, \sadness, \anger and
\joy (+10pp, +6pp, +4pp and +3pp, respectively). This amelioration
mostly stems from an increased recall, i.e., finding
emotions with the help of appraisals seems easier. Only for some emotions there is a drop in
\F, particularly for \surprise ($-$4pp).

The fact that this model relies on gold appraisal information
represents a principled issue, because gold appraisals are typically
not available in classification scenarios. Therefore, as a last
analysis, we examine \TAEmodelPredText which replaces the writers'
ratings with predicted values.  We use the \TAmodelText regressor
trained in Experiment~1 and remap the continuous values that they
produce in the [0:1] interval back to the 5-point scale used by the
human annotators.\footnote{We remap to discrete values both for a
  direct comparison to the original appraisals and because
  Experiment~2 showed that such framework works better than
  the scaled alternative, although to a minimal extent.} We observe
that the performance remains consistent with the gold-aware systems
(Macro-\F is only 1pp lower), in line with our previous finding that
leveraging appraisal predictions as inputs is not detrimental for the
overall emotion recognition task, and that the benefit is more
substantial for some emotions than others.

Also the result that these cognitively-motivated features do not improve the
reconstruction of some emotion categories (e.g., \disgust, \relief) is coherent
with a finding that emerged earlier (Experiment~2): appraisals might not be
equally handy to classify all events. This suggests that, at times, a text
might contain sufficient signal to support an appropriate classification
decision. After all, we have observed that appraisals themselves are fairly
``contained'' in text, as they can be predicted. Thus, one could argue that
text alone is informative enough, but the help of appraisals becomes evident
with other emotion categories (\guilt, \sadness, \anger and \joy). Hence, there
are cases in which exploiting them as input features (standing as explicit
background knowledge involved in the text interpretation) pushes the
classifiers towards the correct inference. Brief, to answer RQ4, we find that
the integration of appraisals into an emotion classifier can have a (partial
but) beneficial impact on its observed performance.

\begin{wraptable}{R}{0.3\textwidth}
\centering\small
\caption{Appraisal contribution to the emotion classification of texts
  with high/low appr.\ IAA. Scores are \F.}
\label{tab:apprcontr}
\setlength{\tabcolsep}{3pt}
\begin{tabular}{llrr}
  \toprule
  && \multicolumn{2}{c}{Agreement} \\
  \cmidrule(l){3-4}
  && High & Low \\
  \cmidrule(lr){3-3}\cmidrule(lr){4-4}
  \multirow{2}{*}{\rotatebox{90}{Appr.}}
  & w/o. & .60 & .54 \\
  & w.   & .62 & .57 \\
  \bottomrule
\end{tabular}
\end{wraptable}
The role of appraisal dimensions can be better appreciated by discussing difficult-to-judge
event descriptions.  In the analysis of
Table~\ref{tab:agreementexamples1} and~\ref{tab:agreementexamples2},
Section~\ref{sec:qualitativedataanalysis}, we conjectured that texts
where validators misunderstood the experience of the writers are more
likely to be correctly classified (emotion-wise) by a model that has
access to the (correct) appraisal information.  To test our
hypothesis, we extract 400 instances from the validation set that have
the highest G--V appraisal agreement value, and 400 instances with the
lowest agreement. We evaluate classification with and without
appraisals (\TAEmodelGoldText and \TEmodelText) on these two sets of
instances.

\F scores averaged over five runs of the models are in
Table~\ref{tab:apprcontr}.  The classification performance is lower
for the 400 datapoints on which annotators disagreed the most (column
Low), regardless of the input.  In both agreement groups, the
predictions informed by appraisals lead to superior \F: there is an
improvement of 2pp for instances with high appraisal agreement, and a
higher improvement for those with low appraisal
agreement (+3pp), as expected.

\subsection{Error Analysis}
\label{sec:qualquantanalysis}
While we provided evidence that appraisal predictions help emotion
recognition in some cases, it remains unclear how they help, that is,
whether there are cases in which they systematically improve the
prediction that would be taken without any access to appraisals, and
what types of mistakes they prevent.  To understand when the input
appraisals' access is convenient, we conduct quantitative and
qualitative analyses.

\subsubsection{Quantitative Analysis}
\begin{figure}
  \mbox{}\hspace{30mm}\sf\TEmodel\hspace{41mm}\TAEmodelPred\\
  \raisebox{42mm}{\rotatebox{90}{\sf Gold}}
  \includegraphics[scale=0.95]{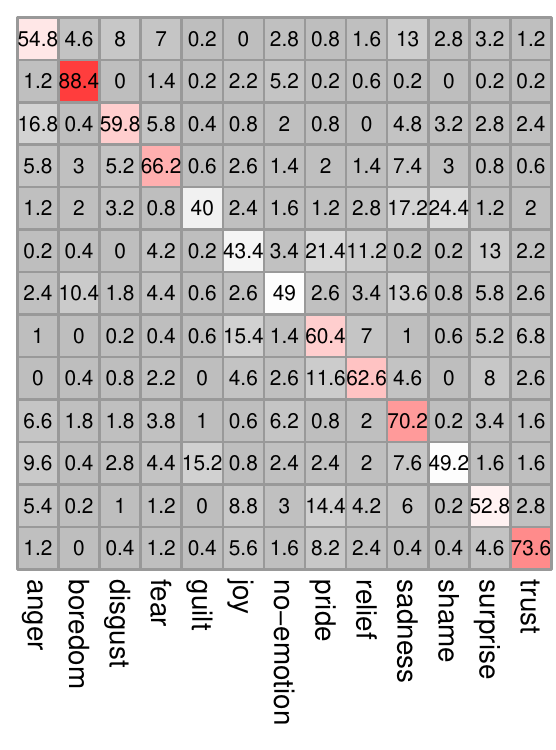}%
  \includegraphics[scale=0.95]{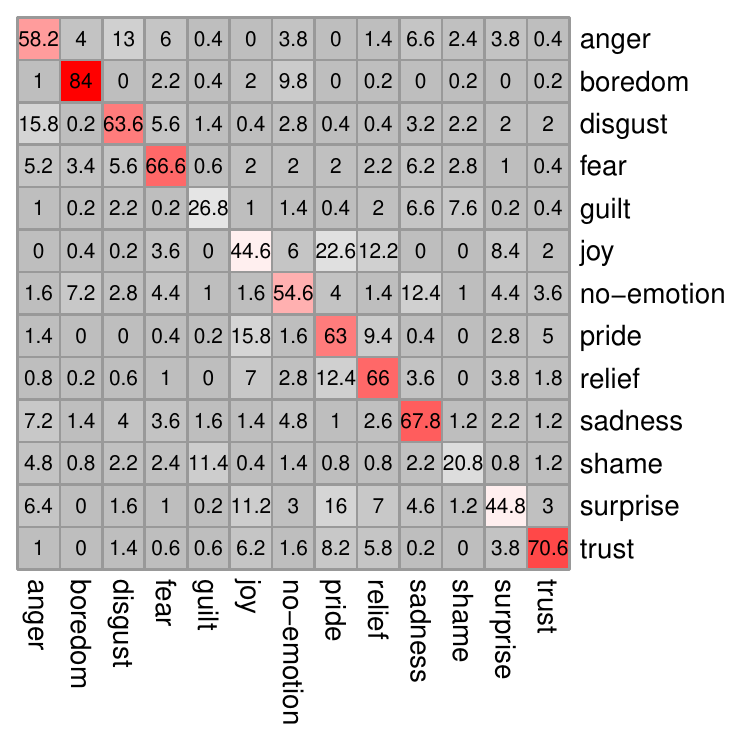}\\
  \mbox{}\hspace{30mm}\sf{}Prediction\hspace{40mm}Prediction\\
  \caption{Confusion matrices to compare \TEmodelText and
    \TAEmodelPredText. The numbers represent counts of correct
    predictions (on diagonal) and mistakes (off-diagonal) averaged
    across 5 runs. Numbers in the guilt and shame rows are multiplied
    by 2 such that they are comparable to the other emotions;
    therefore, the values can also be interpreted as percentages,
    because the number of instances per emotion in the test set is 100.
    Higher values on the diagonal mean better performance; vice versa
    for the off-diagonal cells.}
  \label{fig:heatmaps-models}
\end{figure}
\begin{figure}
  \centering
    \mbox{}\hspace{-10mm}(\TAEmodelPredText) $-$ (\TEmodelText)\\
  \raisebox{42mm}{\rotatebox{90}{\sf Gold}}
  \includegraphics[scale=1]{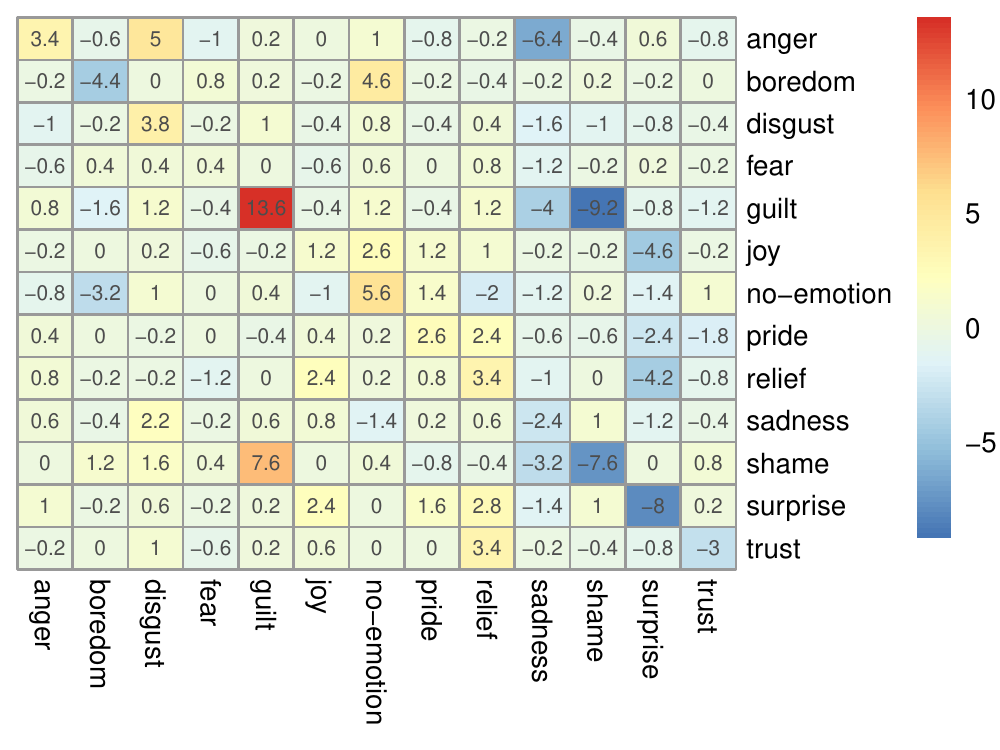}\\
  \mbox{}\hspace{-10mm}\sf{}Prediction\\
  \caption{Cell-by-cell difference between \TAEmodelPredText and
    \TEmodelText in Figure~\ref{fig:heatmaps-models}.  Numbers
    represent the comparison between the count of texts that both of
    them correctly predicted (on diagonal) and those that they both
    misclassified (off-diagonal), averaged across 5 runs of the
    models.  Red cells are those where \TAEmodelPredText performs
    better than \TEmodelText, blue cells show the opposite case. Color
    intensity corresponds to the absolute value of the numbers.}
  \label{fig:heatmap-delta-models}
\end{figure}

The two confusion matrices in Figure~\ref{fig:heatmaps-models} are a
break-down of the performance reported in Table~\ref{tab:text2emotion}
for \TEmodelText and \TAEmodelPredText. They contain the counts of
text labelled correctly (on the diagonal, representing true positives
-- TP) and incorrectly (off-diagonal), averaged across five runs of
the models. Note that the values for the emotions of \guilt and
\shame are multiplied by two to simplify the comparison with the other
emotions. These numbers show what emotion pairs are better
disambiguated through the knowledge of the 21 cognitive variables, and
what pairs, on the contrary, suffer from it.  We summarize the
difference between them in Figure~\ref{fig:heatmap-delta-models}.

We have already seen that predictions of \anger, \fear, \guilt, \joy,
and \sadness benefit particularly from appraisal features
(Table~\ref{tab:text2emotion}).  The comparison of the diagonals in
the two heatmaps mirrors the improvements across such labels:
\TAEmodelPredText predicts on average 6.8 \guilt TP (13.6/2) more than
\TEmodelText; the count of TP of \anger increases of 3.4; for \joy
there are 1.2 more TP, while for \fear 0.4.  For \sadness, the
improvement in \F cannot be found in the number of TP instances, which
in fact decreases by 2.4. It rather stems from a reduction in false
positives (off-diagonal sum in the \sadness columns: $63.6$ for
\TEmodelText and $46$ for \TAEmodelPredText, which is mostly due to a
better disambiguation of \sadness from \anger).
We also notice that the correct and incorrect predictions of the two
models are distributed unevenly across the 13 emotions. Emotion pairs
that are most often confused by \TEmodelText are (gold/predicted)
\disgust/\anger (16.8 FP), \noemotion/\boredom (10.4), \pride/\joy
(15.4), \guilt/\shame (12.2 = 24.4/2 in the matrix), \relief/\pride
(11.6) and \joy/\relief (11.2). Appraisal information in fact slightly
adds confusion to these particularly challenging classes (\disgust/\anger: 1, \noemotion/\boredom: 3.2,
\pride/\joy: 0.4, \relief/\pride: 0.8, \joy/\relief: 1), except for
\guilt/\shame where confusion declines by 4.6 (9.2/2) FP.

\subsubsection{Qualitative Analysis}
As a last analysis, we manually inspect the interaction between
appraisals and emotion predictions. We show here 20 examples of texts
whose classification is modified for the better by specific appraisal
information (Table~\ref{tab:agreementexamplesmodels}), selected based
on the agreement reached by the annotators.

\begin{table}[t]
  \centering\small
  \setlength{\tabcolsep}{2pt}
  \renewcommand{\arraystretch}{0.9}
  \caption{Examples in which the \TEmodelText is corrected by the
    \TAEmodelPredText. The top (bottom) part of the table shows the
    examples where the G--V agreement is high (low) on appraisal
    evaluations ($\textrm{RMSE}$).}
  \label{tab:agreementexamplesmodels}
  \begin{tabularx}{1.0\linewidth}{lccccX}
    \toprule
      Id & Gold & \TEmodel & \TAEmodelPred & $\textrm{RMSE}$ & Text \\
      \cmidrule(r){1-1}\cmidrule(lr){2-2}\cmidrule(lr){3-3}\cmidrule(lr){4-4}\cmidrule(lr){5-5}\cmidrule(l){6-6}

      1 & fear & sadness & fear & $1.02$ & \textit{When I found out my mum had cancer} \\
      2 & pride & surprise  & pride & $1.04$ & \textit{I got my degree} \\
      3 & relief & trust & relief & $1.04$ & \textit{When my child settled well into school} \\
      4 & disgust & surprise & disgust & $1.08$ & \textit{someone dropped meat on the floor at work and used it.} \\
      5 & no-e. & boredom & no-e. & $1.15$ & \textit{travelling to Cooktown Queensland}\\
      6 & anger  & anger & disgust & $1.15$ & \textit{I felt ... when my partner waited to tell me 3 months later that he had texted his ex-partners.} \\
      7 & pride & joy & pride & $1.26$ & \textit{I bought my car recently}\\
      8 & shame & guilt & shame & $1.27$ & \textit{broke an expensive item in a shop accidently}\\
      9 & relief & surprise & relief & $1.28$ & \textit{I'm supposed to speak publicly but the event gets cancelled.} \\
      10 & sadness & surprise & sadness & 1.29 & \textit{I found out that my ex-wife was divorcing me.}\\
      \cmidrule(r){1-1}\cmidrule(lr){2-2}\cmidrule(lr){3-3}\cmidrule(lr){4-4}\cmidrule(lr){5-5}\cmidrule(l){6-6}
      60 & anger & trust & anger & $1.36$ & \textit{someone moved my personal belongings}\\
      61 & anger & shame & anger & $1.40$ &  \textit{my mother made me feel like a child}\\
      62 & anger & sadness & anger & $1.41$ & \textit{I was lied to about money} \\
      63 & anger & sadness  & anger & $1.47$ & \textit{when youths dont respect their elders} \\
      64 & guilt & sadness & guilt & $1.53$ & \textit{I ate some food from the fridge which belonged to my flatmate without her permission} \\
      65 & relief & pride & relief & $1.54$ & \textit{I passed my Irish language test}\\
      66 & no-e. & relief & no-e. & $1.52$ & \textit{when getting my roof inspected for storm or wind damage.} \\
      67 &  relief & joy & relief & $1.67$ & \textit{When I found my dog}\\
      68 & no-e. & boredom & no-e. & $1.67$ & \textit{Completing my degree. Should have felt pride, didn't feel ... but a headache.} \\
      69 & guilt & shame & guilt & $1.70$ & \textit{I took the last shirt in the right size when my friend wanted it too.} \\
      70 & joy & surprise & joy & $1.73$ & \textit{When I received a invite to a wedding}\\
      71 & disgust & pride & disgust & $2.02$ & \textit{His toenails where massive} \\
      \bottomrule

  \end{tabularx}
\end{table}

The appraisal-aware model rectifies the example corresponding to Id 64
(``I ate some food from the fridge which belonged to my flatmate
without her permission'') from \sadness to \guilt.  The dimensions
relevant for disentangling these two emotions are \pleasantness,
\unpleasantness and \situationalControl \cite{Smith1985}. The
classification improvement here correlates precisely with a low
\situationalControl ($1$) and moderate \pleasantness ($3$) -- a common
appraisal association of an event annotated for \sadness has low
values.  Also \ownresponsibility ($5$) and moderate \ownControl ($3$)
might have played a role.  We see a similar pattern for example 69, initially
associated with \shame but corrected to \guilt with the dimensions
related to the perception of one's agency.

Example 61 (``my mother made me feel like a child'') shows how \anger
is disambiguated from \shame. There, a score of $4$ is predicted for
\otherResp, then used as input. This makes intuitive sense: the
dimension is typical of events in which the experiencer undergoes what
someone else has caused (``my mother''), and in fact, it
differentiates \anger from other negative emotions such as \shame and
\guilt (where the responsibility falls on the self)
\cite{Smith1985}. Anger is further characterized by patterns in
\ownControl, \othercontrol, with the former being lower than the
latter \cite{Smith1985,Scherer1997}.  Example 60 (``someone moved my
personal belongings'') further highlights their importance
(\ownControl: $2$, \othercontrol:$5$).

The emotion of \disgust is confused by \TEmodelText with \surprise and
\pride in the examples 4 and 71, respectively.  Once more, these events
are about something caused or belonging to \textit{others}
(accordingly, \otherResp and \othercontrol are rated both as $5$, and
\ownresponsibility as $1$).  The prediction of \notconsider in the
lower end of the rating scale might have been taken as a further
indicator of the negative connotation of the text by \TAEmodelPredText.
\textit{Suddenness}, \urgent, and \goal, that the theories correlate
to \fear \cite{Scherer2001}, stand out in ``When I found out my mum
had cancer'' (where they are all rated as $5$).  Further, the correct
prediction for example 7 (``I bought my car recently'') and example 2
(`` I got my degree'') is accompanied by strong degrees of appraisals
that are typical of \pride{} -- \ownControl ($4$ and $2$),
\ownresponsibility ($5$ and $5$), \goal ($5$ and $5$), and
\goalSupport ($5$ and $5$).

In general, for positive emotions such as \relief, \trust, \surprise,
and \pride, it is more difficult to identify patterns of appraisals
that differentiate them, and even annotators disagree more. This
could be due to our set of 21 variables, which does not include some
dimensions recently proposed to tackle positive
emotions specifically \cite{Smith2020}.

Another class for which \TAEmodelPredText tends to recover the correct
label is \noemotion, that \TEmodelText mistakes for \boredom
(examples 5 and 68) and \relief (example 66).  All of them can be
thought of as non-activating states, but the confusion with \boredom
is especially foreseeable, in that its low motivational relevance
(i.e., \goal), \pleasantness and \unpleasantness
\cite{Smith1985,Yih2020} is shared by the neutral state of \noemotion.
Example 5 (``travelling to Cooktown Queensland'') partially confirms this
pattern, as \goal and \unpleasantness are rated by the model as $1$
(but \pleasantness as $4$).

\section{Discussion and Conclusion}
\label{conclusion}
\paragraph{Contributions and Summary} This paper is concerned with
appraisal theories, and investigates the representation of appraisal
variables as a useful tool for NLP-based emotion analysis: starting
from the collection of thousands of event descriptions in English, it
conducts a detailed analysis of the data, it discusses its annotations
from the writers' and readers' perspectives, and lastly, it describes
experiments to predict emotions and appraisals, separately and
jointly.

We propose the use of 21 appraisal dimensions based on an extensive
discussion of theories from psychology. Appraisals formalize criteria
with which humans evaluate events. As such, they are cognitive
dimensions that underly emotion episodes in real life -- a type of
information that can facilitate systems in interpreting implicit
expressions of emotional content.  They also allow to represent the
structured differences among the phenomena in question.  Nevertheless,
they are mostly dismissed in the literature of affective computing in
text. We provide evidence that their patterns can be leveraged to
represent emotions and are beneficial for the modeling of specific
classes. In fact,
under the assumption that appraisals \textit{are} emotions, modeling
can take place without the need to decide on a set of emotion labels
in advance.  A process of this type is similar to the use of valence
and arousal among studies in the field based on dimensional models
from psychology. At the same time appraisals are a formalism with a
stronger expressive power, as they can differentiate emotion
categories via more fine-grained underlying mechanisms \cite{Smith1985},
have a theoretically-motivated mapping to
emotions, and fit the analysis of events from the perspective of the
people who lived through them. In contrast, valence and arousal models
focus on affect, which is more related to a subjective feeling than a
cognitive processing module.

Our appraisal labels are the result of a crowdsourcing study.
Participants were tasked to describe events that provoked a specific
emotion in them; further, they qualified their
experience along the 21 appraisals. This gold standard data served as
a basis for an evaluation of other human annotators: being presented
with (a subset of) the event descriptions, readers had to
recover both the original emotion and the original appraisals. In
turn, their judgment served for comparison with multiple models,
aimed at determining if the task of appraisal prediction is feasible,
and how such predictions can be exploited for the automatic detection
of emotion from text.  Validators and systems turned out to perform
similarly on the task of emotion and appraisal prediction. Therefore,
we conclude that text provides information for humans and
classifiers to recover appraisals (RQ1).

It is noteworthy that the readers agree to a higher extent with other
readers on the appraisal assignment than with the texts'
authors. Based on qualitative analyses, we exemplified the
correspondence between textual realizations and appraisal ratings
(RQ2) rated by both systems and humans, highlighting how certain
texts have a more typical emotion connotation, while others require
more elaborate interpretation (e.g., by focusing on different parts of
the texts, different appraisals might fit a description).  In most
cases, the descriptions we collected allow for an event assessment
which is faithful to the original one.
From a quantitative angle, we found a significant relation between validators' traits and their
reliability. Differences between the annotation
conducted by readers with dissimilar traits are, however, small (RQ3). We thus
deduce that appraisals can be annotated in
traditional annotation setups, just like emotions.  Finally, we saw that
appraisals help to predict certain emotion categories, as they correct
mistakes of a system relying on text alone (RQ4).
Overall, appraisal theories proved to be a valid
framework for further research into the modeling of emotions in text.

We make \corpusname publicly available.  Of the 6600 descriptions,
1200 instances are also labeled from the readers'
perspective. Further, we prepare our implementation for future use
and will make it available as easy-to-use pretrained models, to
facilitate upcoming research on the generalizability of appraisals in
other textual domains. \corpusname
includes variables that have not all been fully analyzed in this
paper. This brings us to future work.

\paragraph{Future Work} Our analyses of the data, inter-annotator
agreement and models raise a set of important future work items.
First, we tackled the impact of appraisals on the resolution of
misclassification. With a manual analysis, we interpreted the
differences between the models' behaviors by attempting to match the
predicted appraisal patterns to the patterns documented by the theories.
Their correspondence indicates that appraisals lend themselves well as a tool
to introspect and explain machine learning models, but without a robust, quantitative
approach to the problem, which goes beyond the scope of this paper,
our investigation has only scratched the surface of their potential to
explain emotion decisions.

The patterns identified in the qualitative discussion support the idea
that specific dimensions disambiguate emotions in different cases,
depending on the topic/event in question.  This puts forward another
promising research direction, namely, emotion prediction conditioned
on particular interpretations of events.  Understanding if and when
some appraisals have a systematic effect on a classifier's predictions
would have a valuable application: empathetic dialogue agents could
grasp internal states better by asking users to clarify the relevant
evaluation dimensions (e.g., ``did you feel responsible for the
fact?'', ``could you foresee its consequences?''). In addition, we
only made use of one appraisal vector for modeling (i.e., that
representing or being predicted for the perspective of writers).  Can
we build person-specific emotion and appraisal predictors guided by
demographic properties, personality traits, or current emotion state?
Although we did not find any evidence that personal attributes
influence inter-annotator agreement, it is possible that incorporating
this information in models might make their inferences more fitting to
the expectations of users.

We highlighted the slight but consistent mismatch between humans
and machine learning models. The latter performs better, but strictly
speaking, the two did not undertake the same task: the models were
trained on the writers' perspective, while the readers attempted to
minimize the distance between their own point of view (based on prior
emotion experiences and subjective interpretations) and that of some
unknown text author. A fairer comparison would adopt zero-shot
learning, for instance with natural language inference models or
transformers trained for text generation.

The corpus we collected gives the opportunity
to analyze what lies behind a particular emotion choice.
Can we predict/explain the variations of emotion assignments from validators
with the help of their appraisals?
We found that even when they do
not recover the gold emotion label, they can still be correct about appraisals.
This motivates an adaption of the used measures of inter-annotator agreement
towards an account of the fundamentally similar understanding of texts:
emotion disagreements that come hand in
hand with high appraisal agreement could be weighted as
less relevant. As an alternative,
future work could study if wrong emotion judgments
are considered valid by the writers themselves, by extending the corpus construction
task to a multi-label scenario, where the writers indicate secondary
emotions that are acceptable interpretations of their experiences.

While we focused on English,
our corpus construction procedure can easily be transferred to other
languages and scaled to larger amounts of
texts. Given the finding that the readers' annotation
is reliable, similar data can be collected for other languages,
for specific domains, and going beyond event descriptions
induced experimentally -- an endeavor that has recently taken its first steps
towards verbal productions extracted from social media
\citep{Stranisci2022}. Our expectation is that
the full value of appraisal information in emotion-laden data
will flourish with more spontaneously produced and (ideally)
longer pieces of texts, which can give both human annotators
and classifiers more context to picture the evaluation
stage of an affective episode. Moving to different
domains would also be important to verify if appraisals
promote the recognition of a handful of emotion classes as in our work, or
if our results are an effect of the events described by the writers, and actually,
in other texts many more emotion classes can be better differentiated
through explicit appraisal criteria.

Lastly, appraisals encompass a range of experiences, which they can
account for from various perspectives, including those of the entities
mentioned in text \citep{Troiano2022}. This makes them advantageous
for other studies than emotion modeling, interested in understanding
human judgments more broadly, like argumentative persuasion, analyses
of evaluations from text, and streams of research aimed explaining
their models in a cognitively-motivated manner.

\paragraph{Ethical Considerations}

The task of recognizing appraisals (and emotion categories) is and
will be imperfect.  As \citet{mohammad2022ethics} puts it: ``it is
impossible to capture the full emotional experience of a person
[...]. A less ambitious goal is to infer some aspects of one’s
emotional state''. This applies to our work as well.  The taxonomy of
21 appraisal criteria contains a structured and useful guideline to
investigate certain evaluations involved in humans' affective
reactions. We praised their expressive advantage over the feeling and
motivational traditions. Still, they are not sheltered from criticism
\cite{rosemanappraisal}. For instance, event evaluations are in principle
countless; it might also be doubted that an appraisal, 
or the group of appraisal variables as a whole, is always
sufficient and/or necessary for an emotion to happen, and
consequently, that is always an appropriate approach for computational analyses.

We publish the raw, unaggregated judgments
to account for the naturally diverse emotion recognition sensibilities
of our validators, who ended up producing many
interpretations for the same texts (with the extreme case of
the descriptions produced for $E=\text{\noemotion}$, in which the validators
could read an emotional reaction). Allowing readers to participate 
only once was our strategy to collect divergent voices, precisely. 
For a similar reason, we encouraged variety among the descriptions of
events in the generation phase of \corpusname.

Other than linguistic diversity and disagreeing annotations,
\corpusname displays a rich range of demographics made publicly
available.  Nevertheless, we do not see any particular risk regarding
the profiling of our participants.  First, we pseudonymize their IDs
in respect of their privacy.  Second, for machine learning systems to
learn personal expressive patterns, private affective behaviors, or
personal preferences, a considerable amount of data from the same
person would be needed. Instead, \corpusname has an inconsistent
number of texts coming from different writers, and many of them
produced only one description.  Third, we worked with experimental
texts: while it is reasonable to assume that they represent the
participants' language use, in a more spontaneous occasion people
might have written about other aspects of their life, and might not
necessarily have expressed emotion content by focusing on events.
Fourth, such texts are taken in isolation: within larger textual
contexts, they could be associated to different emotions.

Like other studies in computational emotion analysis, ours endorses
the assumption that language is a window on people's mental lives.
As such, it favours human-assisting applications (e.g., for 
chatbots in the healthcare domain) but is also prone to misuse
(e.g. to profile people's mental wellbeing and preferences, 
to decide on their everyday lives' opportunities). 
We condemn all future applications of the outcome of our work 
breaching people's privacy or testing their emotional states and appraisals
without consent.

\section*{Acknowledgements}
We thank Kai Sassenberg for support with the formulation of the items
in the questionnaires and general consultation in the area of emotion
theories.  This research is funded by the German Research Council
(DFG), project ``Computational Event Analysis based on Appraisal
Theories for Emotion Analysis'' (CEAT, project number KL 2869/1-2).

\starttwocolumn
\bibliography{lit}

\onecolumnnew

\clearpage

\appendix

\section{Comparison of Appraisal Dimensions Formulations to the Literature}
\label{comparison-appendix}
Table~\ref{tab:comparisontable} reports a comparison of 
the appraisal statements that we used in the generation phase of
\corpusname with the original formulations in
\citet{Scherer1997} and \citet{Smith1985}.  
Our statements were rated from 1 to 5 (with 1 being ``not at all''
and 5 ``extremely''). Similarly, answers for \citet{Scherer1997}
were picked on a 5-point Likert scale between ``not at all'' over
``moderately'' to ``extremely'', with an addition option ``N/A''.
\citet{Smith1985} chose a 11-point scale. 

\newcommand{\horizhead}[1]{\multicolumn{3}{l}{\textbf{#1}}\\*\cmidrule{1-3}}
\newcommand{\verthead}[2]{#2}
\newcommand{\sep}{\cmidrule(r){1-1}\cmidrule(rl){2-2}\cmidrule(rl){3-3}}

{\small
\begin{longtable}{p{2cm}p{6cm}p{5cm}}
  \caption{Comparison of formulations of items between
  \citet{Scherer1997} (SW), \citet{Smith1985} (SE), and our study.}
  \label{tab:comparisontable}\\
  \toprule
  Dim. & SW/SE & \corpusname \\
 \cmidrule(r){1-1} \cmidrule(rl){2-2}\cmidrule(rl){3-3}
  \endfirsthead
  \multicolumn{3}{l}{\ldots continued}\\
  \toprule
  Dim. & SW/SE & \corpusname \\
 \cmidrule(r){1-1} \cmidrule(rl){2-2}\cmidrule(rl){3-3}
  \endhead
  \horizhead{Relevance Detection: Novelty Check}
  Suddenness
       & SW: At the time of experiencing the emotion, did you think that the event happened very suddenly and abruptly?
               &  The event was sudden or abrupt.
  \\\sep
  Familiarity
       & SW: At the time of experiencing the emotion, did you think that  you were familiar with this type of event?
               & The event was familiar.
  \\\sep
  Event Predict\-ability
       & SW: At the time of experiencing the emotion, did you think that  you could have predicted the occurrence of the event?
               & I could have predicted the occurrence of the event.
  \\\sep
 Attention, Attention Removal
       & SE: Think about what was causing you to feel happy in this situation. When you were feeling happy, to what extent did you try to devote your attention to this thing, or divert your attention from it.
               & I paid attention to the situation.\par I tried to shut the situation out of my mind.
  \\\sep
  \horizhead{Relevance Detection: Intrinsic Pleasantness}
  Unpleasantn.\par Pleasantness
       & SW: How would you evaluate this type of event in general, independent of your specific needs and desires in the situation you reported above?\par Pleasantness\par Unpleasentness
               & The event was pleasant for me.\par The event was unpleasant for me.
  \\\sep
  \horizhead{Relevance Detection: Goal Relevance}
  Relevance
       & SW: At the time of experiencing the emotion, did you think that the event would have very important consequences for you?
               & I expected the event to have important consequences for me.
  \\\sep
  \horizhead{Implication Assessment: Causality: agent}
  Own, Others', Situational Responsibility
       & SW: At the time of the event, to what extent did you think that one or more of the following factors caused the event?\par Your own behavior.\par The behavior of one or more other person(s)\par Chance, special circumstances, or natural forces.
               & The event was caused by the my own behavior.\par The event was caused by somebody else’s behavior.\par The event was caused by chance, special circumstances , or natural forces.
  \\\sep
  \horizhead{Implication Assessment: Goal Conduciveness}
  Goal Support
       & SW: At the time of expriencing the emotion, did you think that real or potential consequences of the event...\par ... did or would bring about positive, desirable outcomes for you (e.g., helping you to reach a goal, giving pleasure, or terminating an unpleasant situation)?\par ...did or would bring about negative, undesirable outcomes for you (e.g., preventing you from reaching a goal or satisfying a need, resulting in bodily harm, or producing unpleasant feelings)?
               & At that time I felt that the event had positive consequences for me.
  \\\sep
  \horizhead{Implication Assessment: Outcome Probability}
  Consequence Anticipation
  & SW: At the time of experiencing the emotion, did you think that the real or potential consequences of the event
    had already been felt by you or were completely predictable?
               & At that time I anticipated the consequences of the event.
  \\\sep
  \horizhead{Implication Assessment: Urgency}
  Response\par urgency
  & SW: After you had a good idea of what the probable consequences of the event would be, did you think 
    that it was urgent to act immediately?
               & The event required an immediate response.
  \\\sep
  \horizhead{Coping Potential: Control}
  Own, Others', Chance Control
  & SE: When you were feeling happy, to what extent did you feel that\par you had the ability to influence what was happening in this situation?\par someone other than yourself was controlling what was happening in this situation?\par circumstances beyond anyone’s control were controlling what was happening in this situation?
               & I had the capacity to affect what was going on during the event.\par Someone or something other than me was influencing what was going on during the situation.\par The situation was the result of outside influences of which nobody had control.
  \\\sep
  \horizhead{Coping Potential: Adjustment Check}
  Anticipated Acceptance
       & SW: After you had a good idea of what the probable consequences of the event would be, did you think that you could live with, and adjust to, the consequences of the event that could not be avoided or modified?
               & I anticipated that I could live with the unavoidable consequences of the event.
  \\\sep
  Effort
       & SE: When you were feeling happy, how much effort (mental or physical) did you feel this situation required you to expend?
               & The situation required me a great deal of energy to deal with it.
  \\\sep
  \horizhead{Normative Significance: Control}
  Internal Standards Compatibility
       & SW: At the time of experiencing the emotion, did you think that the actions that produced the event were morally and ethically acceptable?
               & The event clashed with my standards and ideals.
  \\\sep
  External Norms Compatibility
       & SW: At the time of experiencing the emotion, did you think that the actions that produced the event violated laws or social norms?
         & The event violated laws or socially accepted norms.
  \\
 \bottomrule
\end{longtable}
}

\clearpage

\section{Study Details}
\label{study-details}
Table~\ref{tab:generation-overview} reports an overview of the
participants and the cost involved in the generation of
\corpusname. For each round, we indicate the strategy used 
in the text production task:
\begin{compactitem}
\item Strategy 0:  participants were free to write
  any event of their choice.
\item Strategy 1: they were asked to
  recount an event special to their lives.
\item Strategy 2: they were
  shown the list of topics to avoid, described in
  Section~\ref{ssec:data-generation}, Table~\ref{tab:off-limits}).
\end{compactitem}

The row ``Workers'' reports the number of different participants 
accepted in each round, hence in the column ``$\sum$'' is the total number
of (unique) annotators whose answers entered the corpus 
(with the exception of those who contributed to round 1*, the pretest
that we do not include in \corpusname).
Note that the same worker could
participate in multiple rounds; for this reason the sum of workers
across rounds exceeds 2379.

Table~\ref{tab:validation-overview} shows the same information for the 
validation phase. $\sum$=\textsterling\,1768.09 refers to the cost prior to releasing the bonus:
we rewarded an extra payment of \textsterling\,5
to the 60 best performing validators, amounting to \textsterling\,420 (i.e., \textsterling\,300
for the bonus in total + commission charges).

\vspace{5mm}
\begingroup
\small
\setlength{\tabcolsep}{3pt}
\captionof{table}{Generation: overview of study details, for each round separately (with the relative text variability induction strategies) and by aggregating them.}
\label{tab:generation-overview}
\begin{tabular}{lccccccccccccr}
\toprule
Rounds & 1* & 2 & 3 & \multicolumn{2}{c}{4} &  \multicolumn{2}{c}{5} & 6 & 7 & 8 & 9 & $\sum$ \\
  \cmidrule(l){2-2}\cmidrule(l){3-3}\cmidrule(l){4-4}\cmidrule(l){5-6}\cmidrule(l){7-8}\cmidrule(l){9-9}\cmidrule(l){10-10}\cmidrule(l){11-11}\cmidrule(l){12-12}\cmidrule(l){13-13}
  Strategies & 0 & 0 & 0 & 1 & 2 & 1 & 2 & 2 & 2 & 2 & 2\\
Workers & -- &111&526& \multicolumn{2}{c}{476}& \multicolumn{2}{c}{846}&349&81&13&15& 2379\\
Cost (\textsterling{}) & 156.1&154.7&870.1&571.2&552.3&917.8&858.2&616.7& 102.9&10.5& 14.7& 4825.2\\
\bottomrule
\end{tabular}
\endgroup

\vspace{15mm}
\begin{center}
  \begingroup
  \small
\setlength{\tabcolsep}{5pt}
\captionof{table}{Validation: overview of study details, per round and after aggregation.}
\label{tab:validation-overview}
\begin{tabular}{l c c cccc c}
\toprule
Rounds & 1 & 2 & 3 & 4 & 5 & 6 & $\sum$ \\
\cmidrule(r){2-2}\cmidrule(r){3-3}\cmidrule(r){4-4}\cmidrule(r){5-5}\cmidrule(r){6-6}\cmidrule(r){7-7}\cmidrule(r){8-8}
Workers & 20 & 1048&120 &25 &3&1 & 1217\\
Cost (\textsterling{}) & 84&1474.1 & 167.99&36.4 & 4.2& 1.4&1768.09\\
\bottomrule
\end{tabular}
\end{center}
\endgroup

\clearpage

\section{Details on the Data Collection Questionnaires}

The questionnaires in the generation and the validation
phases of building \corpusname are formulated in a comparable manner. Table~\ref{tab:questionnairetemplate} makes
the variants transparent to the reader, showing differences
between the templates in two phases, and across the multiple rounds in
the generation phase. Screenshots of the 
  questionnaires as presented to the readers are available in the 
  supplementary \ material, together with the corpus data.
  
Note that some workers skipped the demographics- and 
personality-related portion of the survey, which had to be completed
for them to be rewarded. We allowed them to 
answer those questions in a separate form, containing only such
questions. We include it in the supplementary\ material as well.

{
  \small
\begin{longtable}{p{.69cm}p{11cm}p{.8cm}}
  \caption{Template of the Questionnaire. The first column specifies where
  the question has been asked. G$n$: Generation (G$E$: prompted by an emotion $E$, GN: prompted by the label ``no emotion'') with text production strategy $n$ (cf. Section~\ref{study-details}), V: Validation.
  No specification means that it has been asked in all variants.
  For the list of [OFF-LIMITS] topics in $n$=2, refer to Table~\ref{tab:off-limits}.}
  \label{tab:questionnairetemplate}\\
  \toprule
  & Question/Text & Value \\
 \cmidrule(lr){1-1}\cmidrule(lr){2-2}\cmidrule(l){3-3}
  \endfirsthead
  \toprule
  & Question/Text & Value \\
  \cmidrule(r){1-1}\cmidrule(lr){2-2}\cmidrule(l){3-3}
  \endhead
  G$x$ & \textbf{Study on Emotional Events}. Dear participant,
         Thanks for your interest in this study. We aim at understanding your
         evaluation of events in which you either felt a particular emotion
         or did not feel any. Further, we will ask you some demographic and
         personality-related information. The study should take you 4 minutes,
         and you will be rewarded with \textsterling\,0.50.
         Your participation is voluntary. You have to be at least 18 years
         old and a native speaker of English. Feel free to quit at any time
         without giving a reason (note that you won't be paid in this case).
         You can take this survey multiple times. You are also welcome to
         participate to the other versions of the survey that we published on
         Prolific, in which we ask you for your experience with different
         emotions. Note that towards the end of this survey, you will find a
         small set of questions that you only need to answer the first time
         you participate (which will save you time if you'll work on the
         other survey variants).
         The data we collect via Google forms will be used for research
         purposes. It will be made publicly available in an anonymised
         form. We will further write a scientific paper publication about
         this study which can include examples from the collected data (also
         in anonymous form). Nevertheless, please avoid providing information
         that could identify you (such as names, contact details, etc.).
         This study is funded by the German Research Foundation (DFG, Project
         Number KL 2869/1-2). Principle Investigator of this study: Dr. Roman
         Klinger, University of Stuttgart (Germany). Responsible and contact
         person: Enrica Troiano, University of Stuttgart (Germany). For any
         information, contact us at enrica.troiano@ims.uni-stuttgart.de
                  & --- \\
  \cmidrule(r){1-1}\cmidrule(lr){2-2}\cmidrule(l){3-3}
  V & \textbf{Study on Emotional Events}. Dear participant,
      Thanks for your interest in this study. In a previous survey, people
      described events that might have triggered a particular emotion
      in them, and they answered some questions about those events. We
      now ask you to evaluate such events.
      You will read 5 brief event descriptions. For each of them, you
      will be asked the same questions that were answered by the event
      experiencers in the previous survey. Your task is to answer the
      same way as they did. Participants who are able to answer most
      similarly to the original authors will get a bonus of \textsterling\,5. We
      reward this bonus to the best 5\% of participants.
      We will also ask you some demographic and personality-related
      information. There, your task is to provide information about
      yourself, and not about the author of the texts.
      The study should take you 8 minutes, and you will be rewarded with \textsterling\,1.
      Your participation is voluntary. You have to be at least 18
      years old and a native speaker of English. Feel free to quit at
      any time without giving a reason (note that you won't be paid in
      this case).
      The data we collect will be used for research purposes. It will
      be made publicly available in an anonymised form. We will
      further write a scientific paper publication about this study
      which can include examples from the collected data (also in
      anonymous form).
      This study is funded by the German Research Foundation (DFG,
      Project Number KL 2869/1-2). Principle Investigator of this
      study: Dr. Roman Klinger, University of Stuttgart
      (Germany). Responsible and contact person: Enrica Troiano,
      University of Stuttgart (Germany). For any information, contact
      us at enrica.troiano@ims.uni-stuttgart.de & --- \\
  \cmidrule(r){1-1}\cmidrule(lr){2-2}\cmidrule(l){3-3}
  &  
    I confirm that I have read the above information, meet the
    prerequisites for participation and want to participate in
    the study. & Yes/No \\
  \cmidrule(r){1-1}\cmidrule(lr){2-2}\cmidrule(l){3-3}
  &  \textbf{Preliminary Questions}. & \\
  & Please insert your ID as a worker on Prolific. & Text \\
& Do you feel any of the following emotions right now, just before
         starting this survey? 1 means ``not at all'', 5 means ``very
         intensely'' {\color{darkgray}[anger; boredom; disgust; fear; guilt; joy; pride;
      relief; sadness; shame; surprise; trust]}& Matrix with items [1--5]\\
  \cmidrule(r){1-1}\cmidrule(lr){2-2}\cmidrule(l){3-3}  
  G$Ex$ & \textbf{This study is about the emotional experience of}
         $\mathbf{E}$. You will be asked to describe a concrete situation
         or an event which provoked this feeling in you and for which
         you vividly remember both the circumstance and your
         reaction. After that, you will be asked further information
         regarding such emotional experience, by indicating how much
         you agree with some statements on a scale from 1 to 5.
         Note: If you participated in our studies before, please
         describe a different situation now. We cannot accept an
         answer related to the same event you already told us about,
         even if you used different words. Further, we will not accept
         answers if they are not descriptions of events, like "I can't
         remember" or "I do not have that feeling".
                  & --- \\
   G$Nx$ & \textbf{This study is about an experience you had, which 
   did not involve you emotionally.} You will be asked to describe a concrete 
   situation or an event which did not provoke any particular feeling in you and 
   for which you vividly remember both the circumstance and your reaction. 
   After that, you will be asked further information regarding such experience, 
   by indicating how much you agree with some statements on a scale from 1 to 5. 
Note: If you participated in our studies before, please describe a different 
situation now. We cannot accept an answer related to the same event you already
told us about, even if you used different words. Further, we will not accept answers 
if they are not descriptions of events, like "I can't remember" or "I always have feelings".   & --- \\
  V& \textbf{Put yourself in the shoes of other people.} You will read
     five texts. These texts describe events that occurred in the life
     of their authors. Don’t be surprised if they are not perfectly
     grammatical, or if you find that some words are missing. For each
     event, you will assess if it provoked an emotion in the
     experiencer, and if so, what emotion that was. Moreover, you will
     be asked how you think the experiencer assessed the event: you
     will read some statements and indicate how much you agree with
     each of them on a scale from 1 to 5. The writers of these texts
     have answered these questions in a previous survey. Your goal now
     is to guess the answer given by the writers as closely as
     possible. & --- \\
  \cmidrule(r){1-1}\cmidrule(lr){2-2}\cmidrule(l){3-3}
  G$Ex$ &  \textbf{Recall an event that made you feel $E$}. Recall an
       event that made you feel $E$ in the past. & --- \\
  G$Nx$ & \textbf{Recall an event that did not make you feel any emotion in the past.}  & --- \\
   \cmidrule(r){1-1}\cmidrule(lr){2-2}\cmidrule(l){3-3}
    &   It could be an
       event of your choice, or one which you might have experienced
       in one of the following areas:
       health, career, finances, community, fun/leisure, sports, arts,
       personal relationships, travel, education, shopping, learning,
       food, nature, hobbies, work...
       Please describe the event by completing the sentence below,
       including event details or write multiple sentences if this
       helps to understand the situation.& --- \\
        \cmidrule(r){1-1}\cmidrule(lr){2-2}\cmidrule(l){3-3}
  G1 & The event should be special to you, or one which you think the other participants of this survey are unlikely to have experienced. It does not need to be an extraordinary event: it should just tell something about yourself.& --- \\
  G2 & NOTE: We already collected many answers related to [OFF-LIMITS].
       Please recount an event which does not relate to any of these:
       we need events which are as diverse as possible!
                  & --- \\  
  \cmidrule(r){1-1}\cmidrule(lr){2-2}\cmidrule(l){3-3}
  G$Ex$ & Please complete the sentence: I felt $E$ when/because/...
                  &
                    Text
  \\
   G$Nx$ & Please complete the sentence: I felt NO PARTICULAR EMOTION when/because/...
                  &
                    Text
  \\
  V & What do you think the writer of the text felt when experiencing
      this event? {\color{darkgray}[anger; boredom; disgust; fear; guilt; joy; pride;
      relief; sadness; shame; surprise; trust; no emotion]} & single choice \\
      V & How confident are you about your answer? & 1\ldots5\\
  G$x$ & How long did the event last? {\color{darkgray}[seconds; minutes; hours; days;
         weeks]} & single choice \\
         V & How long do you think the event lasted? {\color{darkgray}[seconds; minutes; hours; days;
         weeks]} & single choice \\
  G$Ex$ & How long did the emotion last?  {\color{darkgray}[seconds; minutes; hours; days;
         weeks]} & single choice \\
  G$Nx$ & How long did the emotion last (if you had any)?  {\color{darkgray}[seconds; minutes; hours; days;
         weeks; I had none]} & single choice \\
   V & How long do you think the emotion lasted (if the experiencer had any)? {\color{darkgray}[seconds; minutes; hours; days;
         weeks;  this event did not cause any emotion]} & single choice \\
  
  G$x$ & How intense was your experience of the event? & 1\ldots5 \\
  V& How intense do you think the emotion was?  & 1\ldots5 \\
G$x$  & How confident are you that you recall the event well? & 1\ldots5 \\
  
  \cmidrule(r){1-1}\cmidrule(lr){2-2}\cmidrule(l){3-3}
  G$x$ &  \textbf{Evaluation of that experience}. Think back to when the
    event happened and recall its details. Take some time to remember it
    properly. How much do these statements apply? (1 means "Not at all''
         and 5 means "Extremely'') & --- \\
  V & \textbf{Evaluation of that Experience}. Put yourself in the
      shoes of the writer at the time when the event happened, and try
      to reconstruct how that event was perceived. How much do these
      statements apply? (1 means “I don’t agree at all” and 5 means “I
      completely agree”) \\
  & The event was sudden or abrupt. & 1\ldots5 \\
  G$x$ &  The event was familiar. &1\ldots5 \\
  V&The event was familiar to its experiencer.	 & 1\ldots5 \\
  G$x$  &  I could have predicted the occurrence of the
          event. &1\ldots5 \\
  V&The experiencer could have predicted the occurrence of the event.	 & 1\ldots5 \\
  G$x$  &  The event was pleasant for me. &1\ldots5 \\
  V&The event was pleasant for the experiencer.	 & 1\ldots5 \\
  G$x$  &  The event was unpleasant for me. &1\ldots5 \\
  V&The event was unpleasant for the experiencer.	 & 1\ldots5 \\
 G$x$  &  I expected the event to have important consequences for
         me.&1\ldots5 \\
  V&The experiencer expected the event to have important consequences
     for him/herself.	 & 1\ldots5 \\
&  The event was caused by chance, special circumstances, or natural
         forces. &1\ldots5 \\
  G$x$  &  The event was caused by my own behavior.&1\ldots5 \\
  V&The event was caused by the experiencer’s own behavior.	 & 1\ldots5 \\
&  The event was caused by somebody else’s behavior. & 1\ldots5 \\
  G$x$  &  I anticipated the consequences of the event. & 1\ldots5 \\
  V&The experiencer anticipated the consequences of the event.	 & 1\ldots5 \\
  G$x$  &  I expected positive consequences for me. &1\ldots5 \\
  V&The experiencer expected positive consequences for her/himself.	 & 1\ldots5 \\
  &  The event required an immediate response.& 1\ldots5 \\
 G$x$  &  I was able to influence what was going on during the event.&
                                                                       1\ldots5 \\
  V&The experiencer was able to influence what was going on during the event.	 & 1\ldots5 \\
 G$x$  &  Someone other than me was influencing what was going on.&
                                                                    1\ldots5 \\
  V&Someone other than the experiencer was influencing what was going on.	 & 1\ldots5 \\
&  The situation was the result of outside influences of
         which nobody had control. &1\ldots5 \\
 G$x$  &  I anticipated that I would easily live with the unavoidable
         consequences of the event.& 1\ldots5 \\
  V&The experiencer anticipated that he/she could live with the unavoidable consequences of the event.	 & 1\ldots5 \\
 G$x$  &  The event clashed with my standards and ideals. & 1\ldots5
  \\
  V&The event clashed with her/his standards and ideals.	 & 1\ldots5 \\
&  The actions that produced the event violated laws or socially
         accepted norms.& 1\ldots5 \\
  G$x$  &  I had to pay attention to the situation.& 1\ldots5 \\
  V&The experiencer had to pay attention to the situation.	 & 1\ldots5 \\
  G$x$  &  I tried to shut the situation out of my mind.& 1\ldots5 \\
  V&The experiencer wanted to shut the situation out of her/his mind.	 & 1\ldots5 \\
 G$x$  &  The situation required me a great deal of energy to deal with it. &
                                                                         1\ldots5
  \\
  V&The situation required her/him a great deal of energy to deal with
     it.  & 1\ldots5 \\
      \cmidrule(r){1-1}\cmidrule(lr){2-2}\cmidrule(l){3-3}
  V&\textbf{Have you ever experienced an event similar to the one
     described?} &  \\
  &  I experienced a similar event before. & 1\ldots5\\
  \cmidrule(r){1-1}\cmidrule(lr){2-2}\cmidrule(l){3-3}
  G$x$ & \textbf{Is this the first time you participate in one of our
         emotional-event recollection studies?} We would like to know a bit more about you now. We have multiple
         similar studies on Prolific, all called "Recollection of an
         emotion-inducing experience", with the word "emotion" being replaced
         by an actual emotion name. When you participate in more than one of
         these studies, you only need to answer the following questions once. If this
         is the first time you participate, please answer them (otherwise we
         won't be able to approve your contribution), later you will skip
         this step.
         {\color{darkgray}[Yes, first time, I will answer the following questions.;
         No, I participated before and answered the next set of questions.]}
         & single choice \\
    V&     \textbf{Is this the first time you participate in our event evaluation studies?}
     If yes, you need to answer the following questions (otherwise we won't be able to approve your contribution). If no, you can skip them. 
     {\color{darkgray}[Yes, first time, I will answer the following questions.;
     No, I participated before and answered the next set of questions.]}
     &  single choice\\ 
     \cmidrule(r){1-1}\cmidrule(lr){2-2}\cmidrule(l){3-3}
  G$x$ & \textbf{Demographic and Personality-related Questions}. As a
    last step, we ask you to answer some questions about
    yourself. Note: if you take one of our studies in the future,
    you won't fill in these sections again; if this is your first
    time and don't provide such information, we won't be able to
    reward you. & --- \\
  & How old are you? & int \\
  & With which gender do you identify? {\color{darkgray}Female; Male; Gender Variant/Non-Conforming; Prefer not to answer]} & single choice \\
  & What is the highest level of education you completed? {\color{darkgray}[No formal
    qualifications; Secondary education; High school; Undegraduate
    degree (BA/BSc/other); Graduate degree (MA/MSc/MPhil/other);
    Doctorate degree (PhD/other); Don't know/ not applicable]} & single
                                                                choice \\
  & With which of the following ethnic groups do you identify the
    most? {\color{darkgray}[Australian/New Zealander;
    North Asian; South Asian; East Asian; Middle Eastern; European; African;
    North American; South American; Hispanic/Latino; Indigenous;
    Prefer not to answer; Other...]} & single choice \\
  &  Here are a number of personality traits that may or may not apply
    to you. You should rate the extent to which the pair of traits applies to you, even
    if one characteristic applies more strongly than the
    other. {\color{darkgray}[Extraverted, enthusiastic; Critical, quarrelsome;
    Dependable, self-disciplined; Anxious, easily upset; Open to new
    experiences, complex; Reserved, quiet; Sympathetic, warm;
    Disorganized, careless; Calm, emotionally stable; Conventional,
    uncreative]} & Matrix with items [1\dots7]\\
  \cmidrule(r){1-1}\cmidrule(lr){2-2}\cmidrule(l){3-3}
 
G$x$ & \textbf{One Last Question}. Please be assured that your answer
       will in no way influence how we treat your submission (you will
       be rewarded, if you properly followed our instructions).
	Did you actually experience that event or did you make it up
         to? {\color{darkgray}[The event really happened in my life.; I never
         experienced that event, but I really imagined how it would
         make me feel.]} & single choice \\
        \bottomrule
\end{longtable}
}

\clearpage

\section{Details on Results}
Our modeling results for the task of predicting appraisals are
averages across 5 runs of the model.  In
Tables~\ref{tab:table-appraisals-sd},
\ref{tab:table-appraisals2emotions-sd}, and
\ref{tab:table-emotions-sd}, we complement such results with
standard deviation values.

\begin{table}[h!]
    \caption{Appraisal prediction performance of text classifiers and
      regressors (\TAmodelText) as \F and $\textrm{RMSE}$.}
      \label{tab:table-appraisals-sd}
\newcommand{\sd}[2]{#1{\tiny$\pm$#2}}
\centering\small
        \begin{tabular}{lrrrr}
            \toprule
            & \multicolumn{2}{c}{Classification} & \multicolumn{2}{c}{Regression} \\
            \cmidrule(lr){2-3}\cmidrule(lr){4-5}
            & \TAhuman  & \TAmodel &  \TAhuman & \TAmodel \\
            \cmidrule(r){2-2}\cmidrule(lr){3-3}\cmidrule(lr){4-4}\cmidrule(lr){5-5}
            Appraisal       & \F & \F & $\textrm{RMSE}$ & $\textrm{RMSE}$ \\
            \cmidrule(r){1-1}\cmidrule(rl){2-2}\cmidrule(rl){3-3}\cmidrule(rl){4-4}\cmidrule(rl){5-5}
    Suddenness & $.68$ & \sd{.74}{.02} & $1.47$ & \sd{1.33}{.05} \\
    Familiarity & $.53$ & \sd{.79}{.00} & $1.49$ & \sd{1.42}{.09} \\
    Event Pred. & $.56$ & \sd{.75}{.01} & $1.46$ & \sd{1.47}{.17} \\
    Pleasantness & $.83$ & \sd{.88}{.01} & $1.1$ & \sd{1.30}{.06} \\
    Unpleasantness & $.85$ & \sd{.80}{.01} & $1.22$ & \sd{1.26}{.05} \\
    Goal Relevance & $.66$ & \sd{.71}{.01} & $1.52$ & \sd{1.57}{.17} \\
    Situat. Resp. & $.48$ & \sd{.85}{.01} & $1.55$ & \sd{1.43}{.09} \\
    Own Resp. & $.73$ & \sd{.79}{.01} & $1.32$ & \sd{1.40}{.11} \\
    Others' Resp. & $.74$ & \sd{.73}{.02} & $1.54$ & \sd{1.57}{.24} \\
    Anticip. Conseq. & $.52$ & \sd{.69}{.02} & $1.61$ & \sd{1.50}{.11} \\
    Goal Support & $.67$ & \sd{.81}{.01} & $1.36$ & \sd{1.33}{.12} \\
    Urgency & $.54$ & \sd{.61}{.03} & $1.68$ & \sd{1.43}{.05} \\
    Own Control & $.53$ & \sd{.79}{.01} & $1.48$ & \sd{1.35}{.08} \\
    Others' Control & $.76$ & \sd{.62}{.01} & $1.55$ & \sd{1.36}{.07} \\
    Situat. Control & $.51$ & \sd{.87}{.01} & $1.53$ & \sd{1.35}{.06} \\
    Accept. Conseq. & $.43$ & \sd{.64}{.02} & $1.77$ & \sd{1.44}{.06} \\
    Internal Standards & $.57$ & \sd{.82}{.01} & $1.44$ & \sd{1.36}{.09} \\
    External Norms & $.56$ & \sd{.92}{.00} & $1.16$ & \sd{1.34}{.15} \\
    Attention & $.74$ & \sd{.48}{.04} & $1.38$ & \sd{1.27}{.07} \\
    Not Consider & $.54$ & \sd{.77}{.03} & $1.56$ & \sd{1.53}{.13} \\
    Effort & $.61$ & \sd{.70}{.03} & $1.47$ & \sd{1.38}{.06} \\
    \midrule
    Average & $.62$ & \sd{.75}{.00} & $1.46$ & \sd{1.40}{.10} \\
        \bottomrule
    \end{tabular}
\end{table}

\vfill
\mbox{}
\clearpage

\begin{table}
    \caption{Emotion prediction performance of classifiers using appraisals as input (\AEmodelGoldText, \AEmodelPredText).}
    \label{tab:table-appraisals2emotions-sd}
\newcommand{\sd}[2]{#1{\tiny$\pm$#2}}
\centering\small
\begin{tabular}{lcccc}
    \toprule
     & \multicolumn{2}{c}{Discretized (1)} & \multicolumn{2}{c}{Scaled (2)} \\
    \cmidrule(r){2-3}\cmidrule(r){4-5}
& \AEmodelGold& \AEmodelPred& \AEmodelGold& \AEmodelPred\\
    \cmidrule(r){2-2}\cmidrule(r){3-3}\cmidrule(r){4-4}\cmidrule(r){5-5}
    Emotion & \F & \F & \F & \F \\
    \cmidrule(r){1-1}
    \cmidrule(r){2-2}\cmidrule(r){3-3}\cmidrule(r){4-4}\cmidrule(r){5-5}
    Anger & \sd{.37}{.04} & \sd{.32}{.04} & \sd{.37}{.01} & \sd{.37}{.01}\\
    Boredom & \sd{.46}{.02} & \sd{.60}{.02} & \sd{.54}{.01} & \sd{.52}{.01}\\
    Disgust & \sd{.36}{.03} & \sd{.37}{.03} & \sd{.45}{.01} & \sd{.29}{.01}\\
    Fear & \sd{.26}{.03} & \sd{.32}{.03} & \sd{.30}{.01} & \sd{.36}{.01}\\
    Guilt & \sd{.23}{.03} & \sd{.19}{.03} & \sd{.30}{.04} & \sd{.18}{.04}\\
    Joy & \sd{.30}{.05} & \sd{.30}{.05} & \sd{.32}{.04} & \sd{.25}{.04}\\
    No-emotion & \sd{.46}{.03} & \sd{.35}{.03} & \sd{.46}{.02} & \sd{.31}{.02}\\
    Pride & \sd{.35}{.05} & \sd{.36}{.05} & \sd{.33}{.05} & \sd{.28}{.05}\\
    Relief & \sd{.18}{.04} & \sd{.19}{.04} & \sd{.21}{.04} & \sd{.27}{.04}\\
    Sadness & \sd{.29}{.05} & \sd{.34}{.05} & \sd{.34}{.03} & \sd{.32}{.03}\\
    Shame & \sd{.18}{.04} & \sd{.24}{.04} & \sd{.24}{.04} & \sd{.31}{.04}\\
    Surprise & \sd{.44}{.03} & \sd{.44}{.03} & \sd{.41}{.02} & \sd{.28}{.02}\\
    Trust & \sd{.24}{.02} & \sd{.15}{.02} & \sd{.21}{.06} & \sd{.27}{.06}\\
    \cmidrule(r){1-1}\cmidrule(r){2-2}\cmidrule(r){3-3}\cmidrule(r){4-4}\cmidrule(r){5-5}
    Macro avg. & \sd{.31}{.01} & \sd{.32}{.01} & \sd{.35}{.02} & \sd{.31}{.02}\\
    \bottomrule
\end{tabular}
\end{table}

\begin{table}
    \caption{Emotion prediction performance of appraisal-aware text classifiers (\TEmodelText, \TAEmodelGoldText, \TAEmodelPredText).}
    \label{tab:table-emotions-sd}
\newcommand{\sd}[2]{#1{\tiny$\pm$#2}}
\centering\small
        \begin{tabular}{lcccc}
            \toprule
            & \TEhuman  & \TEmodel &  \TAEmodelGold & \TAEmodelPred \\
            \cmidrule(lr){2-3}\cmidrule(lr){4-5}\cmidrule(lr){5-5}
            Emotion       & \F & \F & \F & \F \\

            \cmidrule(r){1-1}\cmidrule(rl){2-2}\cmidrule(rl){3-3}\cmidrule(rl){4-4}\cmidrule(rl){5-5}
            Anger & $.57$ & \sd{.53}{.05} & \sd{.57}{.02} & \sd{.57}{.02} \\
            Boredom & $.73$ & \sd{.84}{.01} & \sd{.83}{.03} & \sd{.83}{.03} \\
            Disgust & $.65$ &\sd{.66}{.00} & \sd{.66}{.04} & \sd{.66}{.04} \\
            Fear & $.73$ & \sd{.65}{.03} & \sd{.67}{.04} & \sd{.67}{.03} \\
            Guilt & $.53$ & \sd{.48}{.06} & \sd{.58}{.05} & \sd{.56}{.07} \\
            Joy & $.49$ & \sd{.45}{.02} & \sd{.48}{.03} & \sd{.47}{.03} \\
            No-emotion & $.33$ & \sd{.55}{.01} & \sd{.56}{.02} & \sd{.56}{.01} \\
            Pride & $.59$ &\sd{.54}{.03} & \sd{.55}{.01} & \sd{.55}{.01} \\
            Relief & $.64$ &\sd{.63}{.02} & \sd{.62}{.01} & \sd{.62}{.02} \\
            Sadness & $.63$ &\sd{.59}{.03} & \sd{.65}{.01} & \sd{.63}{.00} \\
            Shame & $.48$ &\sd{.51}{.01} & \sd{.50}{.08} & \sd{.49}{.07} \\
            Surprise & $.42$ &\sd{.53}{.02} & \sd{.49}{.03} & \sd{.50}{.02} \\
            Trust & $.52$ & \sd{.74}{.02} & \sd{.73}{.04} & \sd{.72}{.03} \\
            \cmidrule(r){1-1}\cmidrule(rl){2-2}\cmidrule(rl){3-3}\cmidrule(rl){4-4}\cmidrule(rl){5-5}
            Macro avg. & $.56$ & \sd{.59}{.01} & \sd{.61}{.02} & \sd{.60}{.02} \\
            \bottomrule
    \end{tabular}
\end{table}

\clearpage

\section{Appraisal Labels across Generation and Validation}
\begin{figure}[h!]
  \centering\small
  \includegraphics[scale=0.50]{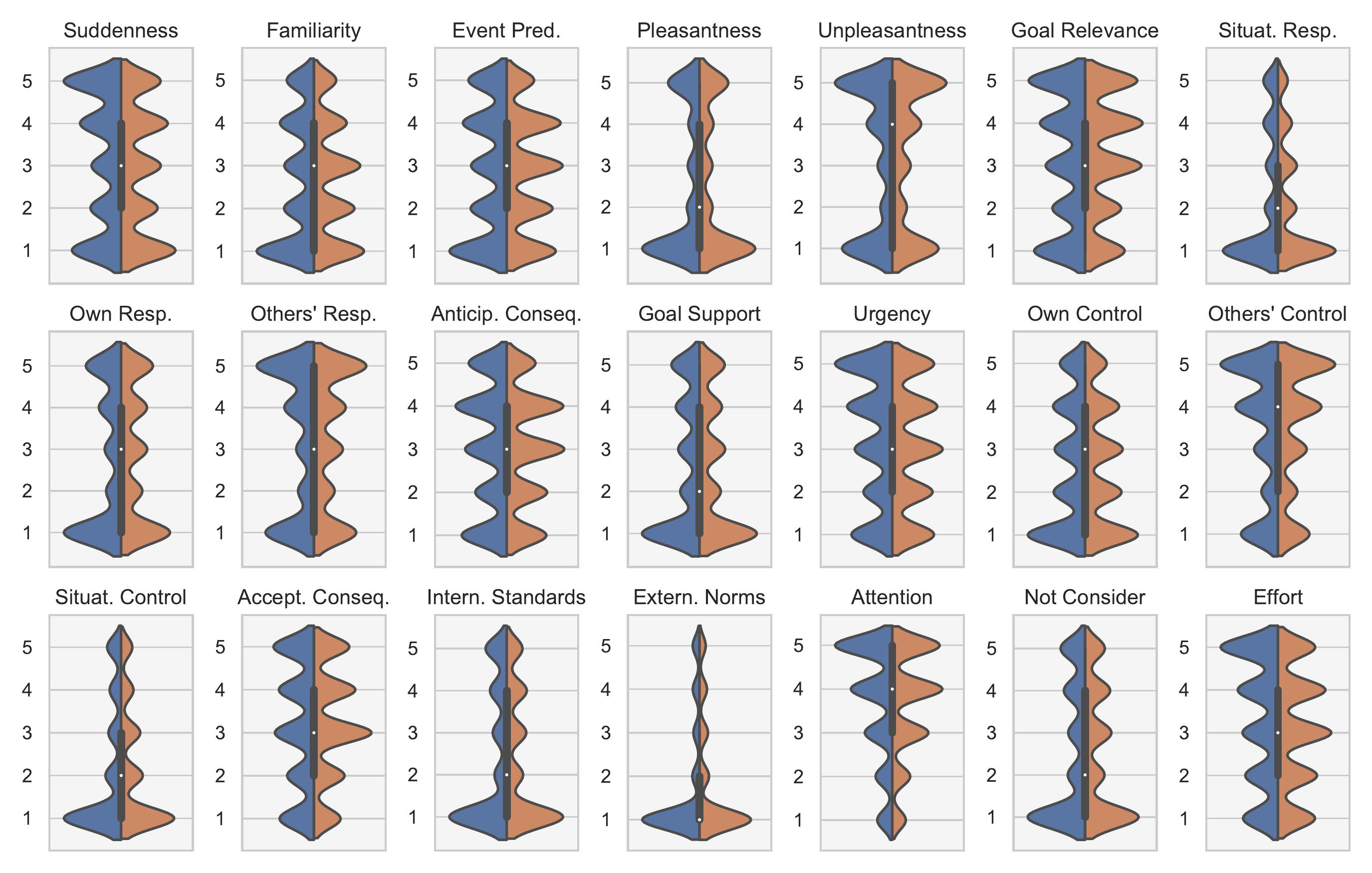}
  \caption{Distributions of appraisal ratings from the two phases  
  of corpus constructions (blue: generation, orange: validation).}
  \label{fig:plot-violins}
\end{figure}

Figure~\ref{fig:plot-violins} shows the distributions of the 21 appraisal
variables. The width of a curve visualizes the relative
frequency of the 5 values for the label in question. The left side (blue)
of each plot represents the generation phase and the right part (orange) corresponds to the
validation-based annotations.

\vfill

\mbox{}

\end{document}